\documentclass[draftclsnofoot, compsoc, peerreview, onecolumn, 12pt]{IEEEtran}

\usepackage[table,xcdraw]{xcolor}
\usepackage{enumitem}
\usepackage{multirow}
\usepackage{url}
\usepackage{bbm}
\usepackage{amsmath,amssymb,amsfonts,amsthm}
\usepackage{booktabs} 
\usepackage{algorithm}
\usepackage{algorithmic}
\usepackage[nocompress]{cite}
\usepackage[caption=false,font=footnotesize,labelfont=sf,textfont=sf]{subfig}
\ifCLASSINFOpdf
  \usepackage[pdftex]{graphicx}
  \graphicspath{{../image/}}
  \DeclareGraphicsExtensions{.pdf,.jpeg,.png}
\else
  \usepackage[dvips]{graphicx}
  \graphicspath{{../image/}}
  \DeclareGraphicsExtensions{.eps}
\fi

\newtheorem{proposition}{Proposition}

\newtheorem{remark}{Remark}
\usepackage{amsmath}
\usepackage{subcaption}
\usepackage{hyperref}
\hypersetup{colorlinks=true,linkcolor=black}

\newcommand{\onlineurl}[1]{[Online]. Available: \textcolor{blue}{\href{#1}{\nolinkurl{#1}}}}

\begin{document}
\title{Information-Theoretic Optimization for Task-Adapted Compressed Sensing Magnetic Resonance Imaging}

\author{
Xinyu Peng,
Ziyang Zheng*,~\IEEEmembership{Member,~IEEE,}
Wenrui~Dai*,~\IEEEmembership{Member,~IEEE,}
Duoduo~Xue,~\IEEEmembership{Member,~IEEE,}
Shaohui~Li,~\IEEEmembership{Member,~IEEE,}
Chenglin~Li,~\IEEEmembership{Member,~IEEE,} 
Junni~Zou,~\IEEEmembership{Member,~IEEE,}
and~Hongkai~Xiong,~\IEEEmembership{Fellow,~IEEE}
\thanks{* Corresponding authors: Ziyang Zheng and Wenrui Dai.}
\thanks{Xinyu Peng is with the Department of Computer Science and Engineering, Shanghai Jiao Tong University, Shanghai 200240, China. E-mail: xypeng9903@sjtu.edu.cn.}
\thanks{Ziyang Zheng, Wenrui Dai, Chenglin Li, Junni Zou, and Hongkai Xiong are with the Department of Electronic Engineering, Shanghai Jiao Tong University, Shanghai 200240, China. E-mail: \{zhengziyang, daiwenrui, lcl1985, zoujunni, xionghongkai\}@sjtu.edu.cn.}
\thanks{Duoduo Xue is with the Department of Computer Science, City University of Hong Kong, Hong Kong SAR, China. E-mail: duoduxue@cityu.edu.hk.}
\thanks{Shaohui Li is with the College of Information Science and Electronic Engineering, Zhejiang University, Hangzhou 310007, China. E-mail: lishaohui@zju.edu.cn.}
\thanks{This is a draft and the final version has been accepted by IEEE TPAMI (DOI: 10.1109/TPAMI.2026.3683201).}
}

\maketitle

\markboth{IEEE Transactions on Pattern Analysis and Machine Intelligence}%
{Peng \MakeLowercase{\textit{et al.}}: Information-Theoretic Optimization for Task-Adapted CS-MRI}

\newpage
\begin{abstract}
Task-adapted compressed sensing magnetic resonance imaging (CS-MRI) is emerging to address the specific demands of downstream clinical tasks with significantly fewer k-space measurements than required by Nyquist sampling. However, existing task-adapted CS-MRI methods suffer from the uncertainty problem for medical diagnosis and cannot achieve adaptive sampling in end-to-end optimization with reconstruction or clinical tasks. To address these limitations, we propose the first task-adapted CS-MRI from the information-theoretic perspective to simultaneously achieve probabilistic inference for uncertainty prediction and adapt to arbitrary sampling ratios and versatile clinical applications. Specifically, we formalize the task-adapted CS-MRI optimization problem by maximizing the mutual information between undersampled k-space measurements and clinical tasks to enable probabilistic inference for addressing the uncertainty problem. We leverage amortized optimization and construct tractable variational bounds for mutual information to jointly optimize sampling, reconstruction, and task-inference models, which enables flexible sampling ratio control using a single end-to-end trained model. Furthermore, the proposed framework addresses two kinds of distinct clinical scenarios within a unified approach, \emph{i.e.}, i) joint task and reconstruction, where reconstruction serves as an auxiliary process to enhance task performance; and ii) task implementation with suppressed reconstruction, applicable for privacy protection. Extensive experiments on large-scale MRI datasets demonstrate that the proposed framework achieves highly competitive performance on standard metrics like Dice compared to deterministic counterpart but provides better distribution matching to the ground-truth posterior distribution as measured by the generalized energy distance (GED).
\end{abstract}


\newpage
\tableofcontents

\newpage
\section{Introduction}

\IEEEPARstart{M}{agnetic} Resonance Imaging (MRI) is a cornerstone for medical diagnostics, renowned for its high contrast and non-invasive nature in providing detailed images of soft tissues. However, the clinical utility of MRI is often limited by the lengthy scan time required to acquire necessary k-space measurement for high-quality imaging. The drive to expedite this process without compromising diagnostic accuracy has stimulated the development of compressed sensing MRI (CS-MRI). Contrary to traditional MRI methods, CS-MRI exploits the inherent sparsity in MR images to reconstruct images with clinically acceptable  quality from significantly fewer samples.

Existing studies on CS-MRI predominantly focus on optimizing image reconstruction using deep learning techniques. Early studies on deep learning-based CS-MRI primarily learn the mapping from the subsampled k-space measurements to the MR images~\cite{sun2016deep, zhang2018ista, RN274}. Recently, joint optimization of MRI sampling patterns and reconstruction has garnered substantial interest~\cite{RN223_Deep, RN276, RN170, RN299, RN283, RN196}. However, these methods are trained for a specified sampling ratio, and thus cannot support adaptive sampling via a single model.

To achieve adaptive sampling and flexibly control sampling ratios, active acquisition methods~\cite{RN209, RN279, RN298, RN224} formalize sampling optimization as a Markov decision process to iteratively generate sampling patterns based on preceding sampling patterns, and  the acquired measurements and the corresponding reconstructed MR images. However, these methods heavily rely on a pre-trained reconstruction model that remains fixed without fine-tuning during sampling optimization. The reconstruction model could fail to align with the sampling patterns generated by the trained sampling strategy and yield  potentially degraded performance~\cite{RN283}.

Beyond reconstruction-centric CS-MRI methods that aim to enhance the reconstruction fidelity of MR images, recent \textit{Task-adapted CS-MRI} methods~\cite{monteiro2020stochastic, schlemper2018cardiac, wu2023learning} directly optimize MR pipeline for accommodating downstream clinical applications~\cite{caballero2014application}. These methods perform downstream tasks directly from k-space measurements without realizing reconstruction or considering reconstruction as an auxiliary process. 
Unfortunately, existing task-adapted CS-MRI methods cannot fully address the inherent \textit{uncertainty problem} that critically affects their reliability in clinical applications~\cite{monteiro2020stochastic,RN199}. The problem arises from the ambiguity in achieving accurate clinical diagnosis due to the factors such as the vague boundaries caused by low imaging quality and diverging expert opinions on the same MR images.
The only one deterministic diagnosis could suffer from the \emph{uncertainty} problem that leads to misdiagnosis or ineffective treatment by disregarding other potentially valid interpretations. The uncertainty becomes even more pronounced for CS-MRI following the Markov chain of ``$\text{task } T\rightarrow \text{ground truth 
image } X \rightarrow \text{measurement } Y$'', especially at low sampling ratios. Due to the data-processing inequality~\cite{cover1999elements}, the uncertainty of task $T$ is inevitably greater given the measurement $Y$ than given the ground truth image $X$. The challenge is also supported by the fact that the presumption of deterministic mapping from \( Y \) to \( T \) is invalid for highly undersampled \( Y \)~\cite{schlemper2018cardiac}. LI-Net~\cite{schlemper2018cardiac} partially addresses the uncertainty problem for segmentation by leveraging latent variables to ensure that predictions lie on the segmentation manifold. However, it remains deterministic and fails to capture all plausible solutions consistent with the measurements.

To address these issues, in this paper, we propose an information-theoretic optimization framework for task-adapted CS-MRI that for the first time simultaneously achieves probabilistic inference for uncertainty prediction and adapts to arbitrary sampling ratios and versatile clinical applications. The proposed framework provides a unified method to optimize acquisition, reconstruction, and downstream clinical tasks for CS-MRI in an end-to-end fashion based on the information-theoretic metric (ITM) \cite{RN47}. Experimental results demonstrate that the proposed framework achieves state-of-the-art performance in task-adapted CS-MRI for various clinical tasks. The contributions of this paper are summarized as below.
\begin{itemize}[leftmargin=0cm, itemindent=0.5cm]
\item \textit{Probabilistic inference for uncertainty prediction:} The proposed framework formalizes task-adapted CS-MRI optimization by maximizing the mutual information between undersampled k-space measurements and clinical tasks to address the uncertainty problem of inferring task \( T \) from measurement \( Y \).
Unlike~\cite{schlemper2018cardiac} that solely predicts segmentation in a deterministic manner, the proposed framework achieves probabilistic uncertainty prediction for both reconstruction-centric and task-adapted CS-MRI.
\item \textit{End-to-end optimized adaptive sampling:} The proposed framework leverages amortized optimization~\cite{amos2023tutorial} and constructs tractable variational bounds for mutual information to support arbitrary sampling ratios for CS-MRI with a single end-to-end trained model. Adaptive sampling is jointly optimized with reconstruction and task-inference models to achieve efficient learning and significant performance gains.
\item \textit{Versatile adaptation to clinical scenarios:} The proposed framework provides a unified approach for a wide range of clinical scenarios characterized by i) jointly optimizing downstream tasks and MR image reconstruction to enhance task performance with auxiliary reconstruction process, and ii) implementing downstream tasks while suppressing MR image reconstruction for scenarios involving privacy-preserving clinical diagnosis and compressed learning~\cite{RN248}. 
\end{itemize}

The rest of the paper is organized as follows. Section~\ref{sec:related-work} reviews the related literature. Section~\ref{sec:proposed-method} elaborates the proposed information-theoretic optimization framework, including the formulation of optimization problems, adaptive sampling via amortized optimization, and procedures addressing the two typical clinical scenarios. Section~\ref{sec:experiements} presents experimental results and comparisons with existing methods. Finally, Section~\ref{sec:conclu} concludes the paper.

Throughout this paper, we use normal symbols for scalars and bold symbols for vectors. We use upper italic cases to represent random
variables (\emph{e.g.}, $X$, $Y$, and $T$) and lower bold cases for their realizations (\emph{e.g.}, $\mathbf{x}$, $\mathbf{y}$, and $\mathbf{t}$).

\section{Related Works} \label{sec:related-work}
\subsection{Deep Learning for MR Sampling Patterns}

Existing approaches for optimizing MRI sampling patterns can be broadly categorized into two classes: \textit{end-to-end methods} and \textit{active acquisition methods}.

\textit{End-to-end methods.} These methods jointly optimize the sampling pattern along with the reconstruction model~\cite{RN223_Deep, RN276, RN170, RN299, RN283, RN196}. To enable the use of stochastic gradient-based optimization for model parameters, these approaches typically require a continuous relaxation of the binary sampling patterns.

\textit{Active acquisition methods.} In these approaches, sampling patterns are generated iteratively based on previously acquired information~\cite{RN209, RN279, RN298, RN224, bakker2022learning}. A pioneering work~\cite{RN209} in this category selects the next measurement that minimizes the current uncertainty. More recently, reinforcement learning tools have been adopted, framing the problem as a partially observable Markov decision process~\cite{RN279, RN298, RN224}.

Different from end-to-end optimized methods that are generally specialized for a fixed sampling ratio, active acquisition methods allow for continuous control of the sampling ratio due to their sequential nature. However, in terms of performance, end-to-end methods significantly outperform active acquisition methods, as demonstrated in~\cite{RN283}. This performance gap is primarily attributed to training-test mismatches arising from the reliance on pre-trained reconstruction models in active acquisition methods.

\subsection{Optimization of Mutual Information}
In this paper, we aim to leverage mutual information for optimizing MR sampling patterns. The criterion for optimizing linear projection (sampling pattern) in CS based on mutual information has its roots in Bayesian CS~\cite{seeger2008compressed, ji2008bayesian}. However, the application of this criterion presents a challenge due to the well-known difficulty in estimating or optimizing Mutual Information. Traditional methods typically rely on approximate Bayesian inference~\cite{seeger2008compressed, nickisch2008bayesian} to handle the intractable posterior involved in mutual information. Recently, approaches utilizing amortized variational inference have been introduced to enable more efficient and accurate posterior inference~\cite{RN49, foster2019variational}, as well as contrastive bounds to bypass explicit posterior computation~\cite{foster2021deep}. Additionally, mutual information can be estimated using various techniques, including ratio estimation~\cite{kleinegesse2019efficient}, neural estimation~\cite{kleinegesse2020bayesian}, and Laplace importance sampling~\cite{beck2018fast}, among others.

\subsection{Task-adapted CS-MRI} \label{sec:taskadaptedmri}
Task-adapted CS-MRI focuses on optimizing the performance of medical tasks based on subsampled k-space measurements, where reconstruction either does not occur or serves as an auxiliary process. Here, we briefly review the two categories of application scenarios for task-adapted CS-MRI.

\textit{Joint Task and Reconstruction.}
Simultaneously inferring $T$ and reconstructing $X$ could provide references or evaluate model predictions for clinical decision making. Pioneering work \cite{caballero2014application} achieved joint reconstruction and segmentation from undersampled k-space measurement based on patch-based sparse representation and Gaussian mixture models. 
In recent years, deep learning has significantly advanced task-adapted CS-MRI. Most deep learning-based task-adapted CS-MRI methods~\cite{RN299, RN258, RN236, RN201, caliva2020breaking} adopt a two-stage model design: first, constructing a reconstruction model with k-space measurements as input, and then cascading a task-inference model for clinical tasks. Both models are jointly optimized using a multi-task loss. More recently, \cite{wu2023learning} introduced a similar model design but proposed a two-step training strategy. This approach trains the reconstruction model independently and fine-tunes both models for clinical tasks, optimizing solely the task-specific loss during the second step.

\textit{Compressed Learning.}
Compressed learning focuses on inferring downstream tasks~\cite{RN248, RN223_Deep} without requiring reconstruction results. This approach is particularly desirable for privacy-preserving medical applications, such as MRI~\cite{RN248}, where sharing sensitive patient data is not feasible. In~\cite{schlemper2018cardiac}, a direct mapping from undersampled k-space measurement to segmentation maps was proposed. However, such methods fail to address the inevitable uncertainty caused by information loss in CS-MRI. LI-Net~\cite{schlemper2018cardiac} partially addresses the uncertainty problem by generating segmentation maps based on latent variables to ensure that predictions lie on the segmentation manifold. Nonetheless, this deterministic approach does not fully resolve the uncertainty issue, as it cannot capture all plausible solutions consistent with the measurement. In Section~\ref{sec:relation}, we make a comprehensive comparison between the proposed method and LI-Net.

\subsection{Stochastic Medical Image Segmentation} 

The advent of deep learning has revolutionized the field of medical image segmentation~\cite{isensee2021nnu}. A critical challenge in this domain is that inferring segmentations from medical images is inherently ambiguous, with annotations from domain experts often exhibiting significant variability. Conventional methods typically rely on pixel-wise categorical distributions to model the conditional distribution of segmentation \(T\) given an MRI image \(X\). However, this approach can be insufficient due to the multimodal nature of the distribution and the necessity for spatial coherence in segmentation~\cite{monteiro2020stochastic}.

To address these challenges, stochastic semantic segmentation methods have emerged, leveraging more expressive distributions to model segmentation uncertainty. These methods primarily integrate deep generative models to generate multiple plausible predictions for a single image, such as variational autoencoders~\cite{RN197, RN230, RN199}, autoregressive models~\cite{zhang2022pixelseg}, normalizing flows~\cite{selvan2020uncertainty}, and diffusion models~\cite{zbinden2023stochastic}. Additionally, several approaches avoid deep generative models, instead employing techniques such as low-rank multivariate normal distributions to model logit distributions~\cite{monteiro2020stochastic} or mixtures of stochastic experts for mode estimation~\cite{gao2022modeling}.

\section{Proposed Method}\label{sec:proposed-method}

This section delves into the optimization of task-adapted CS-MRI from an information-theoretic perspective, enabling probabilistic inference for uncertainty prediction and achieving an end-to-end optimized adaptive CS-MRI framework. Section~\ref{sec:background} introduces the preliminaries on information-theoretic metrics (ITMs) and the linear feature design problem.
Section~\ref{sec:Variational_Information_Optimization} formulates a linear feature design problem based on ITMs for optimizing task-adapted CS-MRI given a fixed sampling ratio.
Section~\ref{sec:amortize} further solves multiple problem instances of task-adapted CS-MRI characterized by varying sampling ratios, and achieves adaptive sampling through a single end-to-end trained model via amortized optimization.
Section~\ref{sec: beta<0} achieves joint optimization of downstream tasks and MRI signal reconstruction, and Section~\ref{sec:beta>0} realizes downstream tasks by removing irrelevant MRI reconstruction to enable privacy preserving medical diagnosis and compressed learning.
Section~\ref{sec:relation} highlights the contributions of the proposed method and discusses its relations to existing methods.

\subsection{Preliminary: Information Theoretic Metric} \label{sec:background}
Consider the linear observation model $\mathbf{y} = \mathbf{\Phi x} + \mathbf{n}$, where $\mathbf{\Phi} \in \mathbb{C}^{M \times N}$ is the measurement matrix, $\mathbf{n}$ represents the noise, $\mathbf{x}$ and $\mathbf{y}$ are realizations of the signal $X$ and observation $Y$, respectively. The Information Theoretic Metric (ITM)~\cite{RN47} provides a concise framework for optimizing $\mathbf{\Phi}$ given the downstream task $T$ that can be inferred from $X$, also known as the \textit{linear feature design problem}. ITM builds a Markov sequence $T\rightarrow X \rightarrow Y$ 
\footnote{Our framework's validity hinges on the conditional independence $T \perp Y \mid X$, meaning the measurements $Y$ are uninformative about the task $T$ given the image $X$. This assumption is invariant to the causal direction, making graphical models like $T \to X \to Y$ and $T \leftarrow X \to Y$ equivalent.} to characterize $X$, $Y$, and $T$, and optimizes $\mathbf{\Phi}$ by balancing the mutual information $I(Y; T)$ for measuring the information preserved in the observation $Y$ for the downstream task $T$ and $I(Y;X)$ for assessing the quality of recovering the signal $X$ from $Y$:
\begin{equation}\label{eq:itm}
\max_{\mathbf{\Phi}}~\text{ITM}(\mathbf{\Phi}, \beta) := \max_{\mathbf{\Phi}} \{ I(Y; T) - \beta I(Y; X) \}.
\end{equation}
Here, the parameter $\beta\in \mathbb{R}$ controls the balance between $I(Y; T)$ and $I(Y; X)$. When $\beta < 0$, signal recovery and downstream task inference are simultaneously considered. In the extreme cases, the downstream task inference is prioritized in analogy to compressed learning~\cite{RN248} for $\beta=0$, while signal recovery is solely considered as $\beta \rightarrow -\infty$. On the contrary, when $\beta > 0$, ITM behaves similarly to the information bottleneck (IB)~\cite{RN252, RN88} that maximally preserves the information about $T$ in $Y$, while discarding the remaining irrelevant information. \textit{Noting that $\mathbf{\Phi}$ does not explicitly appear in ITM($\mathbf{\Phi}, \beta$) but influences $Y$. We omit $\mathbf{\Phi}$ (or equivalent symbols) without introducing confusion.
}

\subsection{End-to-end Variational Information Optimization for Task-Adapted CS-MRI}
\label{sec:Variational_Information_Optimization}
CS-MRI accelerates the MRI acquisition process with subsampling in the k-space. Given a sampling ratio $r=M/N<1$, subsampled measurements $\mathbf{y}\in\mathbb{C}^M$ for a vectorized MR image $\mathbf{x}\in\mathbb{C}^N$ are acquired in the k-space using a linear projection model with $\mathbf{\Phi}\!=\!\mathbf{U}_{\mathbf{m}}\mathcal{F}$, where $\mathbf{U_m}\!\in\!\mathbb{R}^{M\times N}$ is an undersampling matrix selecting $M$ rows from the $N\times N$ identity matrix according to the binary sampling pattern $\mathbf{m}\!\in\!\{0,1\}^N$ and $\mathcal{F}\!\in\!\mathbb{C}^{N\!\times\! N}$ denotes the discrete Fourier transform. 
\begin{equation}\label{eq:pyxm}
\mathbf{y} = \mathbf{U_m} \mathcal{F} \mathbf{x} + \mathbf{n}.
\end{equation}
Here, $\mathbf{n}\in\mathbb{C}^M \sim \mathcal{N}_c(0, \sigma^2 \mathbf{I})$ represents additive complex white Gaussian noise~\cite{sijbers2004maximum} with a standard deviation of $\sigma$. For instance, when $N = 5$, if $\mathbf{m} = [1, 1, 0, 1, 0]$, $\mathbf{U}_{\mathbf{m}} \in\mathbb{R}^{3 \times 5}$ is formed by selecting the first, second, and fourth rows of the 5$\times$5 identity matrix. Without loss of generality, consider $\mathbf{x}$ and $\mathbf{y}$ are realizations from $X$ and $Y$ that represent the underlying distribution of MR images and their measurements, respectively. According to~\eqref{eq:pyxm}, the conditional distribution of $Y$ given $X=\mathbf{x}$ can be characterized as an isotropic Gaussian with mean $\mathbf{U_m} \mathcal{F} \mathbf{x}$ and variance $\sigma^2$, 
\emph{i.e.}, $p(\mathbf{y}|\mathbf{x})=\mathcal{N}_c(\mathbf{y}|\mathbf{U_m} \mathcal{F} \mathbf{x}, \sigma^2 \mathbf{I})$. Note that, in real-world clinical scenarios, noise may exhibit non-Gaussian characteristics, \emph{e.g.}, Rician, Poisson or mixed noise. We could adapt to other noise models by modifying the likelihood term $p(\mathbf{y}|\mathbf{x})$.

For task-adapted CS-MRI, finding an optimal sampling pattern $\mathbf{m}$ can be formulated as a specific instance of the linear feature design problem, \emph{i.e.}, optimizing the linear projection $\mathbf{\Phi}=\mathbf{U_m} \mathcal{F}$ by maximizing the ITM objective in~\eqref{eq:itm}.
Given that the discrete Fourier transform remains fixed, the focus shifts to optimizing $\mathbf{m}$. In CS-MRI, $\mathbf{m}$ satisfies that $\|\mathbf{m}\|_0=M$ and some other physical constraints $\mathcal{C}$ imposed by the real-world MR scanners.
From~\eqref{eq:itm}, we formulate the ITM-based objective function for task-adapted CS-MRI as
\begin{equation}\label{eq:obj}
\max_{\mathbf{m}}\left\{ I(Y; T)-\beta I(Y; X)\right\}, \quad 
\text{s.t.}\ \mathbf{m}\in\mathcal{M},
\end{equation}
where $\mathcal{M} = \left\{ \mathbf{m}: \mathbf{m}\in\{0,1\}^N, \lVert \mathbf{m} \rVert_0 = M, M<N \right\} \cap \mathcal{C}$ represents the sampling pattern constraint.

It is intractable to compute the mutual information between high-dimensional random variables for optimizing~\eqref{eq:obj}. As an alternative, we propose tractable proxy objectives based on variational inference for the case that $\beta< 0$ and marginal approximation for the case that $\beta \geq 0$. 

\subsubsection{$\beta < 0$}
We first establish a tractable lower bound for~\eqref{eq:obj}
using variational inference in Proposition~\ref{prop:ba-bound} for $\beta < 0$.
\begin{proposition}\upshape\label{prop:ba-bound}
Let $\mathbf{t}$ and $\mathbf{y}$ be the realizations of $T$ and $Y$, respectively. Given a variational approximation $q(\mathbf{t|y})$ for the posterior $p(\mathbf{t|y})$, the mutual information $I(Y;T)$ satisfies that
\begin{align}
I(Y;T)=&\mathbb{E}_{p(\mathbf{y})}[D_{KL}(p(\mathbf{t|y})\|q(\mathbf{t|y}))] \nonumber\\
&\qquad +H(T)+\mathbb{E}_{p(\mathbf{t,y})}\log{q(\mathbf{t|y})},
\end{align}
where $D_{KL}$ denotes the Kullback-Leibler~(KL) divergence and $H(T)$ is the entropy of $T$. Furthermore, the Barber-Agakov 
lower bound~\cite{barber2004algorithm} of $I(Y;T)$ is 
\begin{equation}\label{eq:lower_bound1}
I(Y;T)\geq H(T)+ \mathbb{E}_{p(\mathbf{t,y})} \log q(\mathbf{t|y}).
\end{equation}

Let $\mathbf{x}$ be the realization of $X$. The variational lower bound of mutual information $I(Y;X)$ is similarly obtained by 
\begin{equation}\label{eq:lower_bound2}
I(Y;X)\geq H(X)+ \mathbb{E}_{p(\mathbf{x,y})} \log q(\mathbf{x|y}).
\end{equation}
\end{proposition}

Proposition~\ref{prop:ba-bound} suggests that the Barber-Agakov lower bounds \eqref{eq:lower_bound1} and \eqref{eq:lower_bound2} are tight, only when $q(\mathbf{t} | \mathbf{y}) = p(\mathbf{t} | \mathbf{y})$ and $q(\mathbf{x} | \mathbf{y}) = p(\mathbf{x} | \mathbf{y})$ for all possible $\mathbf{y}$. In this case, maximizing the ITM-based objective function in~\eqref{eq:obj} is approximately equivalent to maximizing $\mathbb{E}_{p(\mathbf{x,t,y})} [\log q(\mathbf{t} | \mathbf{y}) - \beta \log q(\mathbf{x} | \mathbf{y})]$. This fact implies that $q$ should approximate $p$ to optimize $\mathbf{m}$. Note that the data entropy terms $H(T)$ and $H(X)$ are omitted in optimizing $\mathbf{m}$, since they do not depend on $\mathbf{m}$. To this end, we develop the proxy objective for maximizing~\eqref{eq:obj} as
\begin{equation}\label{eq:lbo_beta_leq0}
\max_{\mathbf{m},q} \mathbb{E}_{p(\mathbf{x,t,y})} [\log q(\mathbf{t} | \mathbf{y}) - \beta \log q(\mathbf{x} | \mathbf{y})], \quad \text{s.t.} \ \mathbf{m} \in \mathcal{M}.
\end{equation}

According to~\eqref{eq:lbo_beta_leq0}, sampling pattern $\mathbf{m}$, task inference $q(\mathbf{t} | \mathbf{y})$, and MR image reconstruction $q(\mathbf{x} | \mathbf{y})$ can be optimized in an end-to-end fashion. The sampling pattern is adapted to the downstream task and signal recovery for enhanced overall performance. In Section~\ref{sec: beta<0}, we provide a detailed explanation of the procedure for solving \eqref{eq:lbo_beta_leq0}.

\subsubsection{$\beta \geq 0$}
\label{sec: marginal entropy minimization}
The objective function in~\eqref{eq:obj} cannot be maximized using the lower bound of $I(Y;X)$ for $\beta \geq 0$. Since $I(Y; X) = H(Y) - H(Y|X)$ and $H(Y|X)$ remains fixed for all $\mathbf{m} \in \mathcal{M}$, the objective in~\eqref{eq:obj} is reformulated as
\begin{equation}\label{eq:entropy-reg}
\max_{\mathbf{m}}\left\{I(Y; T)-\beta H(Y)\right\}, \quad 
\text{s.t.} \ \mathbf{m}\in\mathcal{M}.
\end{equation}
In~\eqref{eq:entropy-reg}, the regularization term $-\beta H(Y)$ constrains the marginal entropy of $Y$. We develop a tractable proxy objective for~\eqref{eq:entropy-reg} by introducing variational approximation $q(\mathbf{t|y})$ for $p(\mathbf{t|y})$ and marginal entropy estimator $\hat{H}(Y)$ for $H(Y)$.
\begin{equation}\label{eq:obj-beta-gt-0}
\max_{\mathbf{m},q}\left\{\mathbb{E}_{p(\mathbf{x,t,y})} [\log q(\mathbf{t|y})-\beta\hat{H}(Y)] \right\}, \quad
\text{s.t.} \ \mathbf{m} \in \mathcal{M}.
\end{equation}
We calculate $\hat{H}(Y)$ and solve~\eqref{eq:obj-beta-gt-0} in Section~\ref{sec:beta>0}.

\subsection{Adaptive Sampling via Amortized Optimization}\label{sec:amortize}

Both the optimization problems \eqref{eq:lbo_beta_leq0} for $\beta< 0$ and \eqref{eq:obj-beta-gt-0} for $\beta\geq 0$ are formulated to obtain a single sampling pattern $\mathbf{m}$ for a specified sampling pattern constraint $\mathcal{M}$, which corresponds to a fixed sampling ratio $r$. This necessitates training separate models for different sampling ratios.
In this section, we further consider solving multiple problem instances of~\eqref{eq:obj} characterized by varying sampling pattern constraint $\mathcal{M}$ to accommodate to arbitrary sampling ratios via a single end-to-end trained model, \emph{i.e.}, achieving adaptive sampling. Note that it differs from \textit{scalable sampling}, which emphasizes progressive sampling and signal reconstruction.

To address the computational complexity arising from potentially infinite sampling ratios, we employ \textit{amortized optimization}~\cite{amos2023tutorial} to efficiently produce a stochastic sampling strategy that maps $r$ to a distribution of $\mathbf{m}$, supported on the sampling pattern constraint $\mathcal{M}$. The strategy is implemented by a \textit{pattern generation network} (PGN) with learnable parameters $\theta$.  
Given a distribution $p(r)$ over $r$, the objective for optimizing the PGN is 
\begin{equation}\label{eq:amortized-obj}
\!\max_{\theta} \mathbb{E}_{r\sim p(r)} \mathbb{E}_{\mathcal{M}\sim p(\mathcal{M}|r)}\mathbb{E}_{\pi_{\theta}(\mathbf{m}|\mathcal{M})}\! \left[I(Y\!;\!T)\!-\!\beta I(Y\!;\! X) \right]\!,
\end{equation}
where $\pi_{\theta}(\mathbf{m}|\mathcal{M})$ denotes the parametric conditional probability that generates $\mathbf{m}$ given $\mathcal{M}$. 
We then substitute the intractable quantity $I(Y;T) - \beta I(Y; X)$ in~\eqref{eq:amortized-obj} with tractable variational bounds obtained in Section~\ref{sec:Variational_Information_Optimization} to enable optimization. Define $\{r, \mathcal{M}, \mathbf{m}, \mathbf{x}, \mathbf{t}, \mathbf{y}\} \sim p(r)p(\mathcal{M}|r) \pi_{\theta}(\mathbf{m}|\mathcal{M}) p_{data}(\mathbf{x,t}) p(\mathbf{y|x})$. When $\beta < 0$, we introduce~\eqref{eq:lbo_beta_leq0} and obtain:
\begin{equation}\label{eq:amt1}
\max_{\theta,q} \mathbb{E}_{r, \mathcal{M}, \mathbf{m}, \mathbf{x}, \mathbf{t}, \mathbf{y}}[\log q(\mathbf{t|y}) - \beta \log q(\mathbf{x|y})],
\end{equation}
and for $\beta \geq 0$, we consider~\eqref{eq:obj-beta-gt-0} such that:
\begin{equation}\label{eq:amt2}
\max_{\theta,q} \mathbb{E}_{r, \mathcal{M}, \mathbf{m,x,t,y}} [\log q(\mathbf{t|y}) - \beta \hat{H}(Y)].
\end{equation}

\begin{figure}[!t]
\centering
\includegraphics[width=0.9\textwidth]{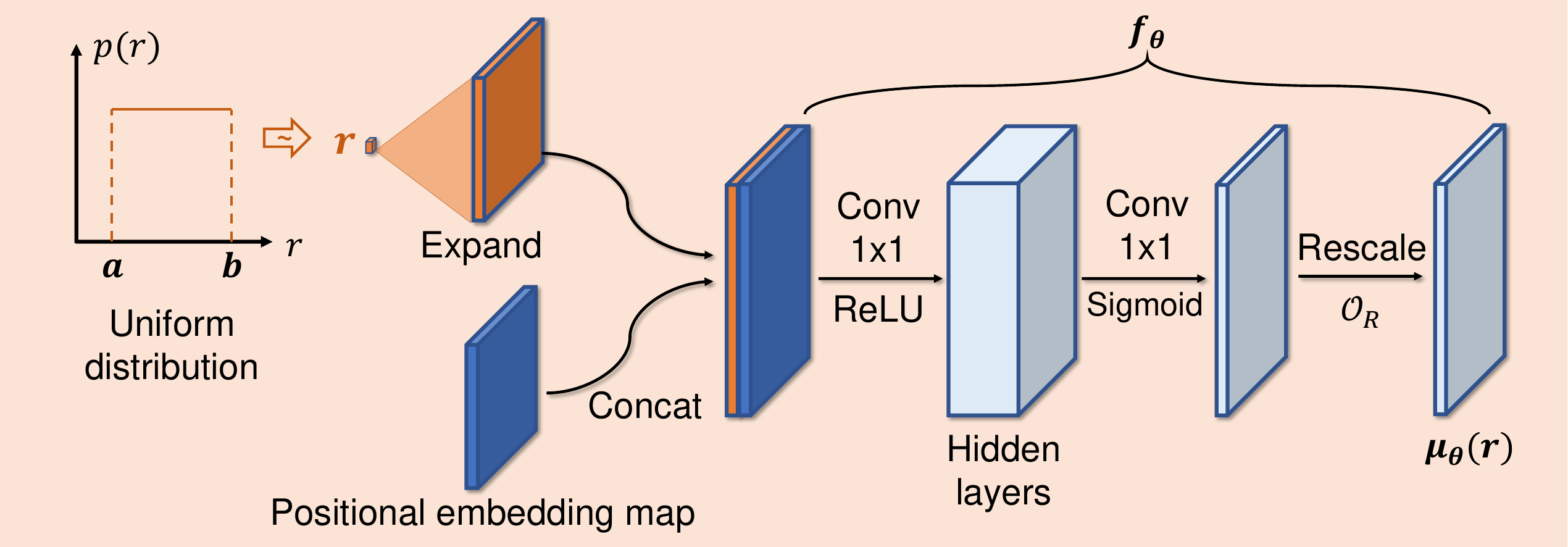}
    \caption{Illustration of the parametric mapping $r \mapsto \mu_{\theta}(r)$. }
    \label{fig:probnet}
\end{figure}

We then describe the implementation of $p(r)p(\mathcal{M}|r) \pi_{\theta}(\mathbf{m}|\mathcal{M})$. Given that the constraint $\mathcal{M}$ on the sampling pattern $\mathbf{m}$ is considered for any sampling ratio $r \in [0,1]$, we define it as follows:
\begin{equation}\label{eq:constraint}
\mathcal{M}=\{\mathbf{m}\in \{0,1\}^N : |\lVert \mathbf{m} \rVert_0 - rN| < \epsilon \},
\end{equation}
where $\epsilon$ is a tolerance hyperparameter to ensure that the sampling pattern is consistent with the sampling ratio. This implies that the unknown distribution $p(\mathcal{M}|r)$ can alternatively be achieved using a specified distribution $p(r)$ over the sampling rate $r$ according to \eqref{eq:constraint}. In this paper, we set $p(r)$ to be a uniform distribution $\mathcal{U}[a,b]$ with $0\leq a\leq b \leq 1$\footnote{The uniform distribution corresponds to a scenario with no preference for sampling rates within the specified range $[a,b]$. However, in cases where there is a preference for a specific range of sampling rates, alternative distributions may be considered.}. 
Once $\mathcal{M}$ is produced from $p(r)p(\mathcal{M}|r)$, we generate the sampling pattern $\mathbf{m}$ using the proposed PGN, i.e., $\mathbf{m}\sim\pi_{\theta}(\mathbf{m}|\mathcal{M})$, and then decide to accept $\mathbf{m}$ as a sample or reject it and return to the sampling step, depending on whether $\mathbf{m}$ belongs to $\mathcal{M}$. This technique is known as \textit{rejection sampling}~\cite{forsythe1972neumann}. If a sample of $\mathbf{m}$ is accepted, it is used to generate the corresponding measurements, obtain the final outputs, and optimize the whole MRI model for downstream tasks and signal recovery.

We now return to the proposed PGN, which implements a stochastic sampling strategy to generate the sampling pattern $\mathbf{m}$ in a differentiable manner, using input $r$ sampled from $\mathcal{U}[a,b]$. We first define the distribution $\pi_{\theta}(\mathbf{m}|\mathcal{M})$ over $\mathbf{m}$ by treating each element $m_i$ of $\mathbf{m}$ as an independent Bernoulli random variable, as follows:
\begin{equation}
    \pi_{\theta}(\mathbf{m}|\mathcal{M}) \in \left\{ \prod_{i=1}^N \mathcal{B}(m_i|\mu_i) : \mu_i \in [0,1] \ \text{for all } i \right\}, 
\end{equation}
where the subscript $i$ indexes the k-space position, $\mu_i$ represents the probability of sampling the point at position $i$, and $\mathcal{B}(m_i|\mu_i)$ denotes that $p(m_i=1)=\mu_i$ and $p(m_i=0)=1-\mu_i$. 
To optimize the sampling strategy in a differentiable manner, we replace $\mu_i$ with a parametric mapping $r \mapsto \mu_{\theta}(r)$ that maps the sampling ratio $r$ to a set of valid probabilities $\mu_{\theta}(r)$, where each element $\mu_{\theta}(r)_i$ lies in $[0,1]$. 
We define a proposal distribution $\tilde{\pi}_{\theta}(\mathbf{m}|\mathcal{M})$ for $\pi_{\theta}(\mathbf{m}|\mathcal{M})$ given $r$ as
\begin{equation} \label{eq:pi(m|M)}
\tilde{\pi}_{\theta}(\mathbf{m}|\mathcal{M}) = \prod_{i=1}^N \mathcal{B}(m_i|\mu_{\theta}(r)_i).
\end{equation}
Thus, $\pi_{\theta}(\mathbf{m}|\mathcal{M})$ is determined by accepting samples $\mathbf{m}$ from $\tilde{\pi}_{\theta}(\mathbf{m}|\mathcal{M})$ when $\mathbf{m} \in \mathcal{M}$, or rejecting them and re-implementing \eqref{eq:pi(m|M)} otherwise.

The parametric mapping $r\mapsto\mu_{\theta}(r)$ is defined as
\begin{equation}\label{eq:PGN}
\mu_{\theta}(r) = f_{\theta}[\text{Concat}(r \cdot \mathbbm{1}_N, M_{pe})],
\end{equation}
where \(M_{pe} \in \mathbb{R}^N\) is a learnable matrix corresponding to position embedding, \(\mathbbm{1}_N\) denotes an all-one vector of dimension \(N\), and \(f_\theta\) is a simple neural network whose input is the concatenation of \(r \cdot \mathbbm{1}_N\) and \(M_{pe}\). In this paper, we set \(f_\theta\) as a multilayer perceptron (MLP) with one hidden layer, followed by a sigmoid activation and a rescale operator proposed in \cite{RN211}, as illustrated in Fig.~\ref{fig:probnet}. 
Note that the linear layers of the MLP are implemented using equivalent one-by-one convolutions.
Given arbitrary \(\mathbf{b} \in \mathbb{R}^N\), the rescale operator \(\mathcal{O}_R\) is defined as
\begin{equation}\label{eq:rescale}
\mathcal{O}_R(\mathbf{b}) =
\begin{cases}
\frac{rN}{\|\mathbf{b}\|_1} \mathbf{b},  & \text{if } \|\mathbf{b}\|_1 \geq r \\
1 - \frac{N - rN}{N - \|\mathbf{b}\|_1}(1 - \mathbf{b}), & \text{otherwise}
\end{cases}.
\end{equation}
From~\eqref{eq:rescale}, \(\mathcal{O}_R(\mathbf{b}) \in [0,1]^N\) and \(\|\mathcal{O}_R(\mathbf{b})\|_1 = rN = M\), thereby guaranteeing that \(\mathbf{m} \in \mathcal{M}\) with high probability. 

We illustrate the training steps with respect to sampling in Algorithm~\ref{alg:amo}. Once the model is well-trained, one can generate $\mathbf{m}$ according to steps 3-10 in Algorithm~\ref{alg:amo} for a specified $r$, achieving adaptive sampling via a single model.

\begin{algorithm}[!ht] 
\renewcommand{\baselinestretch}{1.0}
\renewcommand{\arraystretch}{1.0}
\caption{Adaptive Sampling via Amortized Optimization}\label{alg:amo}
\begin{algorithmic}[1]
\renewcommand{\algorithmicrequire}{\textbf{Input:}}
\renewcommand{\algorithmicensure}{\textbf{Output:}}
\REQUIRE{Tolerance hyperparameter $\epsilon$, MRI data dimension $N$, PGN with parameter $\theta$, constants $a,b$ with $0\!\leq\!a\!\leq\! b \!\leq\!1$}
\WHILE{training the CS-MRI model}
\STATE Sample $r$ from the uniform distribution $r\sim \mathcal{U}[a,b]$
\STATE Determine $\mathcal{M}$ according to \eqref{eq:constraint} 
\STATE Obtain $\mu_\theta(r)$ through PGN according to \eqref{eq:PGN}
\STATE Generate $\mathbf{m}$ through \eqref{eq:pi(m|M)} 
\IF{$\mathbf{m\not\in\mathcal{M}}$}
\STATE Reject $\mathbf{m}$, re-implement \eqref{eq:pi(m|M)}, and return to step 5
\ELSE[$\mathbf{m\in\mathcal{M}}$]
\STATE Accept $\mathbf{m}$ as a sample
\ENDIF
\STATE Use $\mathbf{m}$ to generate measurements, obtain final outputs, and optimize the whole MRI model (including PGN)
\ENDWHILE
\end{algorithmic} 
\end{algorithm}

\subsection{Tractable Optimization With Variational Bounds ($\beta<0$)} \label{sec: beta<0}

In this section, we solve \eqref{eq:amt1} in the case that $\beta < 0$ for joint optimization of downstream tasks and MRI reconstruction. We optimize the tractable variational lower bounds for \eqref{eq:amt1} given the sampling pattern $\mathbf{m}$ obtained through Algorithm~\ref{alg:amo}.

To solve \eqref{eq:amt1}, it is essential to construct tractable formulations of $q(\mathbf{t|y})$ and $q(\mathbf{x|y})$, balancing computational efficiency with fidelity to the true posteriors $p(\mathbf{t|y})$ and $p(\mathbf{x|y})$. 
Note that there exists statistical correlation between elements of $\mathbf{t}$ conditioned on $\mathbf{y}$ for several downstream tasks, such as segmentation. To capture this dependency, we introduce an auxiliary latent random variable $Z$ with a prior distribution $q(\mathbf{z|y})$ and a conditional distribution $q(\mathbf{t|z,y})$, enabling the reformulation of $q(\mathbf{t|y})$ through marginalization as follows:
\begin{equation}\label{eq:likelihood-t}
q(\mathbf{t|y}) = \int q(\mathbf{t|z,y}) q(\mathbf{z|y}) \, \mathrm{d}\mathbf{z}.
\end{equation}
In this section, we use \textit{Segmentation}-adapted CS-MRI as an example to elaborate on the optimization steps with an auxiliary latent variable for this task\footnote{We demonstrate in Section~\ref{sec:seg-performance} that omitting the correlation between elements, in the absence of an auxiliary latent variable, significantly degrades segmentation performance.}. 
While statistical correlations also exist between elements of the MRI reconstruction $\mathbf{x}$ conditioned on $\mathbf{y}$, we model $q(\mathbf{x|y})$ as a diagonal Gaussian distribution in this paper. 
This approach is chosen to balance computational efficiency with accuracy, as maximizing $\log q(\mathbf{x|y})$ is equivalent to minimizing the mean square error (MSE) between the signal $\mathbf{x}$ and a transformation of $\mathbf{y}$ (see more details in Section~\ref{sec:relation}), a typical objective in signal reconstruction tasks.

Directly evaluating the log-likelihood of $q(\mathbf{t|y})$ in \eqref{eq:likelihood-t} remains generally intractable. To address this, we construct the Evidence Lower Bound (ELBO, \cite{kingma2013auto}) for $\log q(\mathbf{t|y})$ by introducing a variational approximation $\hat{q}(\mathbf{z|t,y})$ for the intractable latent posterior $q(\mathbf{z|t,y}) \propto q(\mathbf{t|z,y}) q(\mathbf{z|y})$:\footnote{We omit the derivation of \eqref{eq:elbo} as it follows standard variational inference principles; see \cite{kingma2013auto} for a detailed explanation.}
\begin{align}\label{eq:elbo}
\log q(\mathbf{t|y}) \geq \text{ELBO} := 
\mathbb{E}_{\hat{q}(\mathbf{z|t,y})} [\log q(\mathbf{t}|\mathbf{z}, \mathbf{y}) - D_{KL}[\hat{q}(\mathbf{z|t,y})||q(\mathbf{z|y})]].
\end{align}
By substituting $\log q(\mathbf{t|y})$ in \eqref{eq:amt1} with its ELBO from \eqref{eq:elbo}, we obtain a tractable variational lower bound for \eqref{eq:amt1}:
\begin{align}\label{eq:naive-obj}
\max_{\theta,\phi} \mathbb{E}_{r, \mathcal{M}, \mathbf{m}, \mathbf{x}, \mathbf{t}, \mathbf{y}} \mathbb{E}_{\hat{q}(\mathbf{z|t,y})} [\log q(\mathbf{t}|\mathbf{z}, \mathbf{y})
- D_{KL}[\hat{q}(\mathbf{z|t,y})||q(\mathbf{z|y})] - \beta \log q(\mathbf{x|y})],
\end{align}
where $D_{KL}$ represents the Kullback–Leibler (KL) divergence. 

Notably, \eqref{eq:naive-obj} offers intuitive interpretations. The first term, $\log q(\mathbf{t|z,y})$, aims to reconstruct the ground truth $\mathbf{t}$ sampled from training data given the measurement $\mathbf{y}$ and the latent variable $\mathbf{z}$ generated from $\hat{q}(\mathbf{z|t,y})$. Since $\hat{q}(\mathbf{z|t,y})$ is conditioned on the ground truth $\mathbf{t}$, an ideal reconstruction of $\mathbf{t}$ can be achieved.
The second term, $D_{KL}[\hat{q}(\mathbf{z|t,y})||q(\mathbf{z|y})]$, encourages $q(\mathbf{z|y})$ to generate latent variables $\mathbf{z}$ that resemble those produced by $\hat{q}(\mathbf{z|t,y})$. Here, $\hat{q}(\mathbf{z|t,y})$ acts as a ``teacher'' distribution that leverages the information of $\mathbf{t}$, while $q(\mathbf{z|y})$ serves as a ``student'' distribution, lacking information on $\mathbf{t}$ and attempting to approximate $\hat{q}(\mathbf{z|t,y})$ ``in an average sense''. 
We refer to this approximation as the \textit{average sense} because the optimal solution of $q(\mathbf{z|y})$ that minimizes $\mathbb{E}_{r, \mathcal{M}, \mathbf{m}, \mathbf{x}, \mathbf{t}, \mathbf{y}} \mathbb{E}_{\hat{q}(\mathbf{z|t,y})} D_{KL}[\hat{q}(\mathbf{z|t,y})||q(\mathbf{z|y})]$ is $\mathbb{E}_{p(\mathbf{t|y})}[\hat{q}(\mathbf{z|t,y})]$, referred to as the marginal or aggregated latent posterior~\cite{tomczak2018vae}. 
The third term, $\log q(\mathbf{x|y})$, seeks to reconstruct the MRI signal $\mathbf{x}$ based on the measurement $\mathbf{y}$, thereby ensuring data fidelity.

Starting from here, we present the solving steps for \eqref{eq:naive-obj}.
As mentioned in Section~\ref{sec:background}, the sampling pattern $\mathbf{m}$ influences the measurement $\mathbf{y}$, implying that the posterior distributions $q(\mathbf{t|y})$ and $q(\mathbf{x|y})$ are dependent on $\mathbf{m}$. Therefore, it is necessary to condition these distributions on $\mathbf{m}$, meaning that the input for the task-inference network and signal-reconstruction network should contain information about both $\mathbf{y}$ and $\mathbf{m}$. 
Due to the mismatched dimensionality and different data types of $\mathbf{y}$ (continuous) and $\mathbf{m}$ (discrete), it is undesirable to directly use the paired data $(\mathbf{y}, \mathbf{m})$ to optimize the MRI models. 
To address this, we use the zero-filling reconstruction $\hat{\mathbf{x}}_{zf} := \mathcal{F}^H \mathbf{U_m}^T \mathbf{y} \in \mathbb{C}^N$ as a representation of $(\mathbf{y}, \mathbf{m})$, as it contains all the information about $(\mathbf{y}, \mathbf{m})$. Specifically, $\mathbf{y}$ can be recovered from $\hat{\mathbf{x}}_{zf}$ by $\mathbf{U_m} \mathcal{F}\hat{\mathbf{x}}_{zf}$, and $\mathbf{m}$ can be recovered by identifying the zero elements in $\mathcal{F} \hat{\mathbf{x}}_{zf}$.

To optimize \eqref{eq:naive-obj}, we implement probabilistic modeling for $q(\mathbf{t}|\mathbf{z}, \mathbf{y})$, $q(\mathbf{z|y})$, $\hat{q}(\mathbf{z|t,y})$, and $q(\mathbf{x|y})$ with the latent variable model (LVM) \cite{kingma2013auto}, a powerful off-the-shelf probabilistic model that can be trained by maximum likelihood. Since the input for the task-inference network and signal-reconstruction network is $\hat{\mathbf{x}}_{zf} := \mathcal{F}^H \mathbf{U_m}^T \mathbf{y}$, we first encode the information of $\hat{\mathbf{x}}_{zf}$ into the feature $\hat{\mathbf{y}}$ by a \textit{measurement encoder} $\mathcal{E}_{\phi^Y}$:
\begin{equation}\label{eq:y_hat}
\hat{\mathbf{y}} = \mathcal{E}_{\phi^Y}(\mathcal{F}^H \mathbf{U_m}^T \mathbf{y}).
\end{equation}
$\hat{\mathbf{y}}$ is then fed into the \textit{latent decoder} $\mathcal{D}_{\phi^{Z}}$, \textit{task-inference decoder} $\mathcal{D}_{\phi^{T}}$, and \textit{signal-reconstruction decoder} $\mathcal{D}_{\phi^{X}}$ to characterize $q(\mathbf{z|y})$, $q(\mathbf{t}|\mathbf{z}, \mathbf{y})$, and $q(\mathbf{x|y})$, respectively. 
A typical choice of variational family to formulate these conditional probability distribution is the mean-field variational family, where the elements of a variable are assumed to be mutually independent~\cite{blei2017variational}. 
$q(\mathbf{x|y})$ is modeled as a diagonal Gaussian distribution whose mean and variance are produced by $\mathcal{D}_{\phi^{X}}$:
\begin{equation} \label{eq:q(x|y)}
q(\mathbf{x|y}) = \mathcal{N}(\mathbf{x}|\mathcal{D}_{\phi^{X}}(\hat{\mathbf{y}})).
\end{equation}
$q(\mathbf{z|y})$ is also modeled as a diagonal Gaussian distribution with learnable mean and variance that is commonly adopted for modeling the conditional distribution of latent variables.
\begin{equation}\label{eq:q(z|y)}
q(\mathbf{z|y}) = \mathcal{N}(\mathbf{z}|\mathcal{D}_{\phi^{Z}}(\hat{\mathbf{y}})).
\end{equation}
$q(\mathbf{t}|\mathbf{z}, \mathbf{y})$ is formulated associated with the tasks. Taking MRI segmentation task as an example, $q(\mathbf{t|z,y})$ is modeled as an independent Categorical distribution for each pixel, whose class probability is produced by $\mathcal{D}_{\phi^{T}}(\mathbf{z},\hat{\mathbf{y}})$.
\begin{equation}\label{eq:q(t|zy)}
q(\mathbf{t|z,y}) = \prod_i \mathrm{Categorical}(t_i|\mathcal{D}_{\phi^{T}}(\mathbf{z},\hat{\mathbf{y}})_i).
\end{equation}
As the KL divergence of $\hat{q}(\mathbf{z|t,y})$ aims to match the prior distribution $q(\mathbf{z|y})$, we also formulate $\hat{q}(\mathbf{z|t,y})$ as a diagonal Gaussian distribution, using the same decoder $\mathcal{D}_{\phi^{Z}}$ as $q(\mathbf{z|y})$ and another \textit{measurement encoder} $\mathcal{E}_{\Tilde{\phi}^Y}$:
\begin{equation}\label{eq:qhat(z|t,y)}
\hat{q}(\mathbf{z|t,y}) = \mathcal{N}(\mathbf{z}|\mathcal{D}_{\phi^{Z}}(\mathcal{E}_{\Tilde{\phi}^Y}(\mathbf{t}, \mathcal{F}^H \mathbf{U_m}^T \mathbf{y}))),
\end{equation}
where $\mathcal{E}_{\Tilde{\phi}^Y}$ shares the same architecture to $\mathcal{E}_{\phi^Y}$ except the input has an additional channel for the ground truth segmentation map $\mathbf{t}$.  
Note that \eqref{eq:q(z|y)} and \eqref{eq:q(t|zy)} can be performed multiple times to generate different conceivable segmentation maps that are consistent with the posterior samples from $p(\mathbf{t|y})$, thereby naturally addressing the uncertainty problem imposed by the MR subsampling discussed in Section~\ref{sec:taskadaptedmri}. 

\begin{figure*}[!t]
\centering \includegraphics[width=0.95\textwidth]{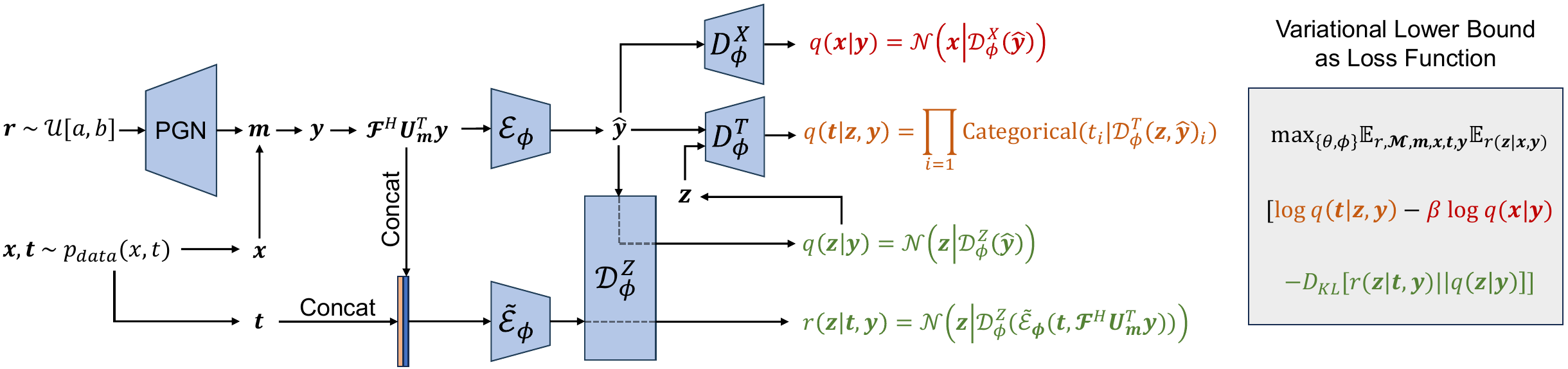}
\caption{Illustration of Algorithm~\ref{alg:optimize} for optimizing \eqref{eq:naive-obj}. At each training iteration, a pair of data $(\mathbf{x}, \mathbf{t})$ is sampled from the training set, along with a sampling ratio $r$ drawn from $\mathcal{U}[a,b]$. Using $r$ as the input, the PGN generates a sampling pattern $\mathbf{m}$, which is then used to create the measurements $\mathbf{y}$. The information in $\mathbf{y}$ is encoded into the feature $\hat{\mathbf{y}}$ by passing the zero-filled reconstruction $\mathcal{F}^H \mathbf{U_m}^T \mathbf{y}$ through the measurement encoder $\mathcal{E}_{\phi^Y}$. Subsequently, $\hat{\mathbf{y}}$ is fed into the decoders $\mathcal{D}_{\phi^{X}}$, $\mathcal{D}_{\phi^{T}}$, and $\mathcal{D}_{\phi^{Z}}$ to calculate $q(\mathbf{x|y})$, $q(\mathbf{t|z,y})$, $q(\mathbf{z|y})$, and $\hat{q}(\mathbf{z|t,y})$. The variational lower bound (loss function) is then evaluated, and the entire MRI model comprising the PGN, encoders $\mathcal{E}_{\phi^Y}$, $\mathcal{E}_{\Tilde{\phi}^Y}$, and decoders $\mathcal{D}_{\phi^{X}}$, $\mathcal{D}_{\phi^{T}}$, $\mathcal{D}_{\phi^{Z}}$ is optimized through backpropagation.
}\label{fig:train}
\end{figure*}

The training steps for optimizing \eqref{eq:naive-obj} are summarized in Algorithm~\ref{alg:optimize} and illustrated in \figurename~\ref{fig:train}. The trained model is then used to obtain the MRI reconstruction and a series of segmentation maps according to Algorithm~\ref{alg:inference}. 

\begin{algorithm}[!ht] 
\renewcommand{\baselinestretch}{1.0}
\renewcommand{\arraystretch}{1.0}
\caption{Optimizing \eqref{eq:naive-obj} for \textit{Segmentation}-Adapted CS-MRI}\label{alg:optimize}
\begin{algorithmic}[1]
\renewcommand{\algorithmicrequire}{\textbf{Input:}}
\renewcommand{\algorithmicensure}{\textbf{Output:}}
\REQUIRE{Initialized encoders and decoders: measurement encoders $\mathcal{E}_{\phi^Y}$ and $\mathcal{E}_{\Tilde{\phi}^Y}$, latent decoder $\mathcal{D}_{\phi^{Z}}$, task-inference decoder $\mathcal{D}_{\phi^{T}}$, signal-reconstruction decoder $\mathcal{D}_{\phi^{X}}$.}
\WHILE{$\phi,\theta$ not converge}
\STATE Sample $\mathbf{x},\mathbf{t}$ from the training set: $\mathbf{x},\mathbf{t}\sim p_{data}(\mathbf{x,t})$ 
\STATE Sample $\mathbf{m}$ according to Algorithm~\ref{alg:amo}
\STATE Obtain $\mathbf{y}$ and $\hat{\mathbf{x}}_{zf}= \mathcal{F}^H \mathbf{U_m}^T \mathbf{y} \in \mathbb{C}^N$
\STATE Encode $\hat{\mathbf{x}}_{zf}$ in the feature $\hat{\mathbf{y}}$ following \eqref{eq:y_hat}
\STATE Sample $\mathbf{z}\sim \hat{q}(\mathbf{z|t, y})$ following 
\eqref{eq:qhat(z|t,y)}
\STATE Evaluate $\log q(\mathbf{x|y})$ following \eqref{eq:q(x|y)}
\STATE Evaluate $\log q(\mathbf{t|z,y})$ following \eqref{eq:q(t|zy)}
\STATE Evaluate $D_{KL}[\hat{q}(\mathbf{z|t, y})\|q(\mathbf{z|y})]$ following \eqref{eq:q(z|y)}, \eqref{eq:qhat(z|t,y)}
\STATE Do back-propagation of \eqref{eq:naive-obj} to compute gradient with respect to $\phi$ and $\theta$
\ENDWHILE
\end{algorithmic} 
\end{algorithm}
\vspace{-12pt}
\begin{algorithm}[!ht] 
\renewcommand{\baselinestretch}{1.0}
\renewcommand{\arraystretch}{1.0}
\caption{Inference for \textit{Segmentation}-Adapted CS-MRI}\label{alg:inference}
\begin{algorithmic}[1]
\renewcommand{\algorithmicrequire}{\textbf{Input:}}
\renewcommand{\algorithmicensure}{\textbf{Output:}}
\REQUIRE{Measurement encoder $\mathcal{E}_{\phi^Y}$, latent decoder $\mathcal{D}_{\phi^{Z}}$, segmentation decoder $\mathcal{D}_{\phi^{T}}$, image decoder $\mathcal{D}_{\phi^{X}}$, measurement $\mathbf{y}$, sampling pattern $\mathbf{m}$.}
\STATE Obtain $\mathbf{y}$ and $\hat{\mathbf{x}}_{zf}= \mathcal{F}^H \mathbf{U_m}^T \mathbf{y} \in \mathbb{C}^N$
\STATE Encode $\hat{\mathbf{x}}_{zf}$ in the feature $\hat{\mathbf{y}}$ following \eqref{eq:y_hat}
\STATE Obtain posterior approximation $q(\mathbf{x|y})$ following \eqref{eq:q(x|y)}
\WHILE{obtaining sufficient segmentation maps}
\STATE Sample $\mathbf{z}\sim q(\mathbf{z|y})$ following 
\eqref{eq:q(z|y)}
\STATE Sample $\mathbf{t}\sim q(\mathbf{t|z,y})$ following 
\eqref{eq:q(t|zy)}
\ENDWHILE
\ENSURE{$q(\mathbf{x|y})$ and a series of segmentation maps $\mathbf{t}$s. The mean of $q(\mathbf{x|y})$ can be taken as a specific MRI reconstruction image.}
\end{algorithmic} 
\end{algorithm}

\subsection{Tractable Optimization With Marginal Entropy Minimization ($\beta\geq 0$)}
\label{sec:beta>0}

In this section, we consider the case that $\beta\geq 0$ in~\eqref{eq:amt2}, \emph{i.e.}, implementing downstream tasks without reconstructing the original image, including privacy preserving clinical decision making and compressed learning. We solve \eqref{eq:amt2} with $\beta\geq 0$ by maximizing the log-likelihood of $q(\mathbf{t|y})$ and minimizing the estimated marginal entropy $\hat{H}(Y)$. 

For tasks that consider statistical correlations between elements of $\mathbf{t}$ conditioned on $\mathbf{y}$, the log-likelihood of $q(\mathbf{t}|\mathbf{y})$ is maximized using the approach described in Section~\ref{sec: beta<0}, which involves introducing an auxiliary latent variable. However, tasks producing scalar outputs such as pathological classification do not require capturing statistical correlations and introducing an auxiliary latent variable. In this section, we use \textit{Classification}-adapted CS-MRI as an example to illustrate the steps for maximizing $\log q(\mathbf{t}|\mathbf{y})$. Specifically, $q(\mathbf{t}|\mathbf{y})$ can be directly modeled as a Categorical distribution, with class probabilities produced by $\mathcal{D}_{\phi^{T}}(\hat{\mathbf{y}})$:
\begin{equation}\label{eq:classification_qty}
q(\mathbf{t}|\mathbf{y}) = \mathrm{Categorical}\left(\mathbf{t} | 
 \mathcal{D}_{\phi^{T}}(\hat{\mathbf{y}})\right).
\end{equation}
Note that, different from \eqref{eq:q(t|zy)}, \eqref{eq:classification_qty} does not involve the pixel-wise product of the probability and the latent input $\mathbf{z}$ for $\mathcal{D}_{\phi^{T}}$.

Subsequently, we develop $\hat{H}(Y)$ to estimate $H(Y)$ with three steps, including i) approximating the marginals of the noisy k-space measurement $\mathbf{y}$, ii) modeling the statistical dependencies due to k-space conjugate symmetry, and iii) estimating the marginal entropy $H(Y)$.

\textbf{Step i): Approximating marginals of $\mathbf{y}$.} 
We first model the marginals of the underlying \textit{noiseless} k-space measurement at each position $k_i=(\mathcal{F}\mathbf{x})_i$, denoted by $q(k_i)$ for $i=1,\cdots,N$. Since the distribution over the \textit{noisy} k-space measurement $y_i$ given $k_i$ is a Gaussian $\mathcal{N}_c(y_i|k_i,\sigma^2)$ according to~\eqref{eq:pyxm}, we consider modeling $q(k_i)$ as Gaussian\footnote{Clearly, assuming $q(k_i)$ to be non-Gaussian introduces computational complexity or intractability, as Eq.~\eqref{eq:qy} no longer holds.}, which allows for the computationally tractable approximation $q(y_i)$ for the true marginal of $y_i$ through analytical Gaussian marginalization:
\begin{equation}\label{eq:qy}
q(y_i) = \int \mathcal{N}_c(y_i|k_i,\sigma^2) q(k_i) \mathrm{d}k_i.
\end{equation}

\textbf{Step ii): Modeling statistical dependencies for $q(\mathbf{y})$.} From~\eqref{eq:qy}, the marginal of the noisy k-space measurement $\mathbf{y}$ can be simply formulated as 
\begin{equation}\label{eq:simple_marginal}
q(\mathbf{y}) =\prod_{i:m_i=1}q(y_i),~i=1,2,...,N.
\end{equation}
However, $q(\mathbf{y})$ can be further analyzed due to the \textit{conjugate symmetry property} of k-space~\cite{defazio2019offset}, \emph{i.e.}, statistical dependencies between the sample position $i$ and its conjugate symmetry position $i*=N-i$ (also referred to as point-reflected position). 
Given $k_i$, we have the following relation between two measurements~\cite{defazio2019offset}:
\begin{align}
y_i &= k_i + n_i, \ \ n_i\sim \mathcal{N}_c(0, \sigma^2 I)\label{eq:yi}, \\
y_{i*} &= \bar{k}_i + n_{i*}, \ \ n_{i*}\sim \mathcal{N}_c(0, \sigma^2 I)\label{eq:yi*},
\end{align}
where $\bar{k}_i$ is the complex conjugate of $k_i$. Due to \eqref{eq:yi} and \eqref{eq:yi*}, sampling at position $i*$ is redundant when sampling at position $i$. Since the joint distribution of $y_i,y_{i*}$ given $k_i$ is the bivariate Gaussian $\mathcal{N}_c((y_i,y_{i*})^T|(k_i,\bar{k}_i)^T,\sigma^2\mathbf{I})$, we can approximate the joint marginal of $y_i,y_{i*}$ similarly to \eqref{eq:qy}:
\begin{equation}\label{eq:qyy}
q(y_i,y_{i*})=\int \mathcal{N}_c((y_i,y_{i*})^T|(k_i,\bar{k}_i)^T,\sigma^2\mathbf{I})q(k_i) \mathrm{d}k_i.
\end{equation}
Let $m_i$ denote the $i$-th element of current sampling pattern $\mathbf{m}$. Define $\mathcal{I}=\{i:m_i=1,m_{i*}=1\}$ as the set of positions where both the sample and its point-reflected position are included, and $\mathcal{J}=\{j:m_j=1,m_{j*}=0\}$ the set of positions where the sample is included but its point-reflected position is not.
We obtain from \eqref{eq:qy} and \eqref{eq:qyy} that
\begin{equation}\label{eq:marginal}
q(\mathbf{y})\!=\! q(y_1, y_2,\cdots,y_M) 
\!=\!\left[\prod_{i \in \mathcal{I}} q(y_i,y_{i*})\right]^\frac{1}{2}\!\prod_{j\in \mathcal{J}}q(y_j),
\end{equation}
where the square root occurs due to the same distributions are multiplied twice, \emph{e.g.}, $q(y_1,y_{N-1})=q(y_{N-1},y_{1})$. Compared to \eqref{eq:simple_marginal}, \eqref{eq:marginal} takes into account the statistical dependencies arising from the conjugate symmetry property of k-space, resulting in a more accurate estimation of $\hat{H}(Y)$.

\textbf{Step iii): Estimating marginal entropy $H(Y)$.} 
We estimate $H(Y)$ by computing the entropy $\hat{H}(Y)$ of $q(\mathbf{y})$. To this end, we first compute $q(y_j)$ and $q(y_i, y_{i^*})$ as in \eqref{eq:marginal} by applying Bayes’ theorem for Gaussian variables~\cite{bishop2006pattern}. Without introducing ambiguity, we define the distribution of a complex scalar random variable through the joint distribution of its real and imaginary parts, i.e., $q(y_j) = q(y_j^r, y_j^c)$, where the superscripts $r$ and $c$ represent the real and imaginary parts of complex variables, respectively. 

Suppose that the real and imaginary parts, $k_i^r$ and $k_i^c$, of the noiseless k-space measurement $k_i$ are independent for simplicity. Define the mean and variance of $k_i^r / k_i^c$ over the training data set $\mathcal{D}$ as $\mu_i^r / \mu_i^c$ and $V_i^r / V_i^c$, respectively, i.e., 
\begin{equation}
\label{eq:qk}
    q(k_i) = \mathcal{N}(k_i^r|\mu_i^r,V_i^r)\mathcal{N}(k_i^c|\mu_i^c,V_i^c).
\end{equation}
Note that $\mu_i^r, \mu_i^c, V_i^r, V_i^c$ are pre-computed before optimizing the MRI model.
Since $q(y_i|k_i) = \mathcal{N}(y_i^r|k_i^r, \sigma^2)\mathcal{N}(y_i^c|k_i^c, \sigma^2)$ according to \eqref{eq:yi}, we obtain that 
\begin{align}
q(y_i) &= \int q(k_i)q(y_i|k_i)\mathrm{d}(k_i)\nonumber\\
&= \mathcal{N}(y_i^r|\mu_i^r, \sigma^2 + V_i^r)\mathcal{N}(y_i^c|\mu_i^c, \sigma^2 + V_i^c). \label{eq:qyi1_form}
\end{align}
according to \eqref{eq:qy}. For the positions in $\mathcal{J}$, similar to the derivation of \eqref{eq:qyi1_form}, we can easily obtain $q(y_i, y_{i*})$ as follows:
\begin{align}\label{eq:qyi2_form}
&q(y_i,y_{i*}) = \mathcal{N}((y_i^r, y_{i*}^r)^T|(1, 1)^T \mu_i^r, \sigma^2 \mathbf{I} + (1, 1)^T V_i^r (1, 1))   \nonumber\\
&\quad\cdot\mathcal{N}((y_i^c, y_{i*}^c)^T|(1, -1)^T \mu_i^c, \sigma^2 \mathbf{I} + (1, 1)^T V_i^c (1, 1)).
\end{align}
From \eqref{eq:qyi1_form} and \eqref{eq:qyi2_form}, we obtain $\hat{H}(Y_i)$ and $\hat{H}(Y_i, Y_{i^*})$ according to the analytical form of entropy of a Gaussian distribution.
\begin{align}
\hat{H}(Y_i) \!&=\! \sum_{\cdot \in \{ r,c \}}\!\frac{1}{2}(1+\log (2\pi) + \log(\sigma^2+V_i^{\cdot})), \nonumber\\
\hat{H}(Y_i, Y_{i*})\!&=\!\sum_{\cdot \in \{ r,c \}}\! 1 \!+\! \log(2\pi) \!+\! \frac{1}{2}(\log\sigma^2 \!+\! \log(\sigma^2 \!+ \!2V_i^{\cdot})), \nonumber
\end{align}
where $V_i=V_i^r+j\cdot V_i^c$, and further compute $\hat{H}(Y)$ according to \eqref{eq:marginal} as follows:
\begin{align}\label{eq:hatH_calc}
    \hat{H}(Y) = \sum_{i\in \mathcal{I}} \frac{1}{2}\hat{H}(Y_i, Y_{i*})+\sum_{j\in \mathcal{J}} \hat{H}(Y_j).
\end{align}
For convenience, $\hat{H}(Y)$ can be computed in a point-wise way for all sampled positions as $\hat{H}(Y)=\sum_i\sum_{\cdot \in \{ r,c \}} \hat{H}_i^{\cdot}$, where $\hat{H}_i^{\cdot}$ is defined as follows for $\cdot \in \{ r,c \}$ and $i\in \{ \mathcal{I},\mathcal{J}\}$:
\begin{equation}\label{eq:H_i_element_wise}
\hat{H}_i^{\cdot}\!=\!\left\{
\begin{aligned}
&\frac{1}{2}(1\!+\!\log 2\pi\!+\!\frac{1}{2}\log\sigma^2 \!+\!\log(\sigma^2\!+\! 2V_i^{\cdot}))), i\!\in\! \mathcal{I}\\
&\frac{1}{2}(1\!+\!\log 2\pi\!+\! \log(\sigma^2\!+\!V_i^{\cdot})), i\!\in\!\mathcal{J}
\end{aligned}
\right.\!.
\end{equation}

\begin{remark}\upshape
\label{remark_sample}
Equation~\eqref{eq:H_i_element_wise} mathematically validates some intuitions about how the choice of $\mathbf{m}$ affects $I(Y;X)$. To find $\mathbf{m}$ that maximizes $I(Y;X)$ (which is equivalent to maximizing the marginal entropy $H(Y)$) when the measurement noise is relatively small, Algorithm~\ref{alg:mee} suggests the following:
\begin{itemize}[leftmargin=0cm, itemindent=0.5cm]
    \item \textit{Avoid redundant sampling.} To maximize the marginal entropy $H(Y)$, one should avoid allocating acquisition budget to $\mathcal{I}$ since $\log \sigma^2 \rightarrow -\infty$ as $\sigma \rightarrow 0$.
    \item \textit{Sample positions with high uncertainty.} Since $H_i$ is monotonically increasing with respect to k-space variance $V_i$ at position $i$, one should prioritize measurements in k-space where the data is most uncertain~\cite{ji2008bayesian}.
\end{itemize}
\end{remark}

We summarize the training steps for optimizing \eqref{eq:amt2} in Algorithm~\ref{alg:mee}. Once the model has been properly trained, classification results can be obtained directly with the suppression of MRI reconstruction, as detailed in Algorithm~\ref{alg:inference2}.

\begin{algorithm}[!ht] 
\renewcommand{\baselinestretch}{1.0}
\caption{Optimizing \eqref{eq:amt2} for \textit{Classification}-Adapted CS-MRI}\label{alg:mee} 
\begin{algorithmic}[1]
\renewcommand{\algorithmicrequire}{\textbf{Input:}}
\renewcommand{\algorithmicensure}{\textbf{Output:}}
\REQUIRE{The k-space variance for the real and imaginary parts over the training set  $\mathbf{V}^r,\mathbf{V}^c \in \mathbb{R}^{N}$, variance of the measurement noise $\sigma^2$, and initialized encoder $\mathcal{E}_{\phi^Y}$ and decoder $\mathcal{D}_{\phi^{T}}$.}
\WHILE{$\phi,\theta$ not converge}
\STATE Sample $\mathbf{x},\mathbf{t}$ from the training set: $\mathbf{x},\mathbf{t}\sim p_{data}(\mathbf{x,t})$ 
\STATE Sample $\mathbf{m}$ according to Algorithm~\ref{alg:amo}
\STATE Compute elements of point-reflected sampling pattern $\Tilde{\mathbf{m}}$: $\tilde{m}_i = m_{i*} \ \text{for} \ i=1,2,\cdots,N.$
\STATE Find positions that are redundantly sampled or not: $\mathcal{I} = \mathbf{m} \odot \tilde{\mathbf{m}}$, $\mathcal{J} = \mathbf{m} \odot (1 - \mathcal{I})$.
\STATE Compute $\hat{H}(Y)$ in a point-wise way according to \eqref{eq:hatH_calc} and \eqref{eq:H_i_element_wise}
\STATE Obtain $\mathbf{y}$ and $\hat{\mathbf{x}}_{zf}= \mathcal{F}^H \mathbf{U_m}^T \mathbf{y} \in \mathbb{C}^N$
\STATE Encode $\hat{\mathbf{x}}_{zf}$ in the feature $\hat{\mathbf{y}}$ following \eqref{eq:y_hat}
\STATE Evaluate $q(\mathbf{t|y})=\mathrm{Categorical}(\mathbf{t}|\mathcal{D}_{\phi^{T}}(\hat{\mathbf{y}}))$ 
\STATE Do back-propagation of \eqref{eq:amt2} to compute gradient with respect to $\phi$ and $\theta$
\ENDWHILE
\end{algorithmic} 
\end{algorithm}
\vspace{-12pt}
\begin{algorithm}[!ht] 
\renewcommand{\baselinestretch}{1.0}
\renewcommand{\arraystretch}{1.0}
\caption{Inference for \textit{Classification}-Adapted CS-MRI}\label{alg:inference2}
\begin{algorithmic}[1]
\renewcommand{\algorithmicrequire}{\textbf{Input:}}
\renewcommand{\algorithmicensure}{\textbf{Output:}}
\REQUIRE{Measurement encoder $\mathcal{E}_{\phi^Y}$, classification decoder $\mathcal{D}_{\phi^{T}}$, measurement $\mathbf{y}$, sampling pattern $\mathbf{m}$}
\STATE Obtain $\mathbf{y}$ and $\hat{\mathbf{x}}_{zf}= \mathcal{F}^H \mathbf{U_m}^T \mathbf{y} \in \mathbb{C}^N$
\STATE Encode $\hat{\mathbf{x}}_{zf}$ in the feature $\hat{\mathbf{y}}$ following \eqref{eq:y_hat}
\WHILE{obtaining sufficient classification results}
\STATE Sample $\mathbf{t}\sim q(\mathbf{t|y})=\mathrm{Categorical}(\mathbf{t}|\mathcal{D}_{\phi^{T}}(\hat{\mathbf{y}}))$ 
\ENDWHILE
\ENSURE{A series of classification results $\mathbf{t}$s}
\end{algorithmic} 
\end{algorithm}

\subsection{Relation to Existing Methods} \label{sec:relation}
\subsubsection{Relation to LI-Net~\cite{schlemper2018cardiac}}

LI-Net is designed to achieve segmentation of MRI data directly from undersampled k-space measurement, bypassing the MRI reconstruction step. LI-Net consists of two main components: (1) a segmentation auto-encoder (AE) with an encoder $\Phi$ and a decoder $\Psi$, and (2) a measurement encoder $\Pi(\mathbf{y})$. The training procedure of LI-Net involves two independent stages.
In stage 1, the AE is trained by minimizing the reconstruction error of the segmentation.
\begin{equation}
\min_{\Phi,\Psi}\mathbb{E}_{\mathbf{t}\sim p_{data}(\mathbf{t})}\|\Psi \circ \Phi(\mathbf{t})-\mathbf{t}\|_2^2,
\end{equation}
where $\mathbf{t}\sim p_{data}(\mathbf{t})$ denotes sampling true segmentation maps from the training data.
Once the AE is well-trained, its parameters are fixed, and the training proceeds to stage 2. In stage 2, the measurement encoder $\Pi(\mathbf{y})$ is trained to predict the latent representation $\mathbf{z} = \Phi(\mathbf{t})$ of the true segmentation map $\mathbf{t}$ by minimizing the MSE in the latent space:
\begin{equation}\label{eq:LINet_Pi}
    \min_{\Pi} \mathbb{E}_{\{\mathbf{y},\mathbf{t}\}\sim p_{data}(\mathbf{y},\mathbf{t})} \lVert \Pi(\mathbf{y}) - \Phi(\mathbf{t}) \rVert_2^2.
\end{equation}
When $\Pi$ is trained, segmentation maps are generated directly from the measurements via $\mathbf{t} \sim \Psi \circ \Pi(\mathbf{y})$.

The scenario that LI-Net addresses corresponds to \eqref{eq:naive-obj} with $\beta=0$ as follows:
\begin{align}\label{eq:naive-obj_beta=0}
\max_{\theta,\phi} \mathbb{E}_{r, \mathcal{M}, \mathbf{m}, \mathbf{x}, \mathbf{t}, \mathbf{y}} \mathbb{E}_{\hat{q}(\mathbf{z|t,y})} [\log q(\mathbf{t}|\mathbf{z}, \mathbf{y})- D_{KL}[\hat{q}(\mathbf{z|t,y})\|q(\mathbf{z|y})].
\end{align}
Although LI-Net and our method share similarities—such as optimizing encoder-decoder structures and introducing an auxiliary latent variable—our approach offers significant advantages over LI-Net in terms of theoretical interpretation, training procedure, sampling, and task inference.

\begin{itemize}[leftmargin=0cm, itemindent=0.5cm]
\item \textit{Theoretical interpretation:}
LI-Net is an intuitively designed “deterministic” method, while our approach is an information-theoretic “statistical” version. Recall that $\hat{q}(\mathbf{z|t,y})$ and $q(\mathbf{z|y})$ in \eqref{eq:naive-obj_beta=0} are modeled as Gaussians (see \eqref{eq:qhat(z|t,y)} and \eqref{eq:q(z|y)}), with means and variances generated by learnable decoders. Let $\mathbf{\mu}_{\hat{q}}(\mathbf{t,y})$, $\mathbf{V}_{\hat{q}}(\mathbf{t,y})$, $\mathbf{\mu}_q(\mathbf{y})$, and $\mathbf{V}_q(\mathbf{y})$ represent the means and variances of $\hat{q}(\mathbf{z|t,y})$ and $q(\mathbf{z|y})$, respectively. At inference, our method can become fully deterministic by setting the latent variable to the Gaussian mean rather than sampling it (e.g., $\mathbf{z} = \mathbf{\mu}_{\hat{q}}(\mathbf{t,y})$ rather than $\mathbf{z} \sim \mathcal{N}(\mathbf{\mu}_{\hat{q}}(\mathbf{t,y}), \mathbf{V}_{\hat{q}}(\mathbf{t,y}))$). In this deterministic scenario, the optimal measurement encoder $\mathbf{\mu}_q^*(\mathbf{y})$ minimizes the MSE loss between $\mathbf{\mu}_q(\mathbf{y})$ and $\mathbf{\mu}_{\hat{q}}(\mathbf{t,y})$:
\begin{equation}\label{eq:deterministic_muq}
    \mathbf{\mu}_q^*(\mathbf{y}) = \arg\min_{\mathbf{\mu}_q(\mathbf{y})} \mathbb{E}\lVert  \mathbf{\mu}_q(\mathbf{y}) - \mathbf{\mu}_r(\mathbf{t,y}) \rVert_2^2.
\end{equation}
Note that \eqref{eq:deterministic_muq} resembles \eqref{eq:LINet_Pi}, with the distinction that the deterministic segmentation encoder $\mathbf{\mu}_{\hat{q}}(\mathbf{t,y})$ is also conditional on $\mathbf{y}$. Moreover, LI-Net intuitively aligns the measurement encoder $\Pi(\mathbf{y})$ with the pre-trained $\Phi(\mathbf{t})$ and generates segmentation maps via $\Psi \circ \Pi(\mathbf{y})$. In contrast, we rigorously derive a principled framework from an information-theoretic perspective.

\item \textit{Training procedure:}
According to \eqref{eq:naive-obj_beta=0} and Algorithm~\ref{alg:optimize}, the latent decoder $\mathcal{D}_{\phi^{Z}}$, task-inference decoder $\mathcal{D}_{\phi^{T}}$, and measurement encoder $\mathcal{E}_{\Tilde{\phi}^Y}$ are jointly optimized in an end-to-end manner, providing a holistic training perspective. By contrast, LI-Net decomposes training into two separate stages, which may prevent it from achieving optimal parameters. For example, in LI-Net’s second training stage, the AE parameters are fixed and cannot be co-optimized with $\Pi$.

\item \textit{Sampling and Task inference:}
LI-Net only supports a fixed sampling ratio during the training stage and is limited to compressive learning. In contrast, our method enables adaptive sampling with a single model and can be adapted to broader scenarios by controlling $\beta$. Additionally, our method provides estimates of prediction uncertainty due to the ``statistical'' nature of variational inference, thereby enhancing robustness and reliability in automated medical diagnosis systems. Experimental results in Section~\ref{sec:exp-segmentation} demonstrate the superior performance of our method over LI-Net.
\end{itemize}

\subsubsection{Relation to Existing End-to-End CS-MRI Reconstruction Methods}\label{sec:relation_e2e}

As mentioned in Section~\ref{sec: beta<0}, we model the MR image posterior, $q(\mathbf{x|y})$, as a diagonal Gaussian distribution. In this formulation, maximizing $\log q(\mathbf{x|y})$ is equivalent to minimizing the MSE between the signal $\mathbf{x}$ and a transformation of $\mathbf{y}$. Specifically, suppose $q(\mathbf{x|y})$ has the mean of $R_{\phi}(\hat{\mathbf{x}}_{zf})$ and the variance of $\sigma_{\phi}^2(\hat{\mathbf{x}}_{zf})$ according to \eqref{eq:q(x|y)}. Then $\max_{\theta,\phi} \mathbb{E}_{r, \mathcal{M}, \mathbf{m}, \mathbf{x}, \mathbf{t}, \mathbf{y}}\log q(\mathbf{x|y})$ in \eqref{eq:naive-obj} is equivalent to 
\begin{equation}\label{eq:gaussian_mse}
\!\min_{\theta,\phi}\mathbb{E}_{r, \mathcal{M}, \mathbf{m}, \mathbf{x}, \mathbf{y}}\!\left[\! \sum_{i=1}^N\!\frac{ ( \mathbf{x}_i\!-\!R_{\phi}(\hat{\mathbf{x}}_{zf})_i )^2}{\sigma_{\phi}^2(\hat{\mathbf{x}}_{zf})_i}\!+\!\log \sigma_{\phi}^2(\hat{\mathbf{x}}_{zf})_i\!\right]\!,
\end{equation}
which aligns with existing end-to-end MRI reconstruction methods that employ an MSE loss~\cite{RN223_Deep, RN276, RN170, RN299, RN283, RN196}. Conversely, these end-to-end methods can be viewed as implicitly maximizing $I(Y;X)$.

A key distinction in our method is that it also learns the uncertainty in MRI reconstruction. According to the first-order optimality condition of \eqref{eq:gaussian_mse}, the optimal variance $\sigma_{\phi}^2(\hat{\mathbf{x}}_{zf})_i$ is the minimum mean square estimator (MMSE) of the reconstruction error given $\mathbf{x}_{zf}$, i.e., $\mathbb{E}[(\mathbf{x}_i - R_{\phi}(\hat{\mathbf{x}}_{zf})_i )^2 | \mathbf{x}_{zf}]$, and therefore serves as a reliable predictor of model reconstruction error, indicating uncertainty. For instance, if the reconstruction mean $R_{\phi}(\hat{\mathbf{x}}_{zf})$ significantly deviates from the ground truth $\mathbf{x}$, reflecting high uncertainty, the variance $\sigma_{\phi}^2(\hat{\mathbf{x}}_{zf})_i$ will increase accordingly to minimize \eqref{eq:gaussian_mse}. Simultaneously, the second term $\log \sigma_{\phi}^2(\hat{\mathbf{x}}_{zf})$ prevents the variance from growing unboundedly.

Additionally, these variances can be interpreted as automatic weights that dynamically adjust the importance of multi-task losses, as discussed in \cite{kendall2018multi}, where, in our context, the multiple tasks correspond to different sampling ratios. Similar variances also appear in the training of diffusion models, where they have been shown to enhance training dynamics \cite{karras2023analyzing}.

\subsubsection{Relation to Existing Task-Adapted MRI Methods}

We clarify the relationship between FSL, Tackle and our work.
\begin{itemize}
    \item \textbf{FSL~\cite{wang2024promoting}:} Similar to SeqMRI, FSL enables sample-level adaptive sampling through its Mask-generating Sub-network Module. It dynamically re-computes a new sampling mask at each iteration of the reconstruction process, conditioned on the intermediate reconstruction. This allows the sampling pattern to adapt in real-time to individual patient characteristics.

    \item \textbf{Tackle~\cite{wu2023learning}:} To our best knowledge, Tackle does not provide real-time adaptive sampling during acquisition. However, it introduces ROI-oriented reconstruction and task-oriented optimization for the sampling mask, thereby providing task-level adaptive sampling that tailors the pattern to the specific diagnostic objective.
    
    \item \textbf{Proposed Method:} InfoMRI introduces a distinct and complementary approach. Its primary novelty lies in its information-theoretic foundation, which provides a unified principle for jointly optimizing sampling, reconstruction, and task inference under a single objective: maximizing task-relevant information within a measurement budget. This information-centric principle is orthogonal to existing adaptive sampling strategies and could potentially be combined with them. For instance, sample-level adaptation for InfoMRI could be enabled by leveraging approaches like deep adaptive design~\cite{foster2021deep} or reinforcement learning~\cite{blau2022optimizing}.
\end{itemize}

Different from FSL and Tackle that offer adaptive sampling strategies at sample and task levels, we provide a new theoretical lens for task-adapted MRI by optimizing information flow throughout the entire imaging-to-diagnosis pipeline.

\section{Experiments} \label{sec:experiements}
In this section, we evaluate the proposed methods on various MRI-related tasks. 
For convenience, we refer to our method as \textit{InfoMRI} throughout the experiments. To validate the efficacy of InfoMRI, we compare with (1) CS-MRI reconstruction methods, (2) task-adapted CS-MRI methods, and (3) compressed learning methods. 
Note that all experimental data are simulated using a single-coil forward model.

We select widely adopted segmentation and classification tasks~\cite{wang2024promoting,wu2023learning,RN299}. The theoretical generality can be applied beyond segmentation and classification, with clear potential for broader applicability.
Experimental results are able to:
\begin{enumerate}
\item Highlight the advantages of amortized optimization proposed in Section~\ref{sec:amortize}, including flexible control of sampling ratios, superior performance, and training efficiency.
\item Demonstrate the robust and reliable prediction and uncertainty quantification by the proposed method to address the uncertainty problem discussed in Section~\ref{sec:taskadaptedmri}. 
\item Validate the effectiveness of marginal entropy minimization proposed in Section~\ref{sec:beta>0} for optimizing a sampling pattern that prevents recovering the original image for privacy protection.
\end{enumerate}

\subsection{Comparisons to CS-MRI Reconstruction Methods} \label{sec:exp-mri-reconstruction}
In this section, we implement pure CS-MRI reconstruction corresponding to $\beta\rightarrow-\infty$, to highlight the adaptive sampling via the proposed amortized optimization,  demonstrating the flexible sampling ratio control and training efficiency compared to current end-to-end methods, as well as superior performance relative to existing active acquisition methods. The training and inference steps are illustrated in Algorithm~\ref{alg:optimize} and \ref{alg:inference}, respectively, except for the steps related to task inference.

\begin{figure}[!t]
\centering
\includegraphics[width=0.95\textwidth]{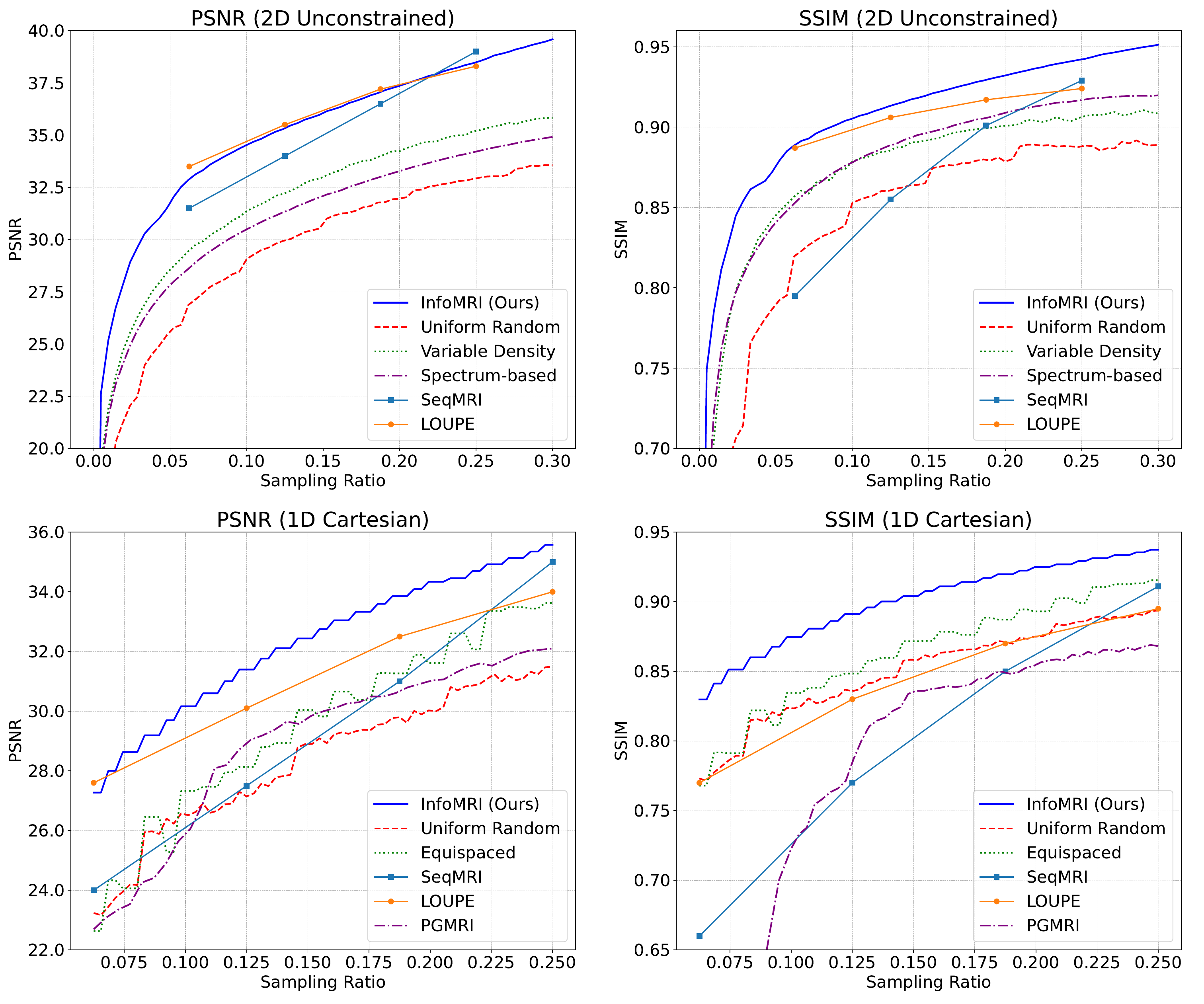}
\caption{The rate-distortion curves of various methods across different sampling ratios in terms of PSNR and SSIM on the fastMRI database~\cite{zbontar2018fastmri}. The curves for LOUPE and SeqMRI are plotted at four fixed sampling ratios of 6.25\%, 12.5\%, 18.75\%, and 25\%. In contrast, the curves for other methods are plotted using 64 points sampled equidistantly from the sampling rate range of [1\%, 30\%] for \textit{2D unconstrained sampling} and [6.25\%, 25\%] for \textit{1D Cartesian sampling}.}\label{fig:rd-curve-new}
\end{figure}

\subsubsection{Experimental Setup} \label{sec:pure-recon-setup}

We compare state-of-the-art deep learning-based CS-MRI methods, including the end-to-end approaches LOUPE~\cite{RN170} and SeqMRI~\cite{RN283}, the active acquisition method PGMRI~\cite{RN279}, as well as traditional sampling strategies such as uniform random~\cite{candes2006robust}, variable density~\cite{lustig2007sparse}, equispaced sampling~\cite{zbontar2018fastmri}, and spectrum-based method~\cite{vellagoundar2015robust}. The goal of this comparison is to demonstrate the superiority of the proposed adaptive sampling method. For a fair comparison, all methods are paired with U-Net~\cite{ronneberger2015u} for reconstruction. Specifically, the end-to-end methods jointly optimize their sampling modules with the reconstruction U-Net, while the active acquisition method and traditional sampling strategies use U-Net to learn the projection from measurements to the target MR images. The proposed InfoMRI employs the same U-Net architecture to outputs the mean and variance of $q(\mathbf{x|y})$.  

We utilize the pre-trained models of LOUPE and SeqMRI from the open-source repository (\url{https://github.com/tianweiy/SeqMRI}), train PGMRI from the open-source repository (\url{https://github.com/Timsey/pg_mri}), and train the other models from scratch. For the traditional sampling strategies and our PGN, the sampling ratio $r$ is randomly generated from $\mathcal{U}[0, 0.3]$ during the training stage. We employ the Adam optimizer with a learning rate of $1 \times 10^{-4}$ and a weight decay of $1 \times 10^{-4}$ to optimize the reconstruction U-Nets for the proposed InfoMRI and traditional sampling strategies. For the proposed PGN, we use the Adam optimizer with a learning rate of $1 \times 10^{-2}$ and no weight decay. The models are trained with a batch size of 16 for 50 epochs.

We benchmark the reconstruction performance of all methods on the NYU fastMRI database~\cite{zbontar2018fastmri}, a widely-used large-scale public dataset for CS-MRI reconstruction research. For the data preprocessing pipeline, we follow the implementation described in~\cite{RN209}. Specifically, we restrict the dataset to single-coil scans, and the slices are cropped to the central $128 \times 128$ region in k-space for computational efficiency. 
Note that cropping the rectangular k-space to a square matrix will result in anisotropic spatial resolution. However, this preprocessing step is a common practice adopted from prior works \cite{RN170,RN283,RN279} to standardize input dimensions for network architectures. To provide a fair and controlled environment for comparison, this preprocessing step was applied uniformly to all methods evaluated in our study, including all baselines. Each 2D slice was treated as an independent sample. The proposed framework can be readily extended to 3D MR acquisitions by adapting the network architectures to utilize 3D convolutions and modifying the PGN to output a 3D k-space mask. Critically, the core information-theoretic principles and the amortized optimization strategy remain identical.

The original validation set is further split into a new validation set and a test set, using the same splitting setup as in~\cite{RN209}. The standard deviation $\sigma$ of the measurement noise is set to $5 \times 10^{-5}$. Peak signal-to-noise ratio (PSNR) and structural similarity index (SSIM) are adopted as performance metrics. For our method, we use the mean of $ q(\mathbf{x|y})$ as the MRI reconstruction image for calculating the metric values and visualization.

Following~\cite{RN170,RN283}, in addition to the acquisition budget constraint, we conduct further experiments that take into account the physical constraints of real-world MRI scanners. Specifically, the sampling positions are constrained to lie along lines in k-space that are orthogonal to the provided read-out direction, a scheme commonly referred to as \textit{1D Cartesian sampling}. In contrast, sampling without this line constraint is referred to as \textit{2D unconstrained sampling} in the following sections of the paper. 
Note that variable density~\cite{lustig2007sparse} only support \textit{2D unconstrained sampling}, while equispaced sampling~\cite{zbontar2018fastmri} and PGMRI~\cite{RN279} only support \textit{1D Cartesian sampling}. Uniform random~\cite{candes2006robust}, LOUPE~\cite{RN170}, SeqMRI~\cite{RN283}, and our InfoMRI support both scenarios. 

\begin{figure*}[!t]
\centering
\includegraphics[width=0.95\textwidth]{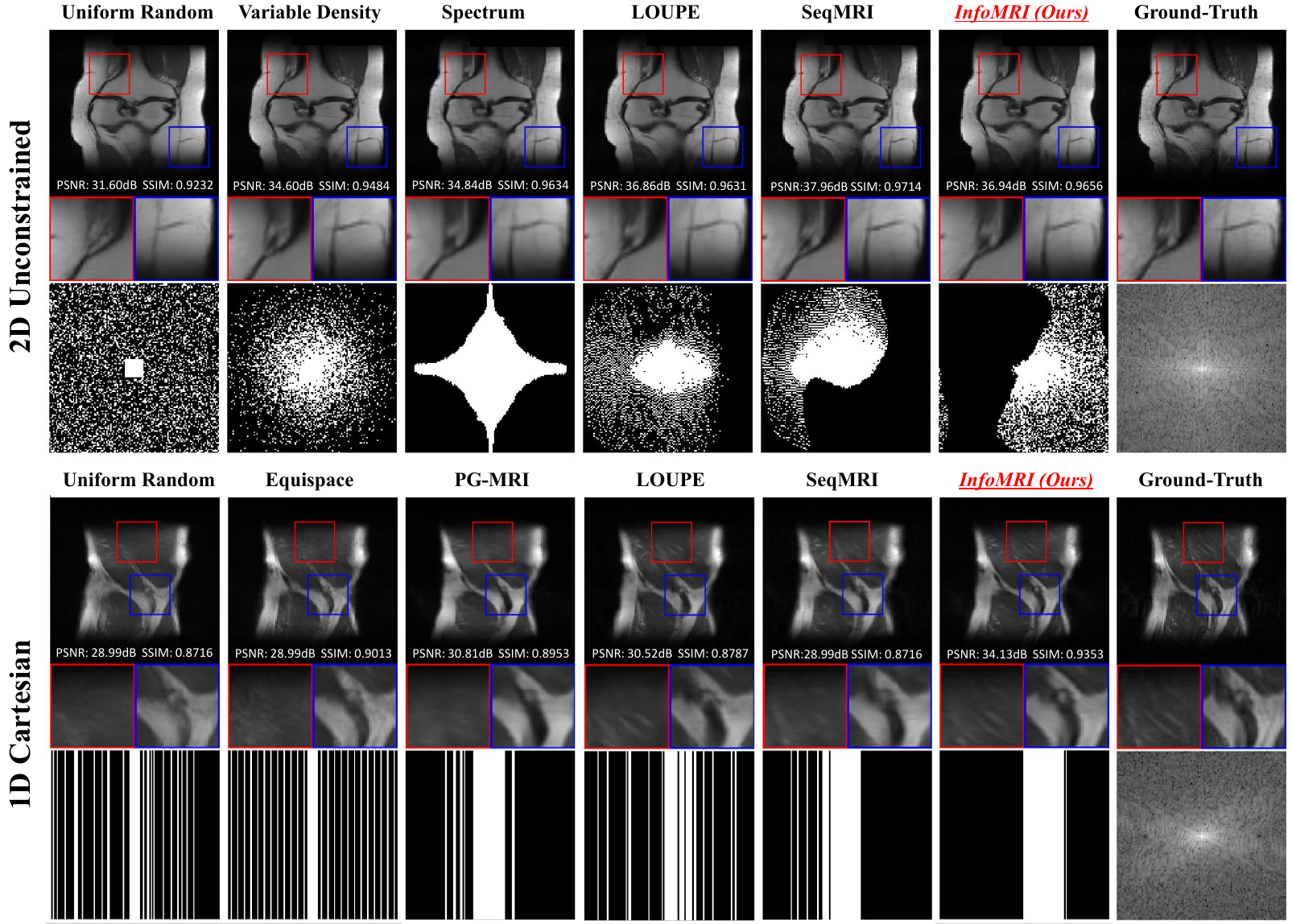}
\caption{Visualizations of MR reconstruction images and the corresponding sampling patterns at the 25\% sampling rate on the fastMRI database. The rightmost column shows the ground truth MR images (top), the amplified regions (middle), and the k-space measurement (bottom). The other columns display the reconstruction images, the amplified areas, and the generated sampling patterns from top to bottom. The PSNR and SSIM values are labeled in the top right corner of the reconstruction images.}\label{fig:2dviz-new}
\end{figure*}

\subsubsection{Reconstruction Performance}

\figurename~\ref{fig:rd-curve-new} illustrates the performance of various methods across different sampling ratios through rate-distortion curves. Note that the end-to-end methods LOUPE and SeqMRI do not support flexible sampling ratio control and are thus plotted at four fixed sampling ratios: 6.25\%, 12.5\%, 18.75\%, and 25\%. Specifically, LOUPE is trained separately for each ratio, and SeqMRI can only achieve these four sampling ratios using a single trained model. \figurename~\ref{fig:2dviz-new} further visualizes the MRI reconstruction images and the corresponding sampling patterns at the 25\% sampling ratio.

In the \textit{2D unconstrained sampling} scenario, our method achieves comparable performance to LOUPE, and outperforms SeqMRI except at the 25\% sampling rate in terms of PSNR. In the \textit{1D Cartesian sampling} scenario, our method is slightly inferior to LOUPE at the 6.25\% sampling rate. In all other cases, our method demonstrates superior performance over all baselines. 
This can be attributed to the fact that LOUPE is trained separately for each ratio, and SeqMRI is specifically optimized for the 25\% sampling rate. Notably, SeqMRI exhibits a significant drop in reconstruction performance at its first three sampling rate points, as these represent intermediate reconstruction stages for the 25\% sampling rate and are not specifically optimized.
In contrast, our method supports flexible sampling ratio control via a single trained model. PGMRI, the active acquisition method that also supports flexible sampling ratio control, shows a substantial decrease in reconstruction performance across all sampling ratios compared to our method. This is because PGMRI solely optimizes the sampling strategy based on a pre-trained MRI reconstruction model, which does not enable joint optimization of sampling and reconstruction.

\subsubsection{Performance with Advanced Reconstruction Methods}

To demonstrate the modularity and broad applicability of our framework, we evaluate its performance when integrated with various state-of-the-art reconstruction architectures, including Plug-and-Play (PnP) algorithms and deep unrolling networks. The experiments were conducted on the Brain MRI dataset from ADMM-Net~\cite{yang2018admm}. Our framework was integrated with three distinct reconstruction models:
\begin{itemize}
    \item \textbf{InfoMRI-PnP:} This model utilizes a pre-trained DRUNet~\cite{zhang2021plug} as the denoiser prior within a PnP framework, allowing for a direct comparison with PnP-Parallel-MRI~\cite{pour2019deep}.
    \item \textbf{InfoMRI-Unroll:} This model adopts the architecture of an unrolling network, specifically ISTA-Net$^+$~\cite{zhang2018ista}, as its reconstruction backbone.
    \item \textbf{InfoMRI-UNet:} This model serves as a baseline using a standard U-Net architecture, enabling comparison with other U-Net-based sampling methods like LOUPE~\cite{RN170}.
\end{itemize}

As shown in Table~\ref{tab:unroll}, replacing U-Net with an unrolled backbone yields higher overall reconstruction quality (InfoMRI-Unroll vs. InfoMRI-UNet), as expected. Crucially, the trends reported in the main tables remain unchanged when switching to the unrolled backbone: (i) learned sampling continues to outperform fixed patterns (e.g., InfoMRI-Unroll v.s. ISTA-Net+); and (ii) amortizing training over a range of sampling ratios matches the performance of a single–sampling-ratio model. InfoMRI-Unroll closely \emph{matches} PUERT in reconstruction fidelity across accelerations while supporting variable sampling ratios within a single trained model. In addition, InfoMRI-Unroll-MSE underperforms InfoMRI-Unroll, highlighting the benefit of our uncertainty-aware information objective, which adaptively weights losses across sampling ratios, reduces gradient variance, and improves final performance.

\begin{table*}
\centering
\caption{Results on the \textit{Brain MRI dataset} from ADMM-Net. InfoMRI-Unroll-MSE means remove the uncertainty weight and train the model solely on MSE.} \label{tab:unroll}
\begin{tabular}{@{}lccccccc@{}}
\toprule
\multirow{2}{*}{Method} & \multirow{2}{*}{Sampling Pattern} &  \multicolumn{2}{c}{$\frac{100}{15}\times$ Acceleration} &  \multicolumn{2}{c}{$\frac{100}{10}\times$ Acceleration} & \multicolumn{2}{c}{$\frac{100}{5}\times$ Acceleration}  \\
\cmidrule(lr){3-4}  \cmidrule(lr){5-6} \cmidrule(lr){7-8}
& &  PSNR~$\uparrow$ & SSIM~$\uparrow$ & PSNR~$\uparrow$ & SSIM~$\uparrow$ & PSNR~$\uparrow$ & SSIM~$\uparrow$ \\
\midrule
MD-Recon-Net~\cite{ran2020md} & Radial & 37.25 & 0.9360 & 35.10 & 0.9120 & 31.05 & 0.8250 \\
PnP-Parallel-MRI~\cite{pour2019deep}   & Radial & 37.33 & 0.9388 & 35.14 & 0.9111 & 31.01 & 0.8221 \\
ADMM-Net~\cite{yang2018admm}     & Radial & 36.83 & 0.9306 & 34.46 & 0.8972 & 30.13 & 0.7958 \\
Self-Supervised MRI~\cite{hu2021self} & Radial & 36.58 & 0.9218 & 33.42 & 0.8924 & 30.18 & 0.8068 \\
ISTA-Net+~\cite{zhang2018ista}   & Radial & 37.07 & 0.9343  & 34.73 & 0.9052 & 30.64 & 0.8176 \\
LOUPE~\cite{RN170}       & Learned & 38.30 & 0.9473 & 36.88 & 0.9381 & 35.34 & 0.9245 \\
PUERT~\cite{RN196}       & Learned & 41.01 & 0.9655 & 39.23 & 0.9565 & 36.65 & 0.9399 \\
\midrule
InfoMRI-PnP         & Learned & 40.08 & 0.9529 & 38.54 & 0.9441 & 35.66 & 0.9215 \\
InfoMRI-UNet        & Learned & 38.35 & 0.9480 & 36.93 & 0.9391 & 35.25 & 0.9232 \\
InfoMRI-Unroll      & Learned & 40.96 & 0.9596 & 39.22 & 0.9591 & 36.61 & 0.9393 \\
InfoMRI-Unroll-MSE  & Learned & 40.32 & 0.9590 & 38.61 & 0.9492 & 35.91 & 0.9293 \\
\bottomrule
\end{tabular}
\end{table*}

\subsubsection{Effectiveness of Amortized Optimization} 
\begin{figure}[!t]
\includegraphics[width=0.95\textwidth]{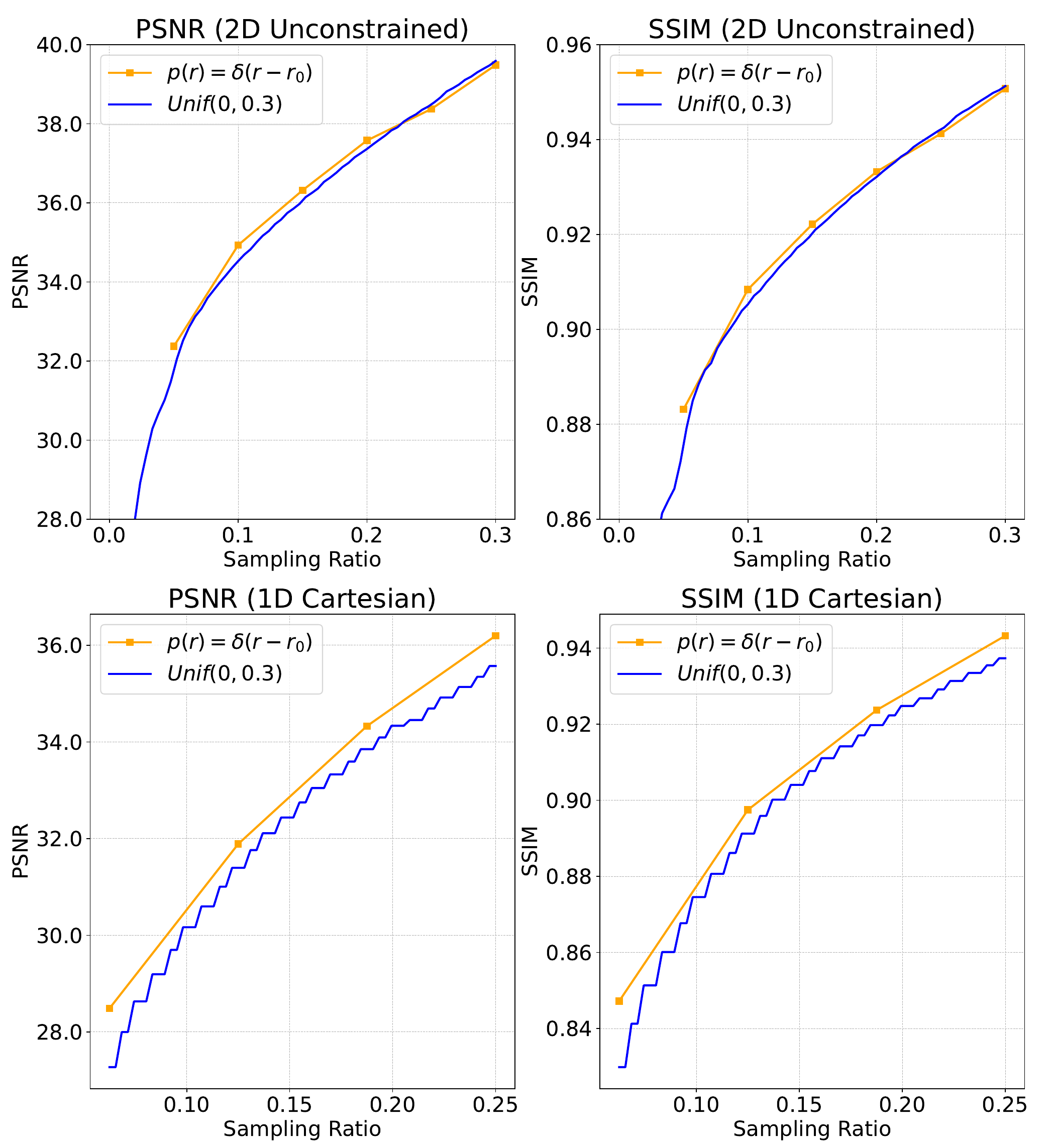}
\caption{Reconstruction performance comparison of models trained with $p(r)=\mathcal{U}[0,0.3]$ versus models trained individually at specific sampling ratios with $p(r)=\delta(r-r_0)$.  The performance is evaluated on the fastMRI database.}
\label{fig:delta-new}
\end{figure}

\figurename~\ref{fig:delta-new} compares models trained with $r \sim p(r) = \mathcal{U}[0,0.3]$ and models individually trained at specific sampling ratios (\emph{i.e.}, with $p(r)$ set as a delta distribution at a specific sampling ratio $r_0$) for the \textit{2D unconstrained sampling} scenario. The results indicate that models trained with $r \sim \mathcal{U}[0,0.3]$ achieve nearly identical rate-distortion performance compared to independently trained models, demonstrating the effectiveness of the proposed amortized optimization approach in achieving adaptive sampling and reducing training complexity.

Due to the conjugate symmetry property of k-space~\cite{defazio2019offset}, sampling both a position and its point-reflected counterpart (rotated 180 degrees for 2D images) introduces redundancy (see Section~\ref{sec:beta>0}).  We have introduced a quantitative \textit{Redundancy Ratio} metric based on the conjugate symmetry properties of MRI k-space. We define the redundancy ratio of a sampling mask $\mathbf{m}$ as:
\begin{equation}
    \text{Redundancy Ratio} = \tfrac{|\mathcal{I}|}{|\mathcal{I}|+|\mathcal{J}|},
\end{equation}
where $\mathcal{I}$ and $\mathcal{J}$ are defined in Section~\ref{sec:beta>0}.

\begin{figure}[!t]
\centering \includegraphics[width=0.95\textwidth]{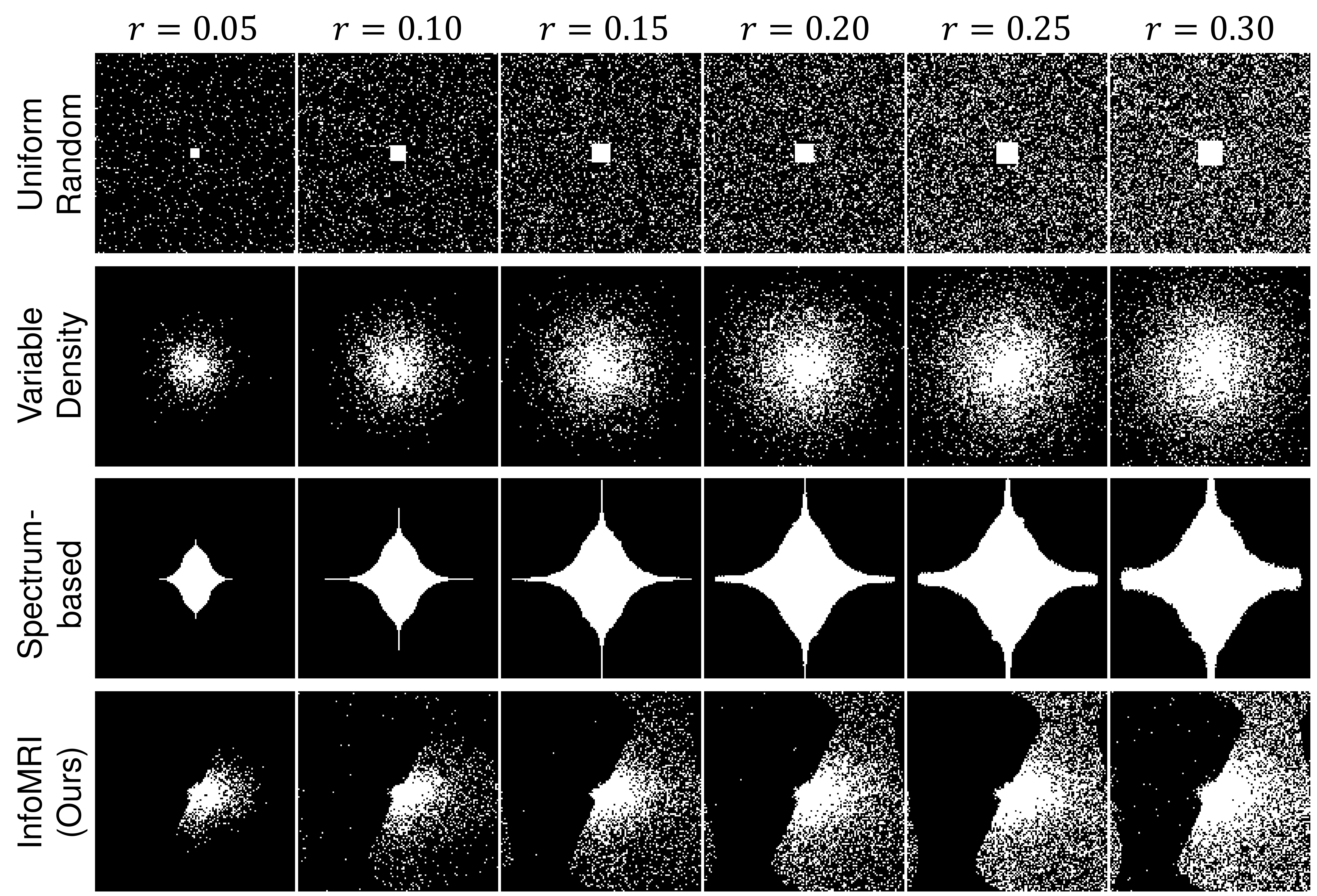}
\caption{Sampling patterns produced by traditional sampling strategies and our PGN as $r$ increases from 0.05 to 0.3. }\label{fig:mask_samples-new}
\end{figure}

In Table~\ref{tab:redundant}, we quantitatively report the redundancy ratio of different sampling patterns, and in \figurename~\ref{fig:mask_samples-new}, we visualize the sampling patterns produced by our PGN and traditional sampling strategies. Notably, even without manually imposed constraints, PGN adaptively learns to exploit this inherent redundancy in k-space, thereby improving acquisition efficiency, demonstrating that our learned PGN sampling pattern is theoretically sound and more efficient at acquiring non-redundant spatial frequencies compared to baseline methods.


\begin{table}[!t]
\centering
\caption{Redundancy ratios for the sampling masks shown in Fig.~\ref{fig:mask_samples-new}. Lower values denote higher sampling efficiency.}\label{tab:redundant}
\centering
\begin{tabular}{@{}lcccc@{}}
\toprule
Methods & $r=0.05$ & $r=0.15$ & $r=0.25$ \\

\midrule
Uniform random    & 0.11 & 0.20 & 0.28 \\
Variable density  & 0.51 & 0.52 & 0.51 \\
Spectrum          & 1.0 & 1.0 & 1.0 \\
InfoMRI (Ours)    & \textbf{0.07} & \textbf{0.09}  & \textbf{0.14} \\
\bottomrule
\end{tabular}
\end{table}

\subsubsection{Reconstruction Uncertainty Quantification} 
\label{exp:uncertainty}

\begin{figure}[!t]
    \centering \includegraphics[width=0.95\textwidth]{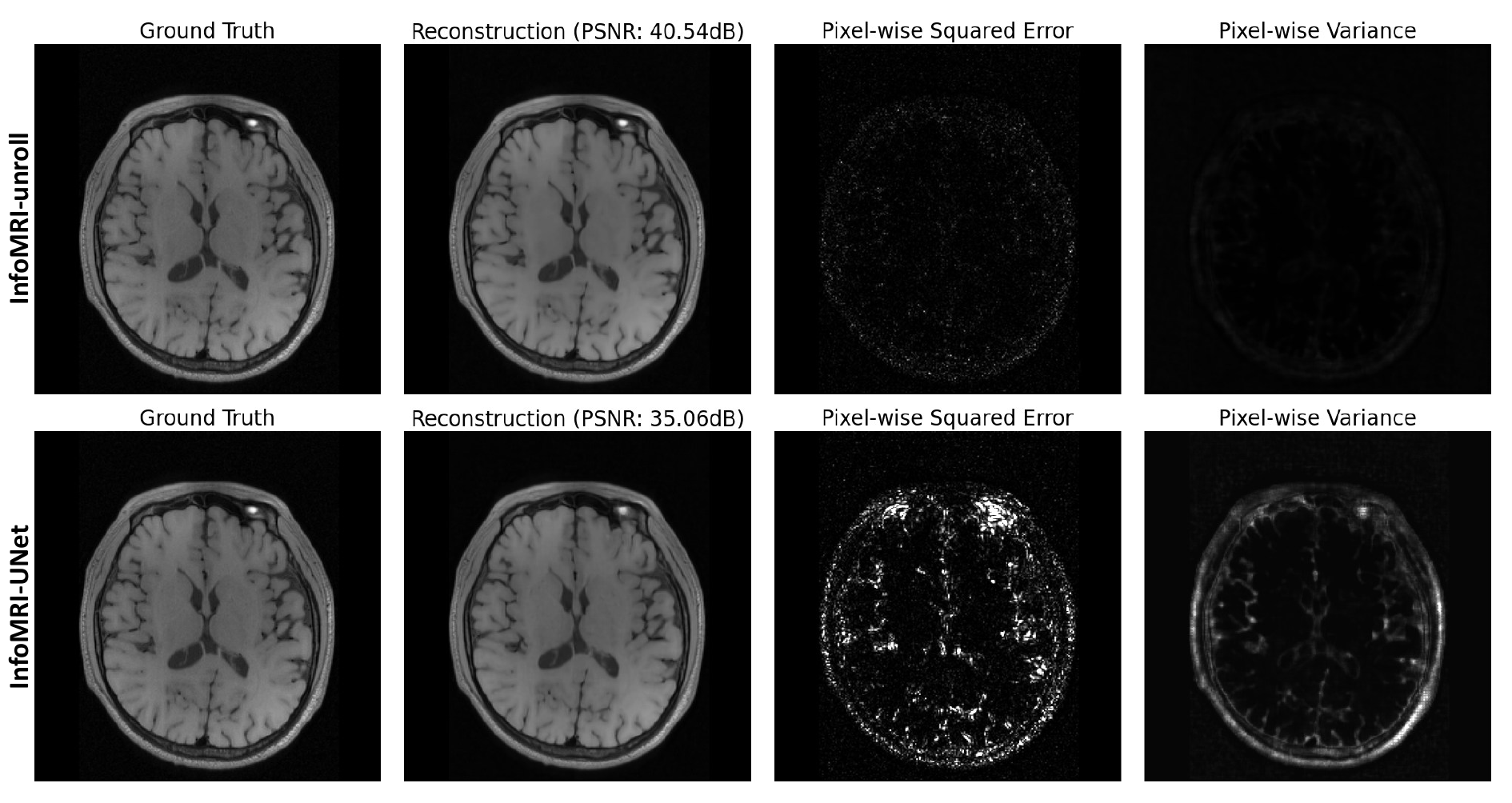}
    \caption{Visualization of InfoMRI at $r=0.15$ for uncertainty quantification on the \textit{Brain MRI dataset}. InfoMRI-Unroll reconstructs better than InfoMRI-UNet and both variants accurately predict the reconstruction error with learned uncertainty.}
    \label{fig:recon-uncertain-new}
\end{figure}

The variance of the posterior approximation $q(\mathbf{x|y})$ is calculated for estimating the reconstruction error and evaluating the reconstruction uncertainty, as elaborated in Section~\ref{sec:relation}. \figurename~\ref{fig:recon-uncertain-new} shows that the pixel-wise variances of $q(\mathbf{x|y})$ highlight regions and textures that closely resemble the ground truth reconstruction error, demonstrating the high accuracy of the uncertainty quantification provided by InfoMRI. This can assist medical professionals in identifying areas where the reconstruction is less clear, ultimately aiding in medical diagnosis.

\begin{figure}[!t]
\centering
\includegraphics[width=0.95\textwidth]{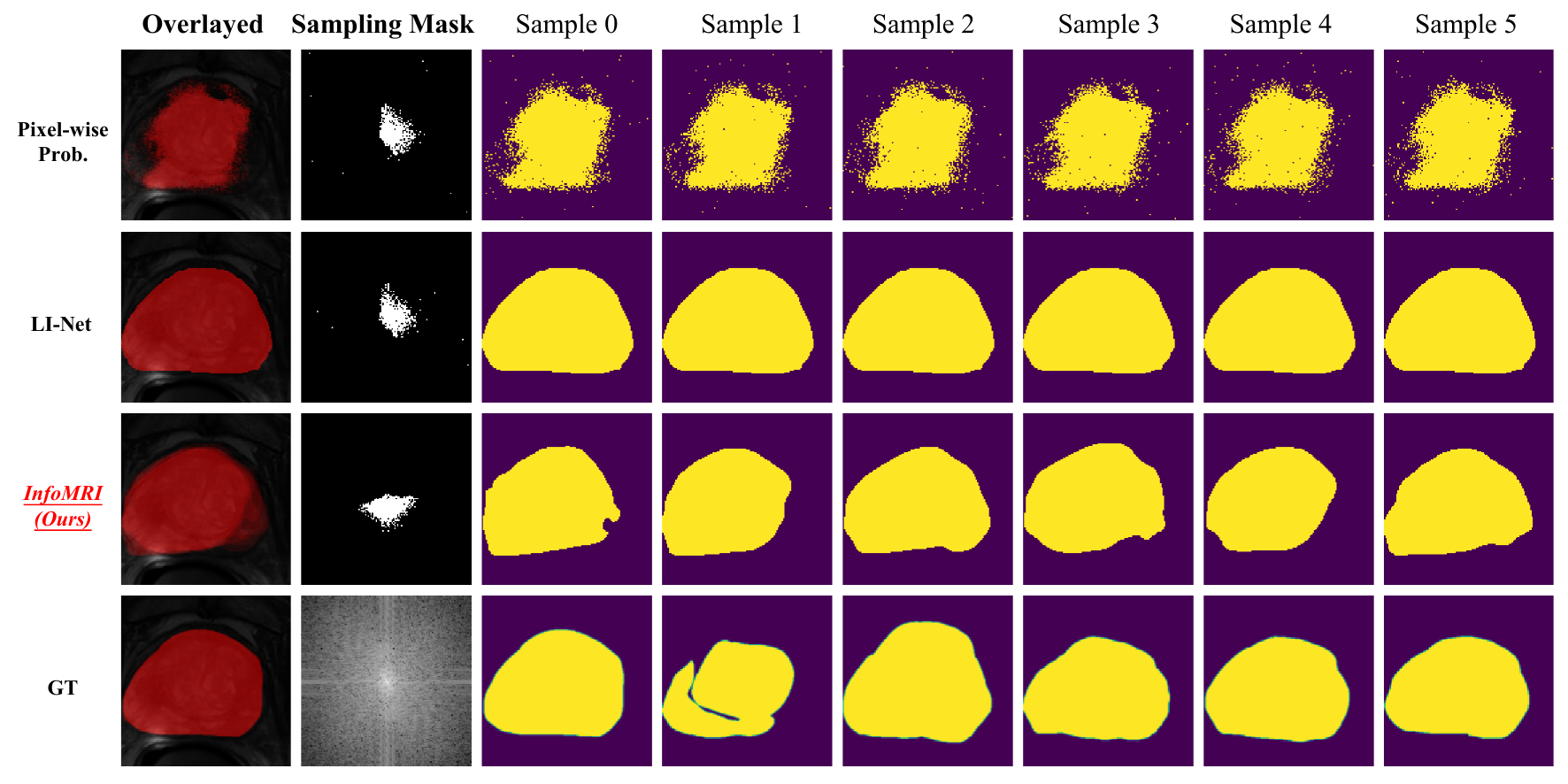}
\caption{Comparison of segmentation predictions on  \textit{QUBIQ 2021 dataset}. Pixel-wise probability methods exhibit spatially incoherent segmentation maps. LI-Net produces spatially coherent segmentation map but no variation. InfoMRI successfully recovers the spatial covariance of the segmentations, and produces diverse and spatially coherent samples.}
\label{fig:seg-compare}
\end{figure}

\begin{table*}[!t]
\centering
\caption{Segmentation performance comparisons  on \textit{QUBIQ 2021 dataset}. Lower GED and Higher DSC indicate better segmentation performance. We use \textbf{bold} and \underline{underline} to highlight the best and the second best, respectively. Note that $8\times$, $16\times$, $24\times$, and $32\times$ accelerations correspond to sampling ratios of 1/8, 1/16, 1/24 and 1/32, respectively.} \label{tab:segmentation-2d}
\begin{tabular}{@{}llcccccccc@{}}
\toprule
\multirow{2}{*}{Method} & \multirow{2}{*}{Sampling Patterns} &  \multicolumn{2}{c}{$8\times$ Acceleration} &  \multicolumn{2}{c}{$16\times$ Acceleration} & \multicolumn{2}{c}{$24\times$ Acceleration} &  \multicolumn{2}{c}{$32\times$ Acceleration}  \\
\cmidrule(lr){3-4}  \cmidrule(lr){5-6}  \cmidrule(lr){7-8}  \cmidrule(lr){9-10}
 &  & GED~$\downarrow$ & DSC~$\uparrow$ & GED~$\downarrow$ & DSC~$\uparrow$ & GED~$\downarrow$ & DSC~$\uparrow$ & GED~$\downarrow$ & DSC~$\uparrow$  \\
\midrule
\multirow{3}{*}{TradPat-UNet-RS} & Uniform random   &  0.2396 & 0.9211 & 0.4265 & 0.8566 & 0.7296 & 0.7275 & 0.7685 & 0.7039 \\
& Poisson disk & 0.2319 & 0.9147 & 0.3124 & 0.8903 & 0.4570 & 0.8343 & 0.5226 & 0.8129 \\
& Variable density &  0.1573 & 0.9474 & 0.2324 & 0.9171 & 0.2801 & 0.9047 & 0.3008 & 0.8979 \\
\midrule
\multirow{3}{*}{LI-Net~\cite{schlemper2018cardiac}} & Uniform random & 0.2516 & 0.9104 & 0.3601 & 0.8737 & 0.4750 & 0.8311 & 0.4872 & 0.8263 \\
& Poisson disk & 0.2089 & 0.9213 & 0.2443 & 0.9109 & 0.3732 & 0.8707 & 0.3871 & 0.8590 \\
& Variable density & 0.1318 & 0.9544 & 0.1585 & 0.9466 & 0.2458 & 0.9150 & {\underline{0.2161}} & {\underline{0.9260}} \\
\midrule
LOUPE-UNet-RS & \multirow{5}{*}{Learned} & 0.1404 & 0.9516 & 0.2145 & 0.9251 & 0.3054 & 0.8802 & 0.3458 & 0.8696 \\
LOUPE-LI-Net & & 0.1215 & 0.9559 & 0.1368 & 0.9503 & {\underline{0.1643}} & {\underline{0.9428}} & 0.2306 & 0.9210 \\
SemuNet~\cite{RN299} & & 0.1166 & \underline{0.9606} & 0.1296 & \underline{0.9556} & 0.2244 & 0.9223 & 0.2549 & 0.9131 \\
Tackle~\cite{wu2023learning} & & \underline{0.1009} & \textbf{0.9644} & \underline{0.1243} & \textbf{0.9570} & 0.1683 & 0.9404 & 0.2216 & 0.9243 \\
InfoMRI (Ours) & & \textbf{0.0987} & 0.9496 & \textbf{0.1009} & 0.9474 & \textbf{0.1086} & \textbf{0.9441} & \textbf{0.1142} & \textbf{0.9429} \\
\bottomrule
\end{tabular}
\end{table*}

\begin{figure}[!t]
\centering \includegraphics[width=0.95\textwidth]{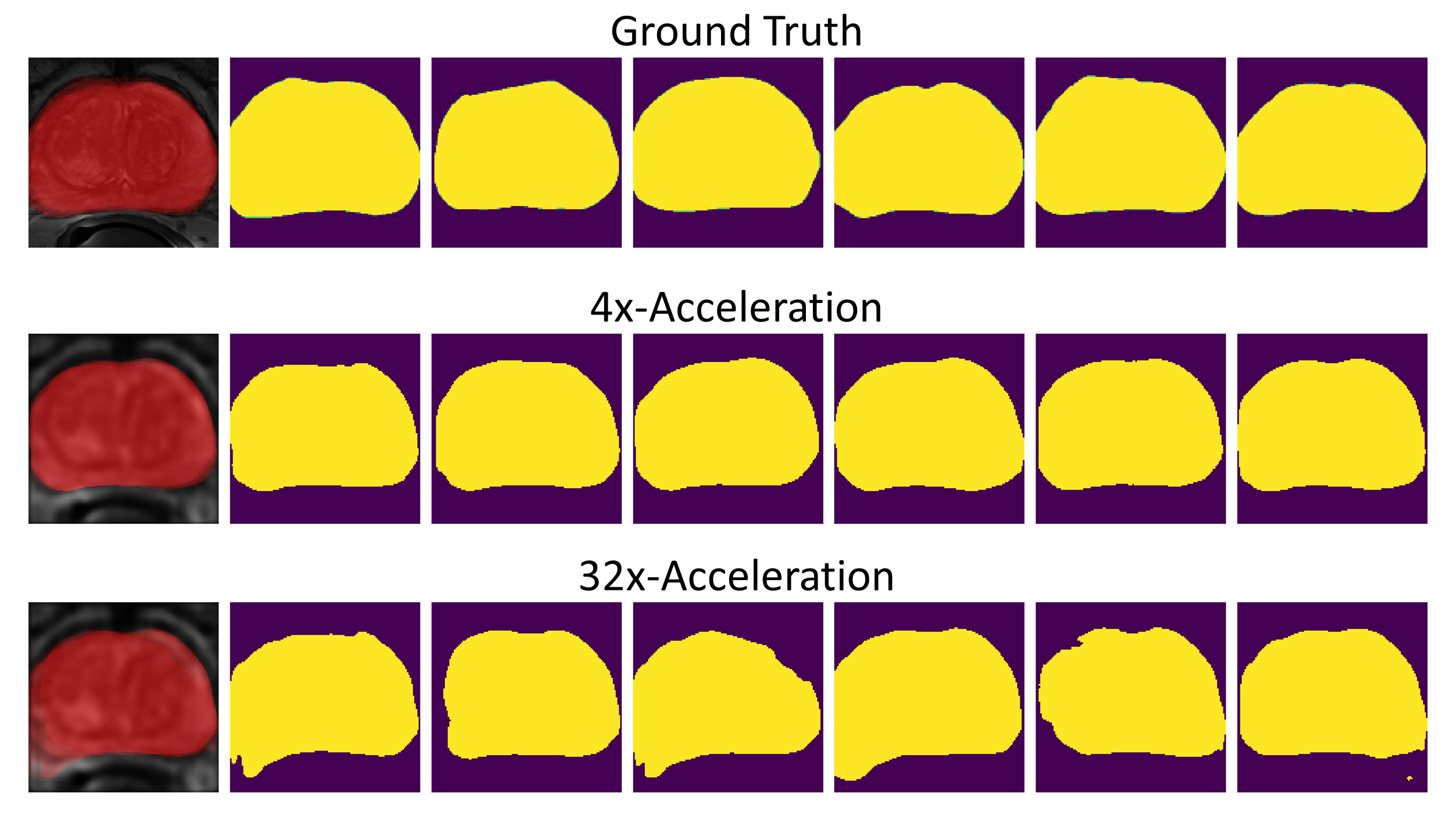}
\caption{Visualization of posterior samples from $q(\mathbf{t|y})$ under different acceleration factors for our InfoMRI  on  \textit{QUBIQ 2021 dataset}. As the acceleration factor increases, $q(\mathbf{t|y})$ generates more diverse samples to address the increasing uncertainty caused by the reduced amount of information.}
\label{fig:viz-uncertainty}
\end{figure}

\subsection{Comparisons to Task-adapted CS-MRI Methods} 
\label{sec:exp-segmentation}
In this section, we implement segmentation-adapted CS-MRI for the case where $ \beta < 0 $, but not approaching $-\infty$, highlighting both the segmentation performance and the simultaneous prediction of segmentation uncertainty.  

\subsubsection{Experimental Setup}

We compare our method with state-of-the-art task-adapted CS-MRI methods for segmentation using undersampled k-space measurement, including LI-Net~\cite{schlemper2018cardiac}, SemuNet~\cite{RN299}, and Tackle~\cite{wu2023learning}. Following~\cite{wu2023learning}, we also compare with LOUPE-UNet-RS, TradPat-UNet-RS, and LOUPE-LI-Net. We categorize the baselines into two groups.
\begin{itemize}[leftmargin=0cm, itemindent=0.5cm]
    \item \textit{Implementing segmentation after MRI reconstruction.} This category includes SemuNet, LOUPE-UNet-RS, and TradPat-UNet-RS. In LOUPE-UNet-RS, the model first generates sampling patterns following LOUPE to produce measurements, then performs MRI reconstruction using a U-Net, and finally applies segmentation with another U-Net. TradPat-UNet-RS follows the same configuration as LOUPE-UNet-RS, except that the sampling patterns are generated using traditional sampling strategies, including uniform random~\cite{candes2006robust}, variable density~\cite{lustig2007sparse}, and Poisson disk methods~\cite{wu2023learning}.
    \item \textit{Implementing segmentation directly based on measurements.} This category includes LI-Net, Tackle, LOUPE-LI-Net, and our method. LOUPE-LI-Net replaces the sampling pattern of LI-Net with a learned sampling pattern from LOUPE.
\end{itemize}

The proposed InfoMRI uses the encoder and decoder architecture from latent diffusion models~\cite{rombach2022high} to construct the encoders \( \mathcal{E}_{\phi^Y}, \mathcal{E}_{\Tilde{\phi}^Y} \) and decoders \( \mathcal{D}_{\phi^{Z}}, \mathcal{D}_{\phi^{T}}, \mathcal{D}_{\phi^{X}} \). 
We employ the same optimizer, learning rate, and weight decay in Section~\ref{sec:pure-recon-setup} to train all the models for 20,000 training steps. 
Following~\cite{higgins2017beta}, we adjust the weight parameters in \eqref{eq:naive-obj} to balance the reconstruction term and KL divergence term. The training objective is modified as
\begin{align}\label{eq:naive-obj-exp}
\max_{\theta,\phi} \mathbb{E}_{r, \mathcal{M}, \mathbf{m}, \mathbf{x}, \mathbf{t}, \mathbf{y}} \mathbb{E}_{\hat{q}(\mathbf{z|t,y})} [w_{1}\log q(\mathbf{t}|\mathbf{z}, \mathbf{y}) 
- w_{2} D_{KL}[\hat{q}(\mathbf{z|t,y})||q(\mathbf{z|y})] + w_{3} \log q(\mathbf{x|y})],
\end{align}
where \( w_{1}, w_{2}, w_{3} \) are set to 1, 50, and 1 in our experiments, and $\phi := (\phi^Y,\, \Tilde{\phi}^Y,\, \phi^{Z},\, \phi^{T},\, \phi^{X})$.

Following~\cite{RN197, RN230, RN199}, we use the Dice Similarity Coefficient (DSC) to evaluate segmentation performance and the Generalized Energy Distance (GED) to measure the similarity between distributions. GED is defined as 
\begin{equation}
D_{GED}^2(q,p)\! =\! 2\mathbb{E}[d_\text{IoU}(\mathbf{\hat{t},t})]\!-\!\mathbb{E}[d_\text{IoU}(\mathbf{\hat{t}, \hat{t}'})]\!-\!\mathbb{E}[d_\text{IoU}(\mathbf{t,t'})],
\end{equation}
where \( d_\text{IoU} = 1 - \text{IoU} \), \( \mathbf{\hat{t}}, \mathbf{\hat{t}'} \) are independently drawn from \( q \), and \( \mathbf{t}, \mathbf{t'} \) are independently drawn from \( p \). Here, \( q \) and \( p \) represent the predicted and ground truth distributions over segmentations, respectively. For deterministic methods, \( q \) is set to a delta peak at the predicted segmentation. In our experiments, we draw 32 samples from the predicted distribution \( q \).

We use the prostate segmentation dataset from the QUBIQ 2021 challenge~\cite{RN301}, which benchmarks methods for addressing uncertainty in medical image segmentation. This dataset consists of MRI images, making it more appropriate for evaluating the proposed InfoMRI method, which is specifically designed for MRI, compared to datasets from other medical imaging modalities (e.g., CT datasets used in~\cite{RN221, RN230}). 
Each prostate image in this dataset includes six labels provided by domain experts, making it ideal for assessing the proposed InfoMRI model's ability to address aleatoric uncertainty by modeling the distribution of segmentations. The standard deviation of the measurement noise $\sigma$ is set to $5 \times 10^{-2}$.

\subsubsection{Quantitative Results} \label{sec:seg-performance}
Table~\ref{tab:segmentation-2d} summarizes the segmentation performance of different methods. It is evident that LI-Net demonstrates an increasing advantage over U-Net for MRI segmentation when using the same sampling patterns. This is observed in the comparisons between LI-Net and TradPat-UNet-RS (both utilizing traditional sampling patterns) and between LOUPE-LI-Net and LOUPE-UNet-RS (both utilizing learned sampling patterns). This advantage can be attributed to LI-Net’s ability to enforce spatial coherence in segmentation, which significantly reduces the solution space and enables correct segmentation to be inferred from fewer measurements.

Additionally, Tackle outperforms SemuNet by a noticeable margin. Both methods are based on sequential models that combine a U-Net for reconstruction with a U-Net for segmentation from the reconstructed image. The primary distinction lies in their approach to the reconstruction stage: Tackle treats the reconstructed image as an intermediate feature without explicitly optimizing its reconstruction quality, whereas SemuNet optimizes the reconstructed image using the $L_1$ distance to the ground truth. This result suggests that deliberately optimizing MRI reconstruction may not necessarily lead to improved performance in MRI segmentation.

For the proposed InfoMRI, the GED performance consistently surpasses all baselines across all acceleration factors, demonstrating a superior match to the ground-truth distribution compared to the baselines. Moreover, InfoMRI exhibits an increasing advantage over the baselines and experiences the smallest performance drop as the acceleration factor increases (in both GED and DSC metrics). This indicates that the proposed method effectively handles information loss in MR subsampling, producing more reliable and stable predictions.

\subsubsection{Segmentation Under Uncertainty}
Due to the information lost in MR subsampling, there is a considerable uncertainty while implementing segmentation.
\figurename~\ref{fig:seg-compare} demonstrates how different methods handle segmentation under an extremely low sampling rate ($32\times$ acceleration). As shown, \textit{pixel-wise prob.} methods (abbreviated as methods that model segmentation using pixel-wise probabilities) result in spatially incoherent segmentation when thresholding the pixel-wise probabilities. For LI-Net, segmentation performance is significantly better than pixel-wise probability methods due to its approach of decoding segmentation from latent representations, resulting in spatially consistent segmentation. However, segmentation errors remain non-negligible, and LI-Net is unable to provide uncertainty quantification due to its deterministic design. In contrast, the proposed method generates multiple segmentations (see the areas of varying shades of red color) by decoding latent representations from multiple samples of the latent posterior \( q(\mathbf{z|y}) \), thereby achieving spatially consistent segmentation while also providing uncertainty quantification. \figurename~\ref{fig:viz-uncertainty} further demonstrates that as the acceleration factor increases, the proposed method produces more diverse segmentations, indicating its ability to provide reasonable uncertainty quantification to address the information loss in MR subsampling.

\subsection{Privacy-Preserving Task-Adapted CS-MRI}

In this section, we implement privacy-preserving classification-adapted CS-MRI for the case where \( \beta \geq 0 \), with larger values of \( \beta \) indicating stronger suppression of MRI reconstruction. {Many clinical workflows aim to obtain specific diagnostic decisions (\emph{e.g.}, classification of conditions) rather than visually inspectable images. However, MR images inherently contain sensitive, patient-identifiable information. Sharing or processing these images, even for automated analysis, poses significant privacy risks, especially in the context of federated learning or cloud-based diagnostics. Our framework is able to learns a sampling and inference pipeline that is explicitly optimized to be \emph{sufficient} for the diagnostic task $T$ while being deliberately \emph{insufficient} for high-fidelity reconstruction of the image $X$. This creates a form of ``privacy-by-design'', where the acquired data is intrinsically difficult to reverse-engineer into a recognizable image, thus protecting patient privacy.}

\subsubsection{Experimental Setup}
The proposed InfoMRI is compared with traditional sampling strategies, including uniform random and variable density, as well as the deep learning-based approach, LOUPE and {A-DPS}. We utilize ResNet18~\cite{he2016deep1} as the classification backbone for all traditional sampling methods and LOUPE. All ResNet18 models are trained using the Adam optimizer with a learning rate of \(1 \times 10^{-4}\), while our PGN is still trained using the Adam optimizer with a learning rate of \(1 \times 10^{-2}\). {We use publicly available code of A-DPS with modifications to enable k-space sampling for fair comparisons.}  

To initially validate our method, we employ the MNIST dataset. This dataset is partitioned into training (50,000 images), validation (5,000 images), and test (5,000 images) subsets. Subsequently, we evaluate our approach in a more clinically relevant setting using the fastMRI knee dataset with slice-level labels provided by~\cite{zhao2021fastmri+}. We split the dataset into training (816 volumes), validation (176 volumes), and test (175 volumes) sets. Following~\cite{yenadaptive}, we predict the presence of Meniscal Tears and ACL sprains in each slice, introducing an additional "abnormal" category to encompass less frequent pathologies as described in~\cite{yenadaptive}. To compare task performance, we use accuracy for the MNIST dataset and the area under the receiver operator curve (AUROC) for the fastMRI dataset, given the imbalanced nature of the dataset. 
For MNIST and fastMRI dataset, we trained for 150 epochs with batch size of 512 and 16, respectively.

To assess the effectiveness of privacy protection, we evaluate the performance of reconstructing MR images from the k-space measurements. A lower PSNR implies a lower likelihood of successful reconstruction and indicates better privacy protection. For the fastMRI dataset, we adopt the same Reconstruction U-Net and training configurations described in Section~\ref{sec:exp-mri-reconstruction}. For the MNIST dataset, we construct a simple convolutional neural network (CNN) for reconstruction. The architecture of the CNN is \texttt{Conv2d(2,32,3,3)}-\texttt{ReLU}-\texttt{Conv2d(32,64,3,3)}-\texttt{ReLU}-\texttt{Conv2d(64,1,1,1)}, where \texttt{Conv2d(2,32,3,3)} denotes a convolutional layer with an input channel size of 2, an output channel size of 32, and a kernel size of 3$\times$3.
The CNN is trained to minimize the MSE loss with a batch size of 64 using the Adam optimizer with a learning rate of $1 \times 10^{-4}$. Training is terminated when the MSE on the validation set stops decreasing.

\begin{table}[!t]
\renewcommand{\baselinestretch}{1.0}
\renewcommand{\arraystretch}{1.0}
\centering
\caption{Privacy preserving classification on MNIST. Higher accuracy and lower PSNR indicate better classification and privacy preserving performance. We use \textbf{bold} to highlight the best and \underline{underline} for the second best.}\label{tab:mnist}
\centering
\begin{tabular}{@{}lcccc@{}}
\toprule
\multirow{2}{*}{Methods} & \multicolumn{2}{c}{$32\times$ Acceleration} & \multicolumn{2}{c}{$48\times$ Acceleration} \\
\cmidrule(lr){2-3}  \cmidrule(lr){4-5}
& PSNR~$\downarrow$ & Accuracy~$\uparrow$ & PSNR~$\downarrow$ & Accuracy~$\uparrow$ \\
\midrule
Uniform random           & \textbf{11.33} & 0.8236 & \underline{11.20} & 0.8061 \\
Variable density         & 14.05 & 0.9410 & 12.74 & 0.8838 \\
LOUPE  & 15.61 & 0.9692 & 13.73 & 0.9253 \\
{ A-DPS }   & { 15.10} & { 0.9773} & { 13.02} & { \textbf{0.9651}} \\
\midrule
InfoMRI ($\beta=0$)      & 14.84 & \underline{0.9776} & 12.19 & 0.9587 \\
InfoMRI ($\beta>0$)      & \underline{11.79} & \textbf{0.9801} & \textbf{10.25} & 
\underline{0.9622} \\
\bottomrule
\end{tabular}
\end{table}
\begin{table}[!t]
\centering
\caption{Privacy preserving diagnosis on fastMRI. Higher AUROC and lower PSNR indicate better classification and privacy preserving performance. We use \textbf{bold} and \underline{underline} to highlight the best and the second best, respectively.}\label{tab:fastMRI}
\begin{tabular}{@{}lcccc@{}}
\toprule
\multirow{2}{*}{Methods} & \multicolumn{2}{c}{$16\times$ Acceleration} & \multicolumn{2}{c}{$32\times$ Acceleration} \\
\cmidrule(lr){2-3}  \cmidrule(lr){4-5}
& PSNR~$\downarrow$ & AUROC~$\uparrow$ & PSNR~$\downarrow$ & AUROC~$\uparrow$ \\
\midrule
Uniform random    & \textbf{22.58} & 0.8968 & \textbf{19.63} & 0.8848 \\
Variable density  & 25.33 & 0.9079 & 23.81 & 0.8989 \\
LOUPE             & 26.20 & 0.9129 & 24.85 & 0.9064 \\
\midrule
InfoMRI ($\beta=0$) & 25.42 & \textbf{0.9206} & 24.00 & \textbf{0.9157} \\
InfoMRI ($\beta>0$) & \underline{24.41} & \underline{0.9152} & \underline{23.56} & \underline{0.9096} \\
\bottomrule
\end{tabular}
\end{table}

\subsubsection{Results}
Tables~\ref{tab:mnist} and~\ref{tab:fastMRI} summarize the reconstruction and task performance for all the methods. The optimal value of \(\beta > 0\) is selected via grid search. An interesting observation is that task performance and reconstruction performance may not be directly correlated, \emph{i.e.}, better reconstruction performance not necessarily leading to better task performance. For instance, Table~\ref{tab:mnist} shows that InfoMRI (\(\beta > 0\)) achieves significantly higher accuracy than variable density and LOUPE (by approximately 4\% and 2\%, respectively) while simultaneously exhibiting worse reconstruction performance (by around 2.3 dB and 3.8 dB, respectively). 

\subsection{Discussion and Limitation}
We provide additional experimental details and results in the supplementary material, including additional reconstruction visualizations on brain MRI data, extended experiments on the \textit{SKM-TEA} and \textit{BRISC 2025} datasets with qualitative analysis of fine-structure tumor segmentation, confidence calibration analysis, sensitivity analysis for hyperparameters, inference complexity analysis, training stability analysis, and performance degradation analysis across acceleration factors.

While the proposed InfoMRI framework demonstrates promising performance in task-adapted sampling and reconstruction, there exist several limitations in the current study. First, the experiments are primarily conducted on single-coil simulated data. Clinical MRI scanners typically use multi-coil acquisition (parallel imaging), which involves more complex forward models (e.g., sensitivity encoding) and noise characteristics. Extending the proposed information-theoretic optimization to multi-coil settings remains an important future direction. Second, our current methodology focuses on 2D slice-wise acquisition, whereas many clinical protocols employ 3D volumetric imaging. The computational complexity of posterior sampling and entropy estimation would increase significantly in 3D, requiring more efficient approximation techniques. Finally, although we demonstrate the utility of uncertainty maps for flagging difficult cases, integrating this ``human-in-the-loop'' mechanism into actual clinical workflows requires further validation with radiologists to assess its impact on diagnostic decision-making.

\section{Conclusions} \label{sec:conclu}

In this paper, we proposed the first adaptive task-adapted CS-MRI framework via ITM-based optimization that addresses the limitations of existing methods in supporting specific clinical tasks and managing diagnostic uncertainty. By maximizing the mutual information between k-space measurements and downstream tasks, our approach enables probabilistic inference and provides a comprehensive solution to the inherent uncertainty in medical diagnosis. Leveraging amortized optimization and constructing tractable variational bounds, we achieved adaptive CS-MRI that jointly optimizes sampling, reconstruction, and task-inference models, allowing flexible control of the sampling ratio within a single end-to-end trained model. Furthermore, our framework unifies two distinct clinical scenarios: enhancing task performance through joint reconstruction and task inference, and implementing tasks with suppressed reconstruction for privacy protection. Extensive experimental results demonstrate the state-of-the-art performance of the proposed framework in terms of reconstruction quality and the accuracy of various clinical tasks.

\newpage

\appendices

\section{Additional Experimental Results}

\subsection{Additional Results on CS-MRI Reconstruction}

\begin{figure*}
    \centering
    \includegraphics[width=1\linewidth]{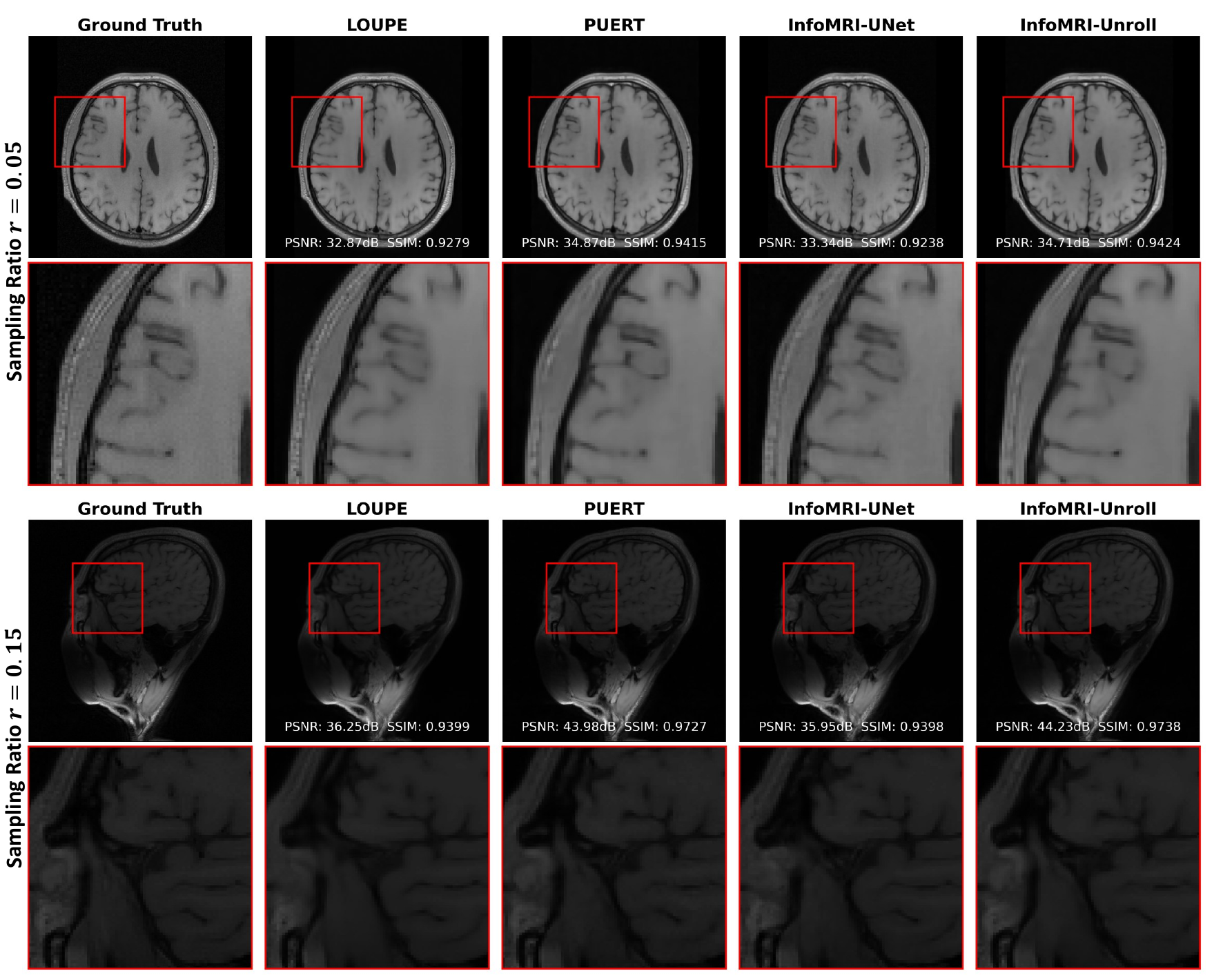}
    \caption{
    Visualizations of MR reconstruction images at the 5\% and 15\% sampling rates on the \textit{Brain MRI dataset}. The leftmost column shows the ground truth MR images (top) and the amplified regions (bottom). The other columns display the reconstruction images and the amplified areas. The PSNR and SSIM values are labeled in the reconstruction images. }
    \label{fig:brain-viz}
\end{figure*}

To evaluate our method on complex, fine-structured anatomical regions, we show visual reconstruction results on brain MRI data in Figure~\ref{fig:brain-viz}.

We further evaluate the Bjøntegaard Delta (BD)-PSNR and BD-Rate of the baselines in Table~\ref{tab:bd-new}, using the proposed method as the benchmark. Table~\ref{tab:bd-new} shows that our method outperforms all baselines except for LOUPE in the \textit{2D unconstrained sampling} scenario on the \textit{fastMRI dataset}.

\subsection{Additional Results on Task-adapted CS-MRI}

\subsubsection{More Results} 

As presented in Table~\ref{tab:2x4x}, at the $2\times$ and $4\times$ acceleration rates, InfoMRI achieves the lowest GED (0.0940 and 0.0943, respectively). This indicates that its predictions provide the closest distributional match to the ground-truth segmentation posterior among all evaluated methods. 

To further demonstrate the clinical practicality of the proposed method, we have further evaluated the performance on 
\begin{itemize}
    \item \textit{SKM-TEA dataset~\cite{desai2021skm}}: A large-scale benchmark of quantitative knee MRI (qMRI) scans with dense manual segmentations, designed for the joint evaluation of MRI reconstruction and analysis tasks.
    
    \item \textit{BRISC 2025 dataset~\cite{fateh2026brisc}}: A clinical benchmark providing expert-annotated brain tumor segmentations from realistic MRI scans, targeting robust segmentation performance in clinical settings.
\end{itemize}

The quantitative results are presented in Table~\ref{tab:skm-tea} and Table~\ref{tab:segmentation-brisc2025}, respectively. The proposed method consistently outperforms prior methods on GED metrics and obtains the best DSC scores for high acceleration factors.

\begin{figure*}
    \centering
    \includegraphics[width=1\linewidth]{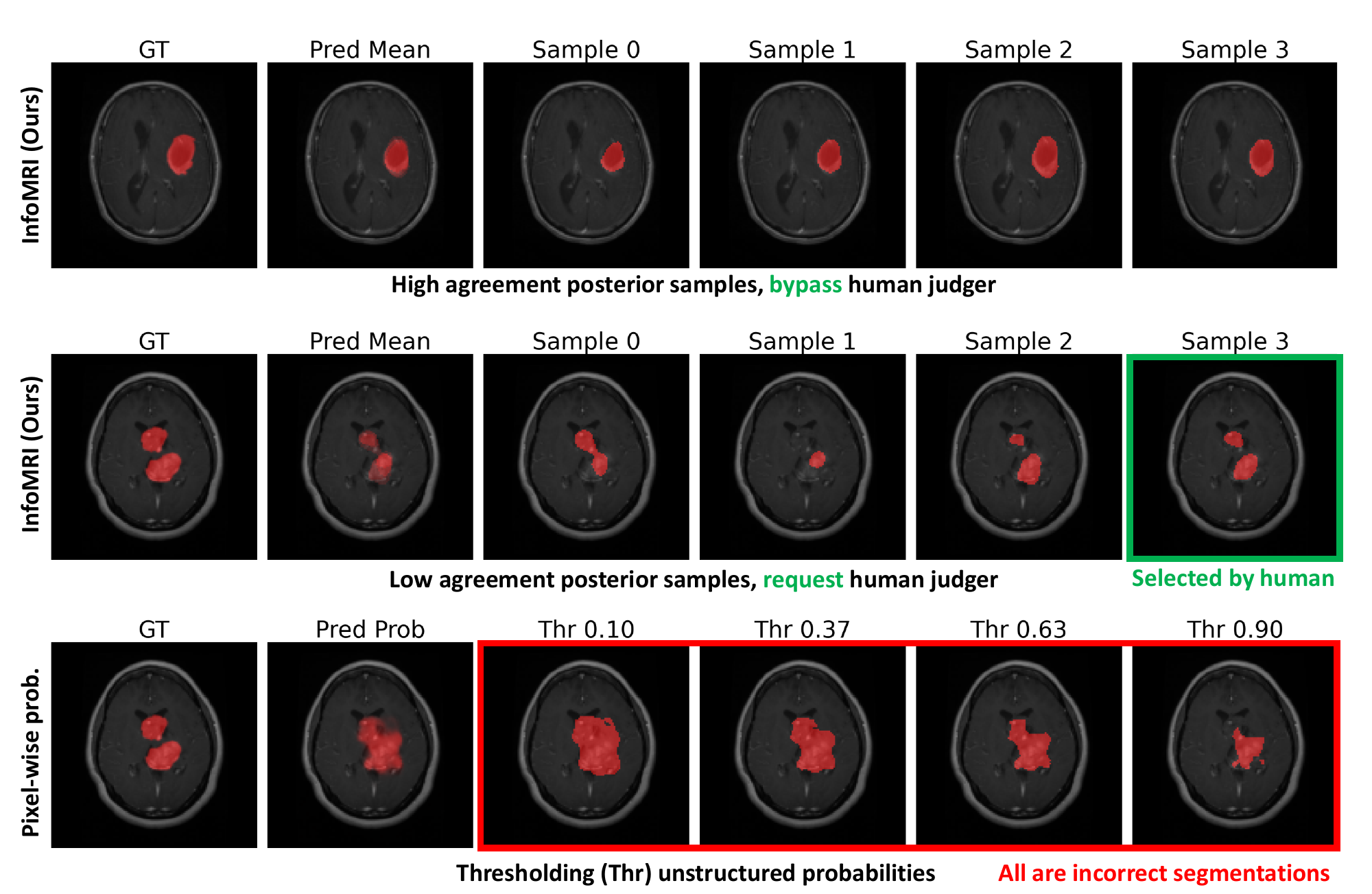}
    \caption{Demonstration of practical utility evaluated on the challenging real-world clinical \textit{BRISC 2025 dataset}. \textbf{Row 1:} A standard clinical case where InfoMRI posterior samples exhibit high consistency, indicating reliable segmentation. \textbf{Rows 2 and 3:} A shared challenging case featuring a small and ambiguous lesion (sharing the same Ground Truth). \textbf{Row 2:} Under high ambiguity, InfoMRI generates diverse, structurally coherent posterior samples. The high inter-sample variance explicitly indicates low confidence, serving as a reliable built-in flag to prompt human expert intervention. \textbf{Row 3:} In contrast, deterministic pixel-wise probability maps fail to capture this diagnostic uncertainty, yielding consistently incorrect segmentation boundaries regardless of the chosen threshold.}
    \label{fig:brisc-viz1}
\end{figure*}

\begin{figure*}
    \centering
    \includegraphics[width=1\linewidth]{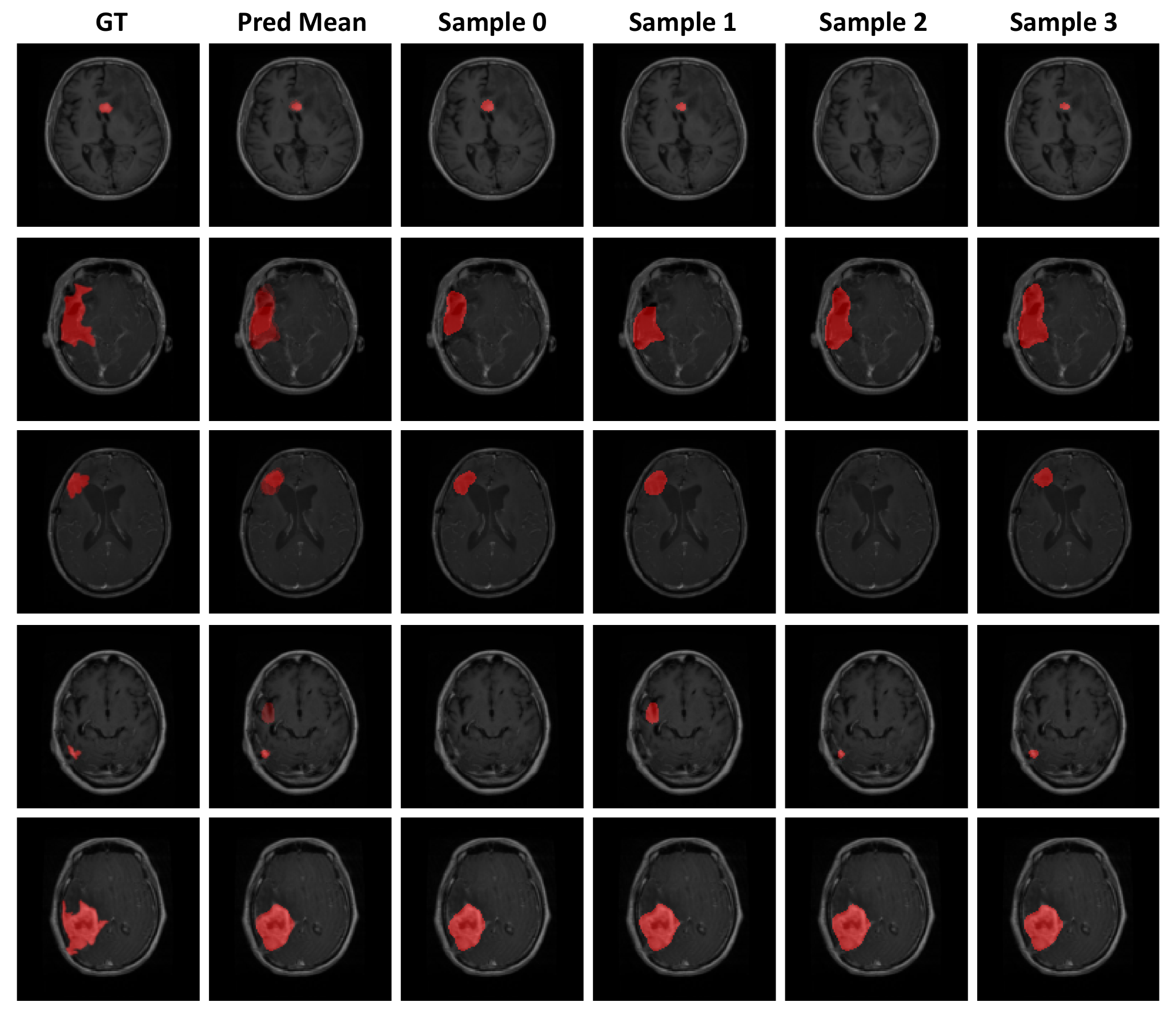}
    \caption{Visual examples of fine-structure tumor segmentation at $16\times$ acceleration on the clinical BRISC 2025 dataset. The proposed method demonstrates a superior capability to accurately delineate intricate lesion boundaries and capture diverse, irregular tumor shapes, thereby maintaining strict structural integrity under significant undersampling.}
    \label{fig:brisc-viz2}
\end{figure*}

Figures~\ref{fig:brisc-viz1} and \ref{fig:brisc-viz2} visualize our fine-structure segmentation capabilities. Unlike standard deterministic models that yield overconfident and often incorrect pixel-wise probability maps under high uncertainty, InfoMRI generates diverse, structurally coherent posterior samples. This enables a highly practical clinical workflow: cases with high sample agreement (Fig.~\ref{fig:brisc-viz1}, Row 1) indicate reliable segmentations that can bypass manual review; conversely, cases with high inter-sample disagreement (Fig.~\ref{fig:brisc-viz1}, Row 2) serve as a built-in uncertainty flag, successfully highlighting ambiguous fine-structure boundaries for human expert (radiologist) review to select the correct sample. In contrast, deterministic models rely on pixel-wise probability maps that yield varying segmentation boundaries depending on the chosen threshold. As demonstrated in Row 3 of Fig.~\ref{fig:brisc-viz1}, under high ambiguity, all such threshold-dependent results remain fundamentally incorrect. Furthermore, Figure~\ref{fig:brisc-viz2} demonstrates that our method robustly captures fine, irregular tumor structures across diverse patient subjects.

\begin{table}[!t]
\renewcommand{\baselinestretch}{1.0}
\renewcommand{\arraystretch}{1.0}
\centering
\caption{BD-PSNR gains (dB) and BD-Rate changes (\%) for the baselines compared with InfoMRI on the \textit{fastMRI dataset}. The results of LOUPE and SeqMRI are averaged over the four sampling ratios of 6.25\%, 12.5\%, 18.75\%, and 25\%, whereas those of other methods are calculated using all sampling ratios within the range [6.25\%, 25\%] (note that there are a total of 100 equidistantly sampled points within the sampling rate range of [1\%, 30\%]). A negative BD-PSNR indicates that the baseline provides lower quality than InfoMRI at the same sampling rate, while a positive BD-Rate means the baseline requires higher sampling rate than InfoMRI to achieve the same quality.} \label{tab:bd-new}
\begin{tabular}{@{}lcccc@{}}
\toprule
\multirow{2}{*}{Method} & \multicolumn{2}{c}{\textit{2D Unconstrained}} & \multicolumn{2}{c}{\textit{1D Cartesian}} \\
\cmidrule(lr){2-3}  \cmidrule(lr){4-5}
& BD-PSNR & BD-Rate & BD-PSNR & BD-Rate \\
\midrule
Uniform random      & -4.22 & 38.97\% & -3.96 & 32.64\% \\
Variable density    & -1.72 & 26.45\%  & -    & - \\
Spectrum based & -2.10 & 30.17\%  & -    & - \\
Equispace         & -     & -      & -3.06   & 26.06\% \\
LOUPE                        & 0.03  & -0.017\%   & -0.89 & 14.22\% \\
SeqMRI                         & -1.48 & 17.66\%    & -3.06 & 25.61\% \\
PGMRI                         & -     & -      & -4.20 & 28.96\% \\
\bottomrule
\end{tabular}
\end{table}

\begin{table}[!t]
\centering
\caption{Segmentation performance comparisons on \textit{QUBIQ 2021 dataset}. Lower GED and Higher DSC indicate better segmentation performance. We use \textbf{bold} and \underline{underline} to highlight the best and the second best, respectively.} \label{tab:2x4x}
\setlength{\tabcolsep}{2.5pt}
\begin{tabular}{@{}llcccc@{}}
\toprule
\multirow{2}{*}{Method} & \multirow{2}{*}{Sampling Pattern}  &  \multicolumn{2}{c}{$2\times$ Acceleration} &  \multicolumn{2}{c}{$4\times$ Acceleration}  \\
\cmidrule(lr){3-4}  \cmidrule(lr){5-6} 
& & GED~$\downarrow$ & DSC~$\uparrow$ & GED~$\downarrow$ & DSC~$\uparrow$ \\
\midrule
\multirow{3}{*}{TradPat-UNet-RS} & Uniform random   &  0.1372 & 0.9548 & 0.1666 & 0.9431 \\
& Poisson disk                                      &  0.1190 & 0.9572 & 0.1266 & 0.9563 \\
& Variable density                                  & 0.1300 & 0.9529 & 0.2029 & 0.9266 \\
\midrule
\multirow{3}{*}{LI-Net} & Uniform random    &  0.1688 & 0.9393 & 0.1868 & 0.9328  \\
& Poisson disk                              &  0.1341 & 0.9520 & 0.1347 & 0.9520  \\
& Variable density                          &  0.1411 & 0.9510 & 0.1704 & 0.9421 \\
\midrule
LOUPE-UNet-RS & \multirow{5}{*}{Learned} &  0.1150 & 0.9609 & 0.1117 & 0.9618 \\
LOUPE-LI-Net &                           &  0.1187 & \underline{0.9568} & 0.1211 & 0.9561 \\
SemuNet &                                &  0.1176 & 0.9565 & 0.1121 & \underline{0.9598} \\
Tackle &                                 &  \underline{0.1104} & \textbf{0.9618} & \underline{0.1073} & \textbf{0.9624} \\
InfoMRI (Ours) &                         & \textbf{0.0940} & 0.9557 & \textbf{0.0943} & 0.9546 \\
\bottomrule
\end{tabular}
\end{table}

\begin{table*}
\centering
\caption{Segmentation performance comparisons on \textit{SKM-TEA dataset}. Lower GED and Higher DSC indicate better segmentation performance. We use \textbf{bold} and \underline{underline} to highlight the best and the second best, respectively. }\label{tab:skm-tea}
\begin{tabular}{@{}llcccccccc@{}}
\toprule
\multirow{2}{*}{Method} & \multirow{2}{*}{Sampling Pattern} & \multicolumn{2}{c}{$2\times$ Acceleration} & \multicolumn{2}{c}{$4\times$ Acceleration} &  \multicolumn{2}{c}{$8\times$ Acceleration} & \multicolumn{2}{c}{$16\times$ Acceleration} \\
\cmidrule(lr){3-4}  \cmidrule(lr){5-6} \cmidrule(lr){7-8} \cmidrule(lr){9-10}
& & GED~$\downarrow$ & DSC~$\uparrow$ & GED~$\downarrow$ & DSC~$\uparrow$ & GED~$\downarrow$ & DSC~$\uparrow$ & GED~$\downarrow$ & DSC~$\uparrow$ \\
\midrule
\multirow{3}{*}{TradPat-UNet-RS} & Uniform random   & 0.3965 & 0.4935 & 0.4028 & 0.4887 & 0.4368 & 0.4742 & 0.5058 & 0.4416 \\
& Poisson disk                                      & 0.3835 & 0.4996 & 0.3859 & 0.4966 & 0.4126 & 0.4863 & 0.4503 & 0.4679 \\
& Variable density                                  & 0.3913 & 0.4944 & 0.3932 & 0.4927 & 0.4236 & 0.4803 & 0.4768 & 0.4571 \\
\midrule
\multirow{3}{*}{LI-Net} & Uniform random            & 0.5038 & 0.4529 & 0.5096 & 0.4482 & 0.5433 & 0.4331 & 0.6283 & 0.3927 \\
& Poisson disk                                      & 0.5008 & 0.4524 & 0.5074 & 0.4489 & 0.5115 & 0.4478 & 0.5429 & 0.4332 \\
& Variable density                                  & 0.5088 & 0.4496 & 0.5074 & 0.4499 & 0.5147 & 0.4464 & 0.5399 & 0.4343 \\
\midrule
LOUPE-LI-Net &   \multirow{5}{*}{Learned}           & 0.4771 & 0.4632 & 0.4792 & 0.4620 & 0.4886 & 0.4573 & 0.5179 & 0.4448 \\
LOUPE-UNet-RS &                                     & 0.3903 & 0.4968 & 0.3887 & 0.4961 & 0.4136 & 0.4866 & 0.4544 & 0.4686 \\
FSL    &                                            & 0.3712 & 0.5035 & 0.3721 & \underline{0.5024} & 0.3930 & 0.4932 & 0.4330 & 0.4756 \\
SemuNet &                                           & 0.3741 & \underline{0.5026} & 0.3749 & 0.5015 & 0.3943 & 0.4928 & 0.4345 & 0.4755 \\
Tackle &                                            & \underline{0.3645} & \textbf{0.5048} & \underline{0.3688} & \textbf{0.5033} & \underline{0.3916} & \underline{0.4936} & \underline{0.4321} & \underline{0.4757} \\
InfoMRI (Ours) &                                    & \textbf{0.3457} & 0.5013 & \textbf{0.3448} & 0.5011 & \textbf{0.3424} & \textbf{0.5032} & \textbf{0.3509} & \textbf{0.4949}  \\
\bottomrule
\end{tabular}
\end{table*}

\begin{table*}[!t]
\centering
\caption{ Segmentation performance comparisons on \textit{BRISC 2025 dataset}. Lower GED and higher DSC indicate better performance. We use \textbf{bold} and \underline{underline} to highlight the best and the second best, respectively.} \label{tab:segmentation-brisc2025}
\begin{tabular}{@{}llcccccccc@{}}
\toprule
\multirow{2}{*}{Method} & \multirow{2}{*}{Sampling Patterns} & \multicolumn{2}{c}{$8\times$ Acceleration} & \multicolumn{2}{c}{$16\times$ Acceleration} & \multicolumn{2}{c}{$24\times$ Acceleration} & \multicolumn{2}{c}{$32\times$ Acceleration} \\
\cmidrule(lr){3-4} \cmidrule(lr){5-6} \cmidrule(lr){7-8} \cmidrule(lr){9-10}
 &  & GED~$\downarrow$ & DSC~$\uparrow$ & GED~$\downarrow$ & DSC~$\uparrow$ & GED~$\downarrow$ & DSC~$\uparrow$ & GED~$\downarrow$ & DSC~$\uparrow$ \\
\midrule
\multirow{3}{*}{TradPat-UNet-RS} & Uniform random   & 0.9109 & 0.6423 & 1.1624 & 0.5132 & 1.4424 & 0.3612 & 1.5017 & 0.3263 \\
& Poisson disk & 0.7598 & 0.7099 & 0.9390 & 0.6237 & 1.1742 & 0.5018 & 1.1890 & 0.4956 \\
& Variable density & 0.6085 & 0.7816 & 0.7307 & 0.7299 & 0.8449 & 0.6770 & 0.9354 & 0.6300 \\
\midrule
\multirow{3}{*}{LI-Net} & Uniform random & 1.0508 & 0.5793 & 1.2417 & 0.4783 & 1.4856 & 0.3409 & 1.5214 & 0.3204 \\
& Poisson disk & 0.9039 & 0.6405 & 1.0580 & 0.5644 & 1.2976 & 0.4347 & 1.3174 & 0.4247 \\
& Variable density & 0.7374 & 0.7284 & 0.8590 & 0.6656 & 0.9584 & 0.6164 & 1.0332 & 0.5785 \\
\midrule
LOUPE-LI-Net & \multirow{5}{*}{Learned} & 0.6473 & 0.7735 & 0.7185 & 0.7389 & 0.7933 & 0.7053 & 0.9290 & 0.6333 \\
LOUPE & & 0.5433 & \underline{0.8091} & \underline{0.6349} & \textbf{0.7733} & 0.7290 & 0.7319 & 0.8583 & 0.6680 \\
SemuNet & & \underline{0.5349} & \textbf{0.8141} & 0.6562 & 0.7638 & \underline{0.7283} & \underline{0.7336} & \underline{0.7881} & \underline{0.7071} \\
Tackle & & 0.6024 & 0.7844 & 0.7303 & 0.7280 & 0.8298 & 0.6839 & 0.9103 & 0.6441 \\
InfoMRI (Ours) & & \textbf{0.4993} & 0.7867 & \textbf{0.5254} & \underline{0.7649} & \textbf{0.5507} & \textbf{0.7472} & \textbf{0.5677} & \textbf{0.7280} \\
\bottomrule
\end{tabular}
\end{table*}

\subsubsection{Confidence Calibration} 
To evaluate whether the methods provide accurate uncertainty estimation, we rely on confidence calibration metrics, Expected Calibration Error (ECE) and Brier Score (BS), for pixel-wise probabilities. Specifically, for a binary segmentation task, given the ground truth segmentation $t_i$ and predicted pixel-wise probabilities $q(t_i = 1)$ for each pixel $i = 1, \ldots, d$, the ECE and BS are defined as:
\begin{align}
    & ECE = \sum_{n=1}^N \frac{|D_n|}{d} |\mathrm{acc}(D_n) - \mathrm{conf}(D_n)|, \\
    & BS = \frac{1}{d} \sum_{i=1}^d (q(t_i = 1) - t_i)^2,
\end{align}
where $d$ is the dimension of segmentation map, $N$ is the number of bins, $D_n$ is the set of pixel indices $i$ whose predicted probabilities $q(t_i = 1)$ fall within the range $\left(\frac{n-1}{N}, \frac{n}{N}\right]$. The terms $\mathrm{acc}(D_n)$ and $\mathrm{conf}(D_n)$ represent the averaged segmentation accuracy and averaged predicted probabilities $q(t_i = 1)$ for pixels with indices in $D_n$, respectively. We set $N = 16$ in the experiments. For the proposed method, pixel-wise probabilities are obtained by averaging segmentation samples from $q(\mathbf{t|y})$.

Table~\ref{tab:calibration} presents the confidence calibration performance of different methods. Notably, although the proposed method does not directly optimize for pixel-wise probabilities, the pixel-wise probabilities obtained by averaging samples from \( q(\mathbf{t|y}) \) exhibit comparable or even superior performance on metrics for confidence calibration. This demonstrates that the uncertainty quantification provided by the proposed method is not only qualitatively reasonable but also quantitatively accurate.

\begin{table*}[!t]
\renewcommand{\baselinestretch}{1.0}
\renewcommand{\arraystretch}{1.0}
\centering
\caption{Calibration performance comparisons on \textit{QUBIQ 2021 dataset}. Lower BS and ECE indicate better calibration performance. We use \textbf{bold} and \underline{underline} to highlight the best and the second best, respectively. Note that $8\times$, $16\times$, $24\times$, and $32\times$ accelerations correspond to sampling ratios of 1/8, 1/16, 1/24 and 1/32, respectively.} \label{tab:calibration}
\begin{tabular}{@{}llcccccccc@{}}
\toprule
\multirow{2}{*}{Method} & \multirow{2}{*}{Sampling patterns} &  \multicolumn{2}{c}{$8\times$ Acceleration} &  \multicolumn{2}{c}{$16\times$ Acceleration} & \multicolumn{2}{c}{$24\times$ Acceleration} &  \multicolumn{2}{c}{$32\times$ Acceleration}  \\
\cmidrule(lr){3-4}  \cmidrule(lr){5-6}  \cmidrule(lr){7-8}  \cmidrule(lr){9-10}
 &  & ECE~$\downarrow$ & BS~$\downarrow$ & ECE~$\downarrow$ & BS~$\downarrow$ & ECE~$\downarrow$ & BS~$\downarrow$ & ECE~$\downarrow$ & BS~$\downarrow$  \\
\midrule
TradPat-UNet-RS & Variable density & 0.0121 & 0.0307 & 0.0204 & 0.0425 & 0.0191 & 0.0505 & \underline{0.0318} & 0.0597 \\
\midrule
LI-Net & Variable density & 0.0228 & 0.0266 & 0.0298 & 0.0337 & 0.0462 & 0.0499 & 0.0415 & 0.0459 \\
\midrule
LOUPE-UNet-RS & \multirow{5}{*}{Learned} & \underline{0.0083} & 0.0223 & \underline{0.0129} & 0.0344 & 0.0156 & 0.0397 & 0.0364 & 0.0681 \\
LOUPE-LI-Net  & & 0.0299 & 0.0299 & 0.0319 & 0.0319 & 0.0407 & 0.0407 & 0.0532 & 0.0532 \\
SemuNet & & 0.0141 & \underline{0.0205} & 0.0178 & \textbf{0.0237} & \underline{0.0155} & \textbf{0.0302} & 0.0223 & \underline{0.0368} \\
Tackle                        & & \textbf{0.0081}    & \textbf{0.0176} & 0.0148 & 0.0255 & 0.0251 & \underline{0.0309} & 0.0330 & 0.0421 \\
InfoMRI (Ours)                                      &                          & 0.0098             & 0.0275          & \textbf{0.0127} & \underline{0.0290} & \textbf{0.0150} & 0.0310 & \textbf{0.0140} & \textbf{0.0316}  \\
\bottomrule
\end{tabular}
\end{table*}

\begin{table*}
\centering
\caption{Sensitivity analysis on hyperparameters of the loss function evaluated on \textit{QUBIQ 2021 dataset}.} \label{tab:test-w123}
\begin{tabular}{@{}ccccccccccc@{}}
\toprule
\multirow{2}{*}{$w_1$} & \multirow{2}{*}{$w_2$} & \multirow{2}{*}{$w_3$} & \multicolumn{2}{c}{$4\times$ Acceleration} &  \multicolumn{2}{c}{$8\times$ Acceleration} & \multicolumn{2}{c}{$16\times$ Acceleration} & \multicolumn{2}{c}{$24\times$ Acceleration} \\
\cmidrule(lr){4-5}  \cmidrule(lr){6-7} \cmidrule(lr){8-9}   \cmidrule(lr){10-11} 
& & & GED~$\downarrow$ & DSC~$\uparrow$ & GED~$\downarrow$ & DSC~$\uparrow$ & GED~$\downarrow$ & DSC~$\uparrow$ & GED~$\downarrow$ & DSC~$\uparrow$ \\
\midrule
1.0 & 1.0 & 1.0 & 0.7087 & 0.3466 & 0.6938 & 0.3053 & 0.6372 & 0.4541 & 0.6489 & 0.4356 \\
1.0 & 5.0 & 1.0 & 0.5884 & 0.4633 & 0.7075 & 0.4463 & 0.6224 & 0.4880 & 0.5750 & 0.5480 \\
1.0 & 10.0 & 1.0 & 0.2388 & 0.8673 & 0.2235 & 0.8826 & 0.2589 & 0.8674 & 0.2352 & 0.8855 \\
1.0 & 50.0 & 1.0 & 0.0943 & 0.9546 & 0.0955 & 0.9542 & 0.1054 & 0.9472 & 0.1208 & 0.9380 \\
1.0 & 100.0 & 1.0 & 0.0953 & 0.9558 & 0.0900 & 0.9574 & 0.1397 & 0.9199 & 0.2032 & 0.8981 \\
\midrule
1.0 & 1.0 & 0.0 & 0.0997 & 0.9540 & 0.0926 & 0.9557 & 0.1413 & 0.9327 & 0.1384 & 0.9391 \\
1.0 & 1.0 & 0.1 & 0.0961 & 0.9567 & 0.0944 & 0.9561 & 0.1128 & 0.9480 & 0.1227 & 0.9448 \\
1.0 & 1.0 & 0.5 & 0.0882 & 0.9565 & 0.0983 & 0.9457 & 0.0946 & 0.9529 & 0.1037 & 0.9534 \\
1.0 & 1.0 & 1.0 & 0.0943 & 0.9546 & 0.0955 & 0.9542 & 0.1054 & 0.9472 & 0.1208 & 0.9380 \\
1.0 & 1.0 & 10.0 & 0.1088 & 0.9507 & 0.1099 & 0.9490 & 0.1367 & 0.9375 & 0.1241 & 0.9437 \\
\bottomrule
\end{tabular}
\end{table*}

\subsubsection{Visualization of Privacy-Protected Compressed Learning}

\begin{figure}[!t]
\renewcommand{\baselinestretch}{1.0}
\centering
\includegraphics[width=0.95\textwidth]{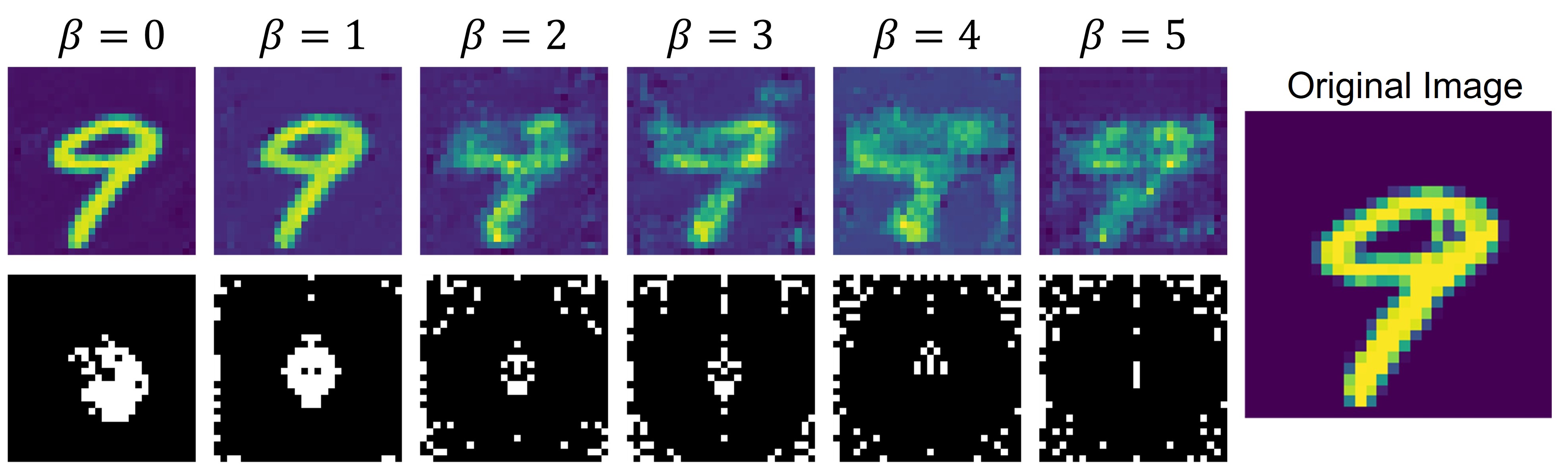}
\caption{Visualization of reconstruction examples under different $\beta$ values  on MNIST. As observed, reconstruction becomes progressively more challenging as $\beta$ increases. Top: Reconstruction; Bottom: sampling patterns.}
\label{fig:ibviz-new}
\end{figure}

We visualize the privacy protection performance in Fig.~\ref{fig:ibviz-new}. When the value of $\beta$ increases, the reconstruction from measurement gradually becomes difficult, and thereby the privacy is better preserved.

\subsection{Ablation Study}

\subsubsection{Sensitivity analysis for hyperparameters}
We have conducted a sensitivity analysis for the key hyperparameters, particularly the weighting parameters $w_1, w_2, w_3$ in our loss function (Eq.~\ref{eq:re-naive-obj-exp}), to assess their impact on final performance. 
\begin{align}\label{eq:re-naive-obj-exp}
\max_{\theta,\phi} \mathbb{E}_{r, \mathcal{M}, \mathbf{m}, \mathbf{x}, \mathbf{t}, \mathbf{y}} \mathbb{E}_{\hat{q}(\mathbf{z|t,y})} [w_{1}\log q(\mathbf{t}|\mathbf{z}, \mathbf{y})  - w_{2} D_{KL}[\hat{q}(\mathbf{z|t,y})||q(\mathbf{z|y})] + w_{3} \log q(\mathbf{x|y})],
\end{align}
Specifically, we varied $w_2$ and $w_3$ while keeping $w_1$ fixed at 1.0. Varying $w_2$ probes the balance between the KL regularizer $D_{KL}[\hat{q}(\mathbf{z}|\mathbf{t},\mathbf{y})||q(\mathbf{z}|\mathbf{y})]$ and the task NLL, while $w_3$ controls the trade-off between reconstruction $\log q(\mathbf{x}|\mathbf{y})$ and task performance. Results are summarized in Table~\ref{tab:test-w123}. 

As shown in Table~\ref{tab:test-w123}, a clear trend emerges as we vary $w_2$. Increasing $w_2$ from 1 to 50 consistently improves both GED and DSC across all acceleration factors. However, when $w_2$ is increased further to 100, performance at $16\times$ and $24\times$ acceleration exhibits a slight degradation, suggesting over-regularization. This indicates a robust operational range for $w_2$ is approximately $10 \sim50$. With $w_2$ fixed, we observe that a moderate reconstruction weight ($w_3 \in [0, 1]$) maintains or slightly improves training stability and performance. Conversely, a large $w_3=10$ significantly harms the task-specific metrics, as the optimization objective shifts focus toward pixel fidelity  at the expense of the downstream task. This suggests a robust range for $w_3$ is $0.1 \sim 1.0$.

The best trade-off between segmentation accuracy and uncertainty calibration is thus achieved with a strong KL regularization term (via $w_2$) and a moderate reconstruction term (via $w_3$). These trends remained consistent across all acceleration rates. Based on this analysis, we selected the default weights ($w_1=1$, $w_2=50$, $w_3=1.0$) for all segmentation experiments (QUBIQ and SKM-TEA). The fact that our method achieves strong performance on both datasets with these fixed hyperparameters indicates the robustness of our proposed framework.

\subsubsection{Analysis on Inference Complexity}
The inference process for a single pass is feed-forward and efficient. The primary computational cost arises from our method's ability to draw multiple stochastic samples at test time, which introduces a linear increase in computation. 
As summarized in the Table~\ref{tab:n_samples}, increasing the number of stochastic samples systematically improves the GED and stabilizes the Dice score, with diminishing returns observed beyond 8-16 samples. The inference time on an RTX 4090 GPU scales nearly linearly, from 0.006s for one sample to 0.075s for 16 samples. This provides a clinically relevant, tunable accuracy-latency trade-off. Compared to single-shot baselines like LOUPE-UNet-RS and SemuNet, InfoMRI achieves markedly better uncertainty calibration (GED) with only a modest increase in latency (e.g., using 4-8 samples), a trade-off we believe is acceptable for many clinical workflows.

\begin{table}
\centering
\caption{Performance and inference time trade-offs evaluated on \textit{QUBIQ 2021 dataset}. Performance results are based on subsampling with $16\times$ Acceleration. Inference times are reported based on one NVIDIA GeForce RTX 4090.} \label{tab:n_samples}
\setlength{\tabcolsep}{2.5pt}
\begin{tabular}{@{}cccccc@{}}
\toprule
Method & Number of Samples & GED & DSC & Infer. Time (s) \\
\midrule
\multirow{6}{*}{InfoMRI (Ours)} & 1  & 0.1369 & 0.9547 & 0.006 \\
& 2  & 0.1146 & 0.9535 & 0.011 \\
& 4  & 0.0936 & 0.9576 & 0.019 \\
& 6  & 0.0955 & 0.9542 & 0.029 \\
& 8  & 0.0869 & 0.9562 & 0.038 \\
& 16 & 0.0872 & 0.9563 & 0.075 \\
\midrule
LOUPE-UNet-RS & 1 & 0.2145 & 0.9251 & 0.004 \\
SemuNet       & 1 & 0.1296 & 0.9556 & 0.004 \\
Tackle        & 1 & 0.1243 & 0.9570 & 0.004 \\
\bottomrule
\end{tabular}
\end{table}

\subsection{Training Stability}
A potential challenge in optimizing information-theoretic objectives, such as the one in our framework, lies in the stability of the mutual information (MI) estimator. Both bias in high-dimensional settings and high variance during optimization can impede effective training. Our framework, InfoMRI, is architected with specific design choices to address these issues and ensure a stable and reliable training process.

\begin{figure*}[!t]
    \centering
    \includegraphics[width=0.95\textwidth]{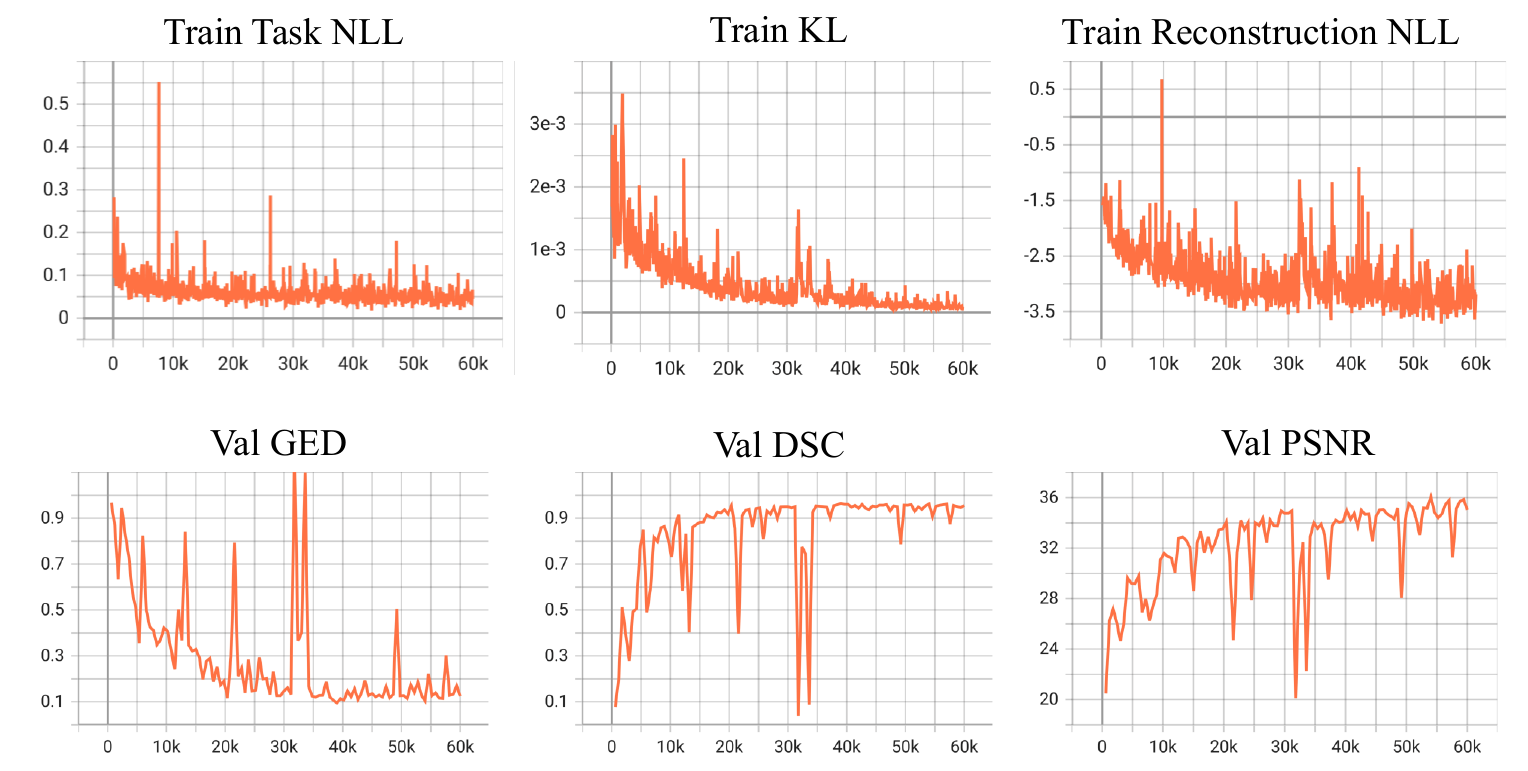}
    \caption{Training dynamics of InfoMRI. We report the evolution of three sub-losses (Task NLL, KL Divergence, and Reconstruction NLL) and the performance metrics on the validation dataset against the number of training steps. The smooth convergence demonstrates the stability of the optimization process.}
    \label{fig:loss}
\end{figure*}

To empirically validate the stability of our training process, we present the training dynamics of InfoMRI in Fig.~\ref{fig:loss}. These curves demonstrate that the three sub-losses—the task NLL ($\log q(\mathbf{t|z,y})$), the KL divergence ($D_{KL}[\hat{q}(\mathbf{z|t,y})||q(\mathbf{z|y})]$), and the reconstruction NLL ($\log q(\mathbf{x|y})$)—all converge smoothly as the number of training steps increases to 60k. Simultaneously, key performance metrics on the validation set (e.g., DSC, PSNR) improve and stabilize as expected. This provides direct empirical evidence that our training process is reliable and not compromised by estimator instability.

\subsubsection{Analysis on Performance vs. Acceleration}
\begin{figure*}
    \centering
    \includegraphics[width=0.8\linewidth]{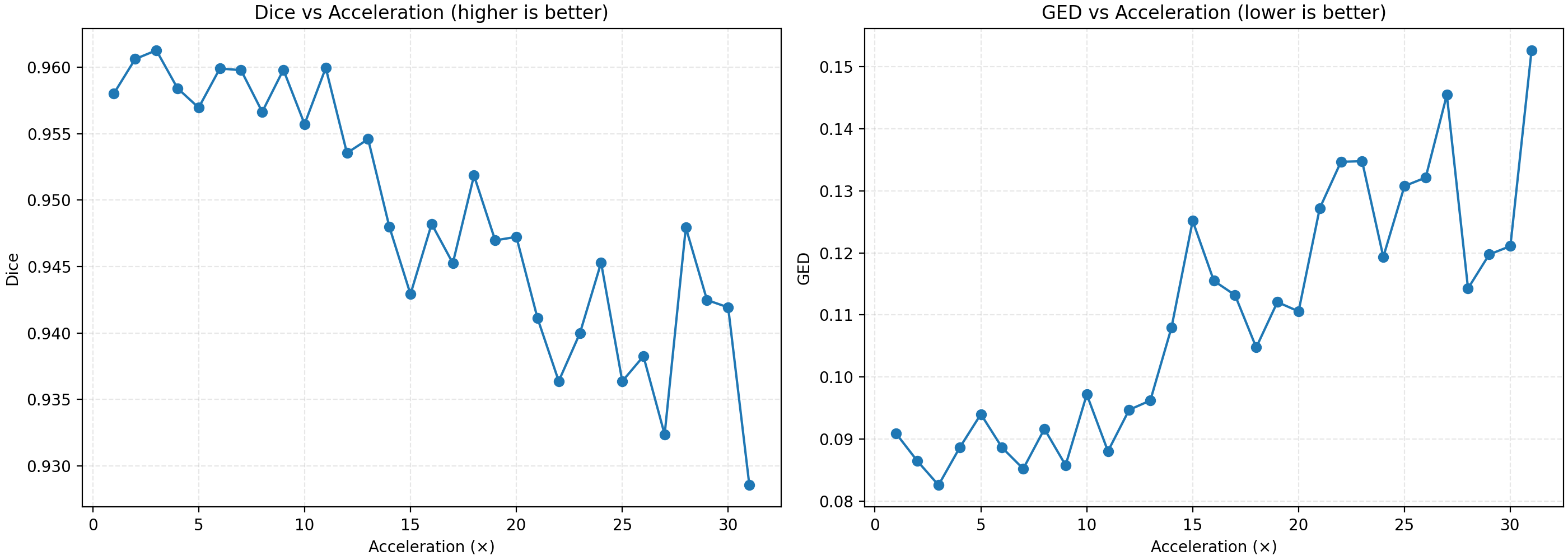}
    \caption{The GED and Dice performance on the \textit{QUBIQ 2021 dataset} for InfoMRI trained with the sampling ratios sampled from the full range [0,1]. The experimental configurations are the same as Table II in the manuscript.}
    \label{fig:performance-vs-accel}
\end{figure*}

We systematically vary the sampling rate and evaluate task performance to identify the ``cliff" where performance begins to degrade significantly compared to the fully-sampled case.
Specifically, we train an InfoMRI model on QUBIQ 2021 dataset with the sampling ratios sampled from the full range [0,1] and evaluate the GED and Dice performance from 1x acceleration (full sample) to 32x acceleration. The experimental configurations are the same as Table II in the manuscript.  As shown in Fig.~\ref{fig:performance-vs-accel}, the performance drop is negligible around 1$\times$ - 10$\times$ acceleration, indicating robust performance across a wide range of acceleration factors.

\section{Additional Experimental Details}

\textbf{* Implementing segmentation after MRI reconstruction:}

\textit{(1) SemuNet}: A two-stage, deterministic pipeline composed of a reconstruction U-Net that maps the zero-filled input to a magnitude image, followed by a separate segmentation U-Net. The learned sampling (LOUPE) and both networks are trained end-to-end with a composite objective that combines a reconstruction loss (MSE) and a segmentation loss (cross-entropy).

\textit{(2) LOUPE-UNet-RS}: Maintains the same two-stage architecture (reconstruction U-Net → segmentation U-Net), but adopts a staged training protocol: first train LOUPE together with the reconstruction U-Net; then freeze the mask and reconstruction U-Net and train only the segmentation U-Net on the reconstructed images. This isolates the segmentation learning while leveraging a learned sampling pattern.

\textit{(3) TradPat-UNet-RS}: Identical two-stage architecture as above, but replaces LOUPE with fixed, handcrafted sampling patterns (uniform random, variable density, or Poisson disk). The reconstruction U-Net is pretrained under each fixed pattern and then frozen; the segmentation U-Net is subsequently trained on the resulting reconstructed images.

\textbf{* Implementing segmentation directly based on measurements:} 

\textit{(1) LI-Net}: LI-Net consists of (i) a measurement encoder that encodes zero-filled reconstructions and produces a low-dimensional latent code (subsampled k-space data are measured using traditional patterns such as uniform random, Poisson, and variable density), and (ii) a segmentation decoder that decodes this latent code to a segmentation map. The decoder is obtained by pretraining a segmentation autoencoder on ground-truth segmentation; After the autoencoder is trained, the segmentation autoencoder remains frozen, and only the measurement encoder is optimized to align its output latent code with the latent code of the ground-truth segmentation map. At the inference time, the measurement encoder predicts the latent of ground-truth segmentation given the measurement, and then the predicted latent is decoded to the predicted segmentation.

\textit{(2) LOUPE-LI-Net}: Identical task head as LI-Net. The only difference is the sampling pattern used to generate subsampled k-space data, i.e., LOUPE-based learned sampling.

\begin{table*}[!t]
\centering
\caption{Training-time trainable parameters by method (task pipeline components). Counts are shown in millions (M) or thousands (K).}
\label{tab:param_trainable}
\begin{tabular}{l l}
\toprule
Method & Trainable Modules \\
\midrule
SemuNet         & Rec-UNet (31.0 M) + Seg-UNet (31.0 M) + LOUPE (16.4 K) = 62.1 M \\
TradPat-UNet-RS & Rec-UNet (31.0 M) + Seg-UNet (31.0 M) = 62.0 M \\
LOUPE-UNet-RS   & Rec-UNet (31.0 M) + Seg-UNet (31.0 M) + LOUPE (16.4 K) = 62.1 M \\
\midrule
LI-Net          & Seg-Enc (34.1 M) + Seg-Dec (49.5 M) + Measure-Enc (34.1 M) = 117 M \\
LOUPE-LI-Net    & Seg-Enc (34.1 M) + Seg-Dec (49.5 M) + Measure-Enc (34.1 M) + LOUPE (16.4 K) = 117 M \\
Tackle          & Rec-UNet (31.0 M) + Seg-UNet (31.0 M) + LOUPE (16.4 K) = 62.1 M \\ 
\midrule
 InfoMRI (Ours)         & Post-Enc (34.3 M) + Prior-Enc (34.3 M) + Seg-Dec (49.5 M) + Recon-Dec (49.5 M) + PGN (16.4 K) = 167 M \\
\bottomrule
\end{tabular}
\end{table*}

\begin{table*}[!t]
\centering
\caption{Inference-time total parameters by method (modules executed at test time). Counts are shown in millions (M) or thousands (K).}
\label{tab:param_inference}
\begin{tabular}{l l}
\toprule
Method &  Total Inference Params\\
\midrule
SemuNet          & Rec-UNet (31.0 M) + Seg-UNet (31.0 M) + LOUPE (16.4 K) = 62.1 M \\
TradPat-UNet-RS  & Rec-UNet (31.0 M) + Seg-UNet (31.0 M) = 62.0 M \\
LOUPE-UNet-RS    & Rec-UNet (31.0 M) + Seg-UNet (31.0 M) + LOUPE (16.4 K) = 62.1 M \\
\midrule
LI-Net           & Seg-Dec (49.5 M) + Measure-Enc (34.1 M) = 83.6 M \\
LOUPE-LI-Net     & Seg-Dec (49.5 M) + Measure-Enc (34.1 M) + LOUPE (16.4 K) = 83.6 M \\
Tackle           & Rec-UNet (31.0 M) + Seg-UNet (31.0 M) + LOUPE (16.4 K) = 62.1 M \\ 
\midrule
 InfoMRI (Ours)       & Prior-Enc (34.3 M) + Seg-Dec (49.5 M) + LOUPE (16.4 K) = 83.8 M \\
\bottomrule
\end{tabular}
\end{table*}

\textit{(3) Tackle}: The task pipeline first applies a reconstruction network to the two-channel measurements (LOUPE-based learned sampling) to obtain a single-channel magnitude image. A subsequent segmentation model (a standard U-shaped convolutional network followed by a sigmoid readout) then predicts the segmentation map from the reconstructed image. Two networks, together with the LOUPE-based learnable sampling, are trained end-to-end for predicting the final segmentation maps from input subsampled k-space data.

\textit{(4) InfoMRI (Proposed)}: A probabilistic, task-adapted pipeline that operates directly on undersampled measurements. Similar to LI-Net, InfoMRI consists of (i) a prior encoder that encodes the measurement into a low-dimensional latent code, (ii) a segmentation decoder that decodes this latent code to a segmentation map, (iii) a posterior encoder that encodes the ground-truth segmentation map. The difference is that InfoMRI additionally consists of a reconstruction decoder and uses PGN for amortizing optimization across different sampling ratios.

\textbf{* Comparison of the parameter counts:} 

To ensure fairness, we report two complementary quantities: (i) the number of trainable parameters used during task training (training-time task pipeline; sampling policy parameters are excluded unless trained jointly with the task), and (ii) the total parameters used at inference for the task pipeline. Exact counts are summarized in Tables~\ref{tab:param_trainable} and~\ref{tab:param_inference}.

Compared with prior works, InfoMRI carries a larger training-time parameter budget (167M vs. 62--117M; Table~\ref{tab:param_trainable}) because it includes (i) a posterior encoder used only to tighten the variational bound during learning, and (ii) a lightweight reconstruction decoder that stabilizes optimization and improves sample quality. Crucially, both modules are discarded at test time. The inference-time footprint (83.8M; Table~\ref{tab:param_inference}) is therefore comparable to LI-Net (83.6M), making deployment overhead modest.

This extra training capacity yields tangible benefits that we observe consistently across experiments:
\begin{itemize}
    \item Better uncertainty calibration at the \emph{same or similar} inference cost: InfoMRI achieves the lowest or among the lowest GED across clinically relevant accelerations (e.g., Table~\ref{tab:2x4x} and Table~\ref{tab:skm-tea}) while maintaining competitive DSC, indicating a closer match to the ground-truth posterior than deterministic counterparts.
    \item Structured uncertainty for downstream use: sampling from $q(\mathbf{t}\mid \mathbf{y})$ produces anatomically coherent hypotheses that can be marginalized by later modules, a capability not afforded by smaller deterministic heads.
    \item Amortized sampling across ratios in a single model: the PGN removes the need to train multiple models per acceleration, trading a one-time increase in training parameters for \emph{substantial} savings in total training runs and easier deployment.
\end{itemize}

In practice, test-time computation involves only the prior encoder, the segmentation decoder, and the PGN. Optional multi-sample evaluation (typically 4--8 samples suffice; Table~\ref{tab:n_samples}) modestly increases latency but materially improves calibration (GED). Taken together, the larger training-time parameter count is an acceptable and deliberate trade-off for improved robustness, calibrated uncertainty, and flexible sampling with minimal impact on inference complexity.


\end{document}